\documentclass[letterpaper,10pt,journal,twoside]{IEEEtran}

\IEEEoverridecommandlockouts                              

\usepackage{times}
\usepackage{graphicx}
\usepackage{subfigure}
\usepackage{capt-of}
\usepackage{amsmath,amssymb,amsopn,amstext,amsfonts,amsthm}
\usepackage{cancel}
\usepackage[space]{cite}
\usepackage{pdfsync}
\usepackage{balance}
\usepackage{color}
\usepackage{mathtools}
\usepackage{bm}
\usepackage{diagbox}
\usepackage{float}
\usepackage{epstopdf}
\usepackage{pifont}
\usepackage{fixltx2e}
\usepackage{multirow}
\usepackage{url}
\usepackage{verbatim}
\usepackage[linkcolor=black,citecolor=black,urlcolor=black,colorlinks=true]{hyperref}
\usepackage{soul,xcolor}
\usepackage[linesnumbered,ruled,vlined]{algorithm2e}
\usepackage{lipsum}
\usepackage{booktabs}
\usepackage{subfiles}
\usepackage{array,booktabs}
\usepackage{mwe}
\usepackage{tablefootnote}
\usepackage{nomencl}
\makenomenclature
\newcolumntype{M}[1]{>{\centering\arraybackslash}m{#1}}

\graphicspath{{./img_compress/}}
\DeclareGraphicsExtensions{.png,.jpg,.eps,.pdf,.jpeg}

\newtheorem{theorem}{Theorem}

\newtheorem{lemma}[theorem]{Lemma}
\newcommand{\etal}{\textit{et al.}}

\newcommand{\vect}{\mathbf}
\newcommand{\matr}[1]{\mathbf{#1}}
\newcommand{\rev}[1]{\textcolor{black}{#1}}

\makeatletter
\let\@oldmaketitle\@maketitle
\renewcommand{\@maketitle}{\@oldmaketitle
    \centering  
    \setcounter{figure}{0}
    \includegraphics[width=2.0\columnwidth]{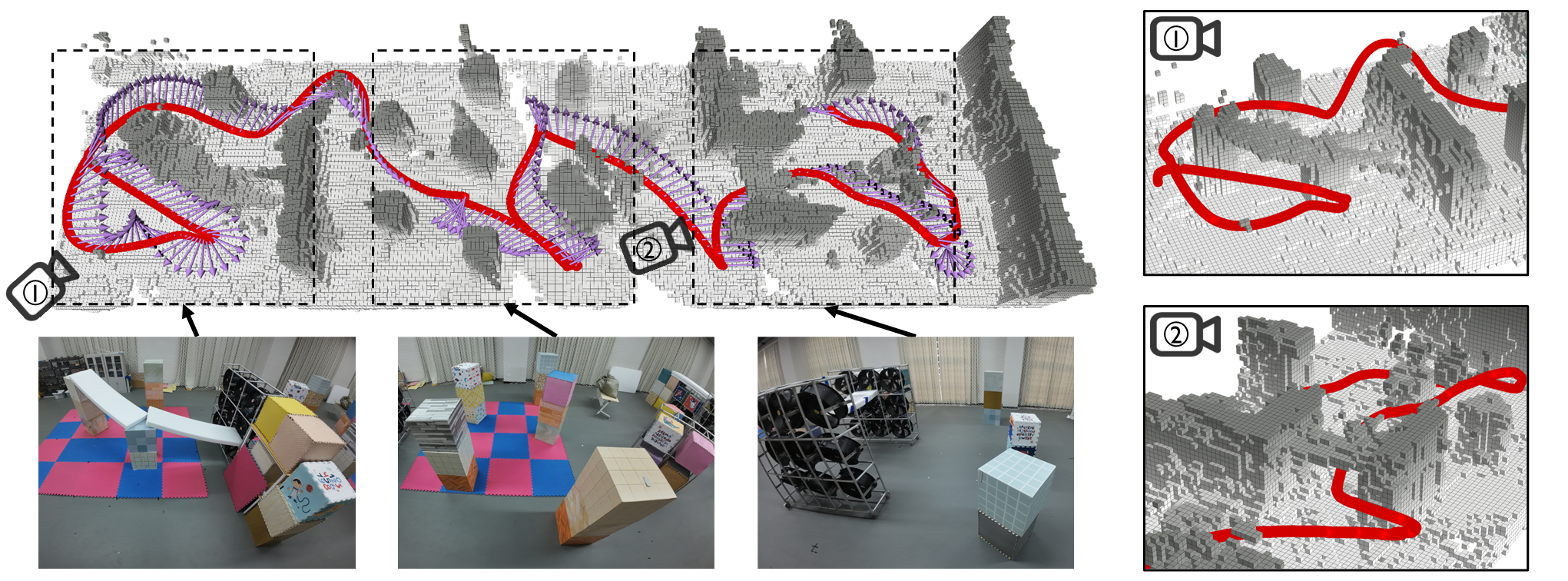}
    \vspace{-0.2cm}
    \captionof{figure}{A quadrotor with limited sensor field of view (FoV) autonomously explored a challenging cluttered indoor environment of dimension $24 \times 6 \times 2$ $m^3$. The online constructed volumetric map and the flight trajectory with \rev{the yaw direction as a purple arrow} are shown in the top-left image. Two detailed close-ups of the bridges from the volumetric map result are displayed in the right images. Video of the experiments is available at \url{https://youtu.be/BGH5T2kPbWw}.}
    \label{fig:realworld_105}
    \vspace{-0.8cm}
}
\makeatother

\title{FALCON: Fast Autonomous Aerial Exploration using Coverage Path Guidance}
\author{Yichen Zhang$^*$, Xinyi Chen$^*$, Chen Feng, Boyu Zhou$^\dagger$, and Shaojie Shen%
\thanks{$^*$ Yichen Zhang and Xinyi Chen are co-first authors.

$^\dagger$ Corresponding author: Boyu Zhou.

This work was supported by HKUST Postgraduate Studentship and HKUST-DJI Joint Innovation Laboratory.

Yichen Zhang, Xinyi Chen, Chen Feng and Shaojie Shen are with the Department of Electronic and Computer Engineering, Hong Kong University of Science and Technology, Hong Kong, China.

Boyu Zhou is with the Southern University of Science and Technology, Shenzhen, China, and the Sun Yat-Sen University, Zhuhai, China.

\tt\footnotesize $\{$yzhangec,xchencq,cfengag,eeshaojie$\}$@ust.hk

\tt\footnotesize uv.boyuzhou@gmail.com
}%
}

\begin{document}
\maketitle

\begin{abstract}
This paper introduces \textbf{FALCON}, a novel \textbf{F}ast \textbf{A}utonomous exp\textbf{L}oration framework using \textbf{CO}verage path guida\textbf{N}ce, which aims at setting a new performance benchmark in the field of autonomous aerial exploration. 
Despite recent advancements in the domain, existing exploration planners often suffer from inefficiencies such as frequent revisitations of previously explored regions.
\textbf{FALCON} effectively harnesses the full potential of online generated coverage paths in enhancing exploration efficiency.
The framework begins with an incremental connectivity-aware space decomposition and connectivity graph construction, which facilitate efficient coverage path planning.
Subsequently, a hierarchical planner generates a coverage path spanning the entire unexplored space, serving as a global guidance.
Then, a local planner optimizes the frontier visitation order, minimizing traversal time while consciously incorporating the intention of the global guidance.
Finally, minimum-time smooth and safe trajectories are produced to visit the frontier viewpoints.
For fair and comprehensive benchmark experiments, we introduce a lightweight \textit{exploration planner evaluation environment} that allows for comparing exploration planners across a variety of testing scenarios using an identical quadrotor simulator.
\rev{Additionally, an in-depth analysis and evaluation is conducted to highlight the significant performance advantages of \textbf{FALCON} in comparison with the state-of-the-art exploration planners based on objective criteria.}
Extensive ablation studies demonstrate the effectiveness of each component in the proposed framework.
Real-world experiments conducted fully onboard further validate \textbf{FALCON}'s practical capability in complex and challenging environments.
The source code of both the exploration planner \textbf{FALCON} and the exploration planner evaluation environment has been released to benefit the community\footnote{\url{https://github.com/HKUST-Aerial-Robotics/FALCON}}.
\end{abstract}


\section{Introduction}
\label{sec:intro}

\IEEEPARstart{A}{utonomous} exploration is the task of mapping unknown environments with mobile robots.
It is a fundamental component of various robotics applications, such as structural inspection \cite{hollinger2013active, bircher2015structural, yoder2016autonomous}, 3D reconstruction \cite{song2021view,feng2023predrecon,feng2024fc}, subterranean navigation \cite{mascarich2018multi, dang2019graph, yang2021graph}, and search and rescue operations \cite{erdelj2017help, marconi2012sherpa, li2023autotrans}.
Thanks to their agility and flexibility, aerial robots are well-suited for these applications in environments that are hazardous or inaccessible \rev{to} human operators.
\rev{For example, during underground cave mining or fire rescue operations, an autonomous aerial robot can efficiently construct detailed maps of the unknown spaces and gather valuable information for object searching and survivor rescue.}
Due to the limited battery life of aerial robot platforms, it is crucial to develop efficient exploration planners that cover accessible space as fast as possible.

Various existing methodologies have been proposed to enhance exploration efficiency \cite{cieslewski2017rapid, dharmadhikari2020motion}.
Early techniques, such as frontier-based \cite{yamauchi1997frontier} and next-best-view-based \cite{papachristos2017uncertainty} approaches, typically employ a greedy strategy that selects the next target according to immediate rewards.
This shortsighted strategy neglects global information, resulting in unnecessary revisitations of already explored areas.
To address this issue, global guidance considering all viewpoints awaiting visitation has been introduced \cite{dai2020fast, selin2019efficient}.
However, a gap persists between these solutions and the objective of the exploration task.
While the ultimate goal of the exploration task is to map entire unknown environments, most global guidance provided by existing methods focuses on visiting all frontier regions or subspaces, overlooking unknown areas beyond.
\rev{Due to the dynamically changing frontiers during exploration, this type of global guidance often provides highly inconsistent intention, leading to back-and-forth movements and failing to reflect the exploration task goal.}
Consequently, this discrepancy leads to inefficient exploration routes that frequently overlap with explored areas, reducing exploration efficiency.

Recently, researchers have attempted to bridge this gap by introducing coverage path (CP) of the entire unexplored space into exploration planning \cite{song2018surface,kan2020online}.
The coverage path serves as a more reasonable global guidance which \rev{better models the complete exploration process}.
However, the potential of coverage paths in improving exploration efficiency remains underexploited due to several limitations. 
Firstly, the coverage paths produced by these approaches often rely on simple space decompositions that inadequately capture the environment's connectivity and topology, potentially resulting in unreasonable coverage paths.
Secondly, many coverage path planning modules suffer from high online computational overhead due to the high complexity of problem resolution.
This impedes the subsequent planner from promptly responding and replanning when the latest environment information is received.
Furthermore, even when the coverage path offers reasonable guidance, the local trajectory may deviate significantly from the global guidance.
This inconsistency reflects the underutilization of the global coverage path's intention during local planning, thereby reducing the significance of the coverage paths.

To address these limitations, we present \textbf{FALCON}, a \textbf{F}ast \textbf{A}utonomous aerial robot exp\textbf{L}oration planner using \textbf{CO}verage path guida\textbf{N}ce, which further realizes the potential of coverage paths in improving exploration efficiency.
Whenever the map is updated using the latest sensor measurements, the entire exploration space is partitioned online into distinct \textit{zones} using the Connected Component Labeling algorithm, which separates disconnected regions based on the latest occupancy information.
From these zones, a connectivity graph is incrementally built to capture the underlying environment topology and boost coverage path planning efficiency. 
The hierarchical planner then utilizes the connectivity graph to promptly solve for a reasonable coverage path over zones, serving as global guidance.
Guided by this coverage path, the local planner optimizes the frontier visitation order using a Sequential Ordering Problem formulation, ensuring coherent motion aligned with the coverage path's intention and minimizing exploration duration.
Finally, minimum-time trajectories are generated to visit the frontier viewpoints while adhering to safety, smoothness and quadrotor dynamics constraints.
This framework reasonably plans coverage paths that consider environmental topology and consciously incorporates the intention of global guidance into local planning, demonstrating significantly enhanced exploration efficiency.

We would like to compare the performance of the proposed planner \textbf{FALCON} with other approaches.
However, to the best of our knowledge, there is a notable lack of an evaluation platform and standards that enable fair and comprehensive comparisons among aerial exploration planners.
Most existing works often restrict the simulation validation of exploration planners to a narrow selection of specific maps, typically inherited from previous studies. 
This may result in algorithms overfitting to those particular maps, without adequately assessing their capabilities in scenarios with diverse characteristics. 
Moreover, the evaluation criteria are often limited to \rev{exploration efficiency}, which only measures the exploration duration without considering other important aspects.
To bridge this gap, we introduce a unified environment and a set of objective criteria to evaluate the efficacy of aerial exploration planners in a fair and comprehensive manner.
On the one hand, we develop a lightweight software-in-the-loop \textit{exploration planner evaluation environment}.
This environment allows for fair and extensive benchmark comparisons with a bundle of state-of-the-art exploration planners \cite{bircher2016receding, zhou2021fuel, zhou2023racer, yu2023echo, zhao2023autonomous} using an identical quadrotor simulator and diverse testing environments.
On the other hand, we propose the \textit{VECO} criteria that an ideal and robust exploration planner should satisfy.
The \textit{VECO} criteria are namely \textbf{V}ersatility, \rev{\textbf{E}xploration Efficiency}, \textbf{C}ompleteness, and \rev{Resp\textbf{O}nsiveness}.

Under the exploration planner evaluation environment, we conduct extensive experiments comparing \textbf{FALCON} with state-of-the-art exploration planners \cite{bircher2016receding, zhou2021fuel, zhou2023racer, yu2023echo, zhao2023autonomous}, exhibiting its extraordinary performance.
The results reveal that \textbf{FALCON} not only achieves $13.81\%\sim 29.67\%$ faster explorations in various testing scenarios, but also satisfies all the \textit{VECO} criteria.
We also present an in-depth analysis of the characteristics of the proposed and benchmarked exploration planners according to the \textit{VECO} criteria.
Additionally, ablation experiments demonstrate the effectiveness of each individual component of \textbf{FALCON}. 
Furthermore, real-world experiments conducted fully onboard using a customized quadrotor in challenging environments confirm the practical operation of \textbf{FALCON}.
The source code has been released to benefit the community.
In summary, the contributions are:
\begin{enumerate}
  \item An incremental connectivity-aware space decomposition and connectivity graph construction method, which captures the environmental topology and facilitates efficient exploration coverage path planning.
  \item A hierarchical exploration planning approach which generates reasonable coverage paths serving as global guidance and optimizes local frontier visitation order preserving the intention of coverage paths.
  \item An exploration planner evaluation environment and evaluation criteria that support fair and comprehensive experiments in comparison with state-of-the-art exploration planners across diverse testing scenarios.
  \item Extensive validation of the proposed exploration planner through benchmark comparisons, ablation studies and real-world experiments. The source code of both \textbf{FALCON} and the evaluation environment has been released.
\end{enumerate}

\bstctlcite{bstctl:etal, bstctl:nodash, bstctl:simpurl}

\begin{table*}[t]
  \caption{A summary of existing exploration studies}
  \centering
  \vspace{-0.2cm}
  \begin{tabular}{cccccc} \toprule\toprule
    \textbf{\begin{tabular}[c]{@{}c@{}}Planning\\Approach\end{tabular}} & \textbf{\begin{tabular}[c]{@{}c@{}}Global\\Consideration\end{tabular}} & \textbf{Prioritization} & \textbf{\begin{tabular}[c]{@{}c@{}}Planning\\Dimension\end{tabular}} & \textbf{References} & \textbf{Sensor Type}\\ \midrule
    \multirow{6}{*}{\begin{tabular}[c]{@{}c@{}}Information\\gain-based\end{tabular}} & \multirow{6}{*}{N/A} & \multirow{4}{*}{Rapid exploration} & \multirow{2}{*}{2D} & \cite{bai2016information} & Laser scanner \\ 
                                                                    &     &                               & & \cite{kaufman2018autonomous}  & LiDAR  \\ \cmidrule{4-6}
                                                                    &     &           & \multirow{2}{*}{3D} & \cite{whaite1997autonomous}  & Laser scanner  \\ 
                                                                    &     &                               & & \cite{tabib2016computationally}  & Camera  \\ \cmidrule{3-6}
                                                                    &     & Reconstruction accuracy    & 3D & \cite{palazzolo2017information}, \cite{palazzolo2018effective} & Camera  \\ \cmidrule{3-6}
                                                                    &     & Localization uncertainty   & 2D & \cite{stachniss2005information}, \cite{bourgault2002information}, \cite{carrillo2018autonomous}& Laser scanner  \\
    \cmidrule{1-6}
    \multirow{6}{*}{\begin{tabular}[c]{@{}c@{}}Sampling-\\based\end{tabular}} & \multirow{6}{*}{N/A} & \multirow{2}{*}{Rapid exploration} & \multirow{2}{*}{3D} & \cite{bircher2016receding}, \cite{bircher2018receding}, \cite{witting2018history}, \cite{duberg2022ufoexplorer} & Camera  \\ 
                                    &                      &                                              & & \cite{dharmadhikari2020motion}  & LiDAR  \\ \cmidrule{3-6}
                                    &                      & Reconstruction accuracy                   & 3D & \cite{schmid2020efficient} & Camera  \\ \cmidrule{3-6}
                                    &                      & \multirow{2}{*}{Localization uncertainty} & \multirow{2}{*}{3D} & \cite{papachristos2017uncertainty}, \cite{papachristos2019autonomous}, \cite{papachristos2017autonomous}& Camera  \\ 
                                    &                      &                                              & & \cite{suresh2020active} & Multibeam sonar  \\ \cmidrule{3-6}
                                    &                      & Object-centric exploration                & 3D & \cite{dang2018autonomous}, \cite{dang2018visual} & Camera  \\

    \cmidrule{1-6}
    \multirow{9}{*}{\begin{tabular}[c]{@{}c@{}}Frontier-\\based\end{tabular}} & \multirow{9}{*}{N/A} & \multirow{6}{*}{Rapid exploration} & \multirow{3}{*}{2D} & \cite{yamauchi1997frontier}, \cite{yamauchi1998frontier} & Laser scanner, Multibeam sonar \\ 
                                    &                      &                                              & & \cite{freda2005frontier}  & Ultrasonic rangefinder  \\ 
                                    &                      &                                              & & \cite{gao2018improved}  & LiDAR  \\ \cmidrule{4-6}
                                    &                      &                          & \multirow{3}{*}{3D} & \cite{shen2012stochastic}, \cite{shen2012autonomous}  & Camera, Laser scanner \\ 
                                    &                      &                                              & & \cite{cieslewski2017rapid}, \cite{zhu20153d}, \cite{senarathne2016towards}  & Camera  \\ 
                                    &                      &                                              & & \cite{faria2019applying}  & LiDAR  \\ \cmidrule{3-6}
                                    &                      & Reconstruction accuracy                   & 3D & \cite{yoder2016autonomous} & LiDAR, Camera  \\ \cmidrule{3-6}
                                    &                      & Localization uncertainty                  & 2D & \cite{stachniss2004exploration} & Laser scanner  \\ \cmidrule{3-6}
                                    &                      & Object-centric exploration                & 3D & \cite{dornhege2013frontier} & Camera, Laser scanner  \\

    \cmidrule{1-6}
    \multirow{9}{*}{Hybrid} & \multirow{2}{*}{N/A}       & Rapid exploration                  & 3D & \cite{dai2020fast}, \cite{yu2023echo} & Camera  \\ \cmidrule{3-6}
                            &                            & Object-centric exploration         & 3D & \cite{papatheodorou2023finding}  & Camera  \\ \cmidrule{2-6}
                            & Graph                      & Rapid exploration                  & 3D & \cite{dang2019graph}, \cite{yang2021graph}  & LiDAR  \\ \cmidrule{2-6}
                            & \multirow{4}{*}{Frontiers} & \multirow{2}{*}{Rapid exploration} & \multirow{2}{*}{3D} & \cite{tang2023bubble}, \cite{cao2021tare}, \cite{cao2021exploring}  & LiDAR  \\ 
                            &                            &                                    &    & \cite{selin2019efficient}, \cite{zhou2021fuel}, \cite{zhao2023autonomous}  & Camera  \\ \cmidrule{3-6}
                            &                            & Localization uncertainty           & 3D & \cite{zhang2022exploration} & Camera  \\ \cmidrule{3-6}
                            &                            & Multi-robots exploration           & 3D & \cite{yan2023mui}, \cite{cao2023representation} & LiDAR  \\ \cmidrule{2-6}
                            & \multirow{5}{*}{\begin{tabular}[c]{@{}c@{}}Coverage\\Path\end{tabular}} & \multirow{3}{*}{Rapid exploration} & \multirow{2}{*}{2D} & \cite{zhao2023tdle} & LiDAR  \\ 
                            &                            &                                    &    & \cite{kan2020online} & LiDAR, Camera  \\ \cmidrule{4-6}
                            &                            &                                    & \rev{3D} & \rev{\textbf{Proposed}} & \rev{Camera}  \\ \cmidrule{3-6}
                            &                            & Reconstruction accuracy            & 3D & \cite{song2018surface}, \cite{song2020online} & Camera  \\ \cmidrule{3-6}
                            &                            & Multi-robots exploration           & 3D & \cite{zhou2023racer} & Camera  \\

    \toprule\toprule
  \end{tabular} \\
  \vspace{-0.4cm}
  \label{tab:literature}
\end{table*}

\section{Related Work}
\label{sec:related}

Autonomous exploration approaches have varying prioritizations depending on application context, including rapid exploration completion \cite{cieslewski2017rapid, dharmadhikari2020motion}, reconstruction accuracy \cite{yoder2016autonomous, schmid2020efficient}, localization uncertainty \cite{bourgault2002information, zhang2022exploration}, object-centric exploration \cite{dang2018autonomous, papatheodorou2023finding}, and multi-robots collaboration \rev{exploration} \cite{zhou2023racer, yan2023mui}.
Regardless of prioritization, the following literature review is presented according to the methodology of exploration planning. 
A summary of the reviewed literature is provided in \rev{Table~\ref{tab:literature}}.

\subsection{Classical Methods}
Information gain-based exploration \rev{has} been widely studied in the past decades \cite{whaite1997autonomous, bourgault2002information}, where candidate viewpoints are chosen based on the expected information gain provided by the observations.
Bai \etal \cite{bai2016information} use Bayesian optimization to select near-optimal sensing actions that reduce map entropy.
Tabib \etal \cite{tabib2016computationally} consider the mutual information between sensor measurements and an environment model using Cauchy-Schwarz quadratic mutual information (CSQMI) \cite{charrow2015csqmi} for exploration of pits and caves.
Kaufman \etal \cite{kaufman2018autonomous} perform 3D Bayesian probabilistic occupancy grid mapping and project the stochastic properties of \rev{the} occupancy grid to 2D spaces for numerically efficient exploration planning.
Besides considering map entropy, Carrillo \etal \cite{carrillo2018autonomous} propose a utility function combining the Shannon and the Renyi entropy which balances between the robot localization and map uncertainty.
Alternatively, Stachniss \etal \cite{stachniss2005information} \rev{considers} vehicle pose and map uncertainty simultaneously by evaluating the expected entropy change of the Rao-Blackwellized particle filter as information gain.
Palazzolo \etal \cite{palazzolo2017information, palazzolo2018effective} select the next best viewpoints that provide the most expected entropy change of the belief about the world state for precise reconstruction.
Bissmarck \etal \cite{bissmarck2015efficient} compare various approaches to compute the information gain for candidate viewpoints selection.

Sampling-based exploration is one of the classic methods, where candidate viewpoints are randomly generated using Rapidly-exploring Random Trees (RRT) \cite{lavalle98rapidly} and the next best view (NBV) \cite{connolly1985determination} is selected based on information gain.
Bircher \etal \cite{bircher2016receding} first introduces the concept of NBV to autonomous exploration using a receding horizon scheme and later \rev{applies} it to surface inspection \cite{bircher2018receding}.
Based on NBV, several subsequent works employ various information gain metrics for different applications.
For instance, Papachristos \etal \cite{papachristos2017uncertainty} minimize the localization and mapping uncertainty during exploration, which is later deployed to visually-degraded underground mine environments \cite{papachristos2019autonomous} as well as \rev{dark environments} such as an old city tunnel \cite{papachristos2017autonomous}.
Dang \etal \cite{dang2018visual} further incorporate the saliency-annotated volumetric mapping to drive the robot towards visually \rev{salient} objects in the environment.
Witting \etal \cite{witting2018history} incorporate navigation history into the sampling scheme to facilitate \rev{the} quick finding of informative regions and avoid being stuck at local minima.
Dharmadhikari \etal \cite{dharmadhikari2020motion} exploit aerial robot dynamics to generate dynamically feasible motion primitives for faster exploration.
Suresh \etal \cite{suresh2020active} develop an active SLAM exploration framework to reduce vehicle pose uncertainty in underwater environments by balancing volumetric exploration and revisitation utilizing RRT nodes.
To ensure scalability, Duberg \etal \cite{duberg2022ufoexplorer} maintain a graph structure for efficient path planning and use a simple exploration heuristic driving the robot to the closest unknown space.

Frontier-based exploration is another popular approach, where frontiers, defined as the boundary between known and unknown space, are identified as targets during exploration.
The technique was initially introduced in \cite{yamauchi1997frontier, yamauchi1998frontier}, where the next target is simply chosen as the closest frontier.
Gao \etal \cite{gao2018improved} modify \cite{yamauchi1997frontier} by adding the robot heading direction and rotation cost into account to maintain robot's orientation for more efficient exploration.
Freda \etal \cite{freda2005frontier} use a frontier-based modification of the Sensor-based Random Tree (SRT) method to bias exploration motion towards frontier regions to improve efficiency.
Stachniss \etal \cite{stachniss2004exploration} extend frontier-based exploration to allow actions that actively revisit explored regions and close loops to reduce uncertainty in SLAM.
Although the aforementioned frontier-based exploration methods operate in 2D space, researchers have made efforts to extend the concept of frontiers into 3D exploration.
Shen \etal \cite{shen2012stochastic, shen2012autonomous} adopts a stochastic differential equation to identify the most significant particle expansion region as \rev{the} next target frontier.
This method avoids poor 3D frontiers generated by traditional 2D frontier-based methods due to incomplete sensor information.
Dornhege \etal \cite{dornhege2013frontier} extend 2D frontier-based method to 3D by introducing void cells to efficiently determine high-visibility locations for autonomous robot exploration and victims searching.
Zhu \etal \cite{zhu20153d} extend frontiers to 3D space using OctoMap representation \cite{wurm2010octomap} and the closest frontier is selected as the goal frontier.
Senarathne \etal \cite{senarathne2016towards} also utilize OctoMap but opt for computationally efficient surface frontiers rather than free-space frontiers to actively map the object surfaces.
To facilitate high-speed exploration, Cieslewski \etal \cite{cieslewski2017rapid} selects the frontier that minimizes the change in velocity to maintain high flight speed and enhance exploration efficiency.
Faria \etal \cite{faria2019applying} combines \rev{frontier-based} exploration with Lazy Theta* path planning to reduce the number of resolution iterations for efficient inspection of large 3D structures.

While these methods are capable of completing exploration \rev{tasks}, their efficiency remains limited and unsatisfactory.
Information gain-based and sampling-based methods enable the explicit quantification of information gain for each \rev{candidate's} viewpoint, but they come with a high computational burden.
More importantly, these methods commonly utilize \rev{a} greedy strategy that selects the next target according to immediate rewards without a holistic perspective on the entire exploration space.
The lack of global consideration can lead to inefficient exploration tours containing back-and-forth movements unnecessarily revisiting explored regions.

\begin{figure*}[t]
	\centering
  \includegraphics[width=1.8\columnwidth]{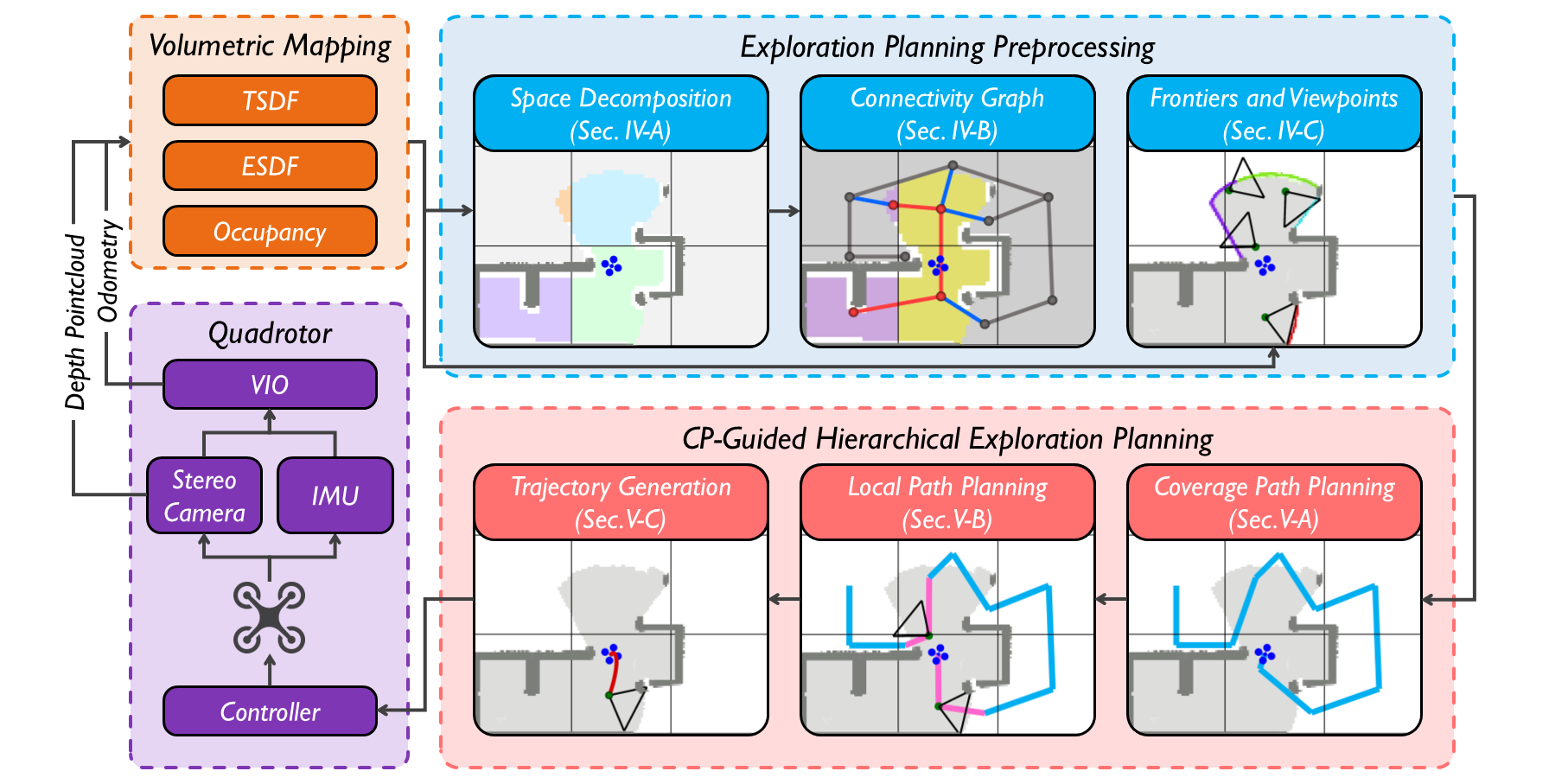}  
  \vspace{-0.3cm}
  \caption{\label{fig:system_overview} Overview of the proposed planner \textbf{FALCON} for fast autonomous exploration. The framework comprises two main components: exploration planning \rev{preprocessing} and CP-guided hierarchical exploration planning. The former part provides fundamental information, including space decomposition and \rev{the} connectivity graph for coverage path planning, as well as extraction of frontiers and viewpoints for local planning. The latter part performs hierarchical exploration planning, which generates a global coverage path (CP) spanning the entire unexplored space and optimizes frontier visitation order  consciously incorporating CP's intention. These modules are consistently updated and replanned until exploration concludes.}
  \vspace{-0.4cm}
\end{figure*}

\subsection{Hybrid Methods With Global Consideration}
In \rev{the} recent decade, researchers attempt to address this problem by capturing and exploiting global information to guide the exploration process.
These hierarchical exploration planning approaches are often hybrid, combining several classical methods mentioned above.
For example, Dai \etal \cite{dai2020fast} propose a hybrid frontier-based and sampling-based strategy that \rev{samples} candidate next-views at the frontier regions and \rev{evaluates} them according to a utility function expressing expected information gain over time.
Selin \etal \cite{selin2019efficient} \rev{utilizes} a receding horizon next-best-view planning approach for local exploration while employing frontier exploration for global planning.
Zhou \etal \cite{zhou2021fuel} first \rev{generates} efficient global paths for detected frontiers and then sample viewpoints around the frontiers as local refinements.
Tang \etal \cite{tang2023bubble} and Yu \etal \cite{yu2023echo} further improve the viewpoints generation and determination strategies of \cite{zhou2021fuel}, demonstrating higher exploration efficiency.
Zhao \etal \cite{zhao2023autonomous} refines the planning \rev{strategy} of \cite{zhou2021fuel} by considering the frontiers location with respect to the exploration boundary.
Yang \etal \cite{yang2021graph} adopt a bifurcated planning strategy to explore local distinctive frontier regions, guided by a sparse topological graph globally.
Cao \etal \cite{cao2021tare, cao2021exploring} first \rev{plans} a global path through subspaces that \rev{requires} detailed exploration and then samples a set of viewpoints to cover the current local subspace.
This method is later extended to \rev{multi-robot} exploration \cite{cao2023representation}.
Although the aforementioned hierarchical approaches are \rev{computationally} efficient since sampling efforts \rev{are} focused on frontier regions, they still suffer from unsatisfactory exploration efficiency.
This inefficiency arises from the fundamental disparity between their concentration solely on frontier regions and the objective of exploration tasks in mapping unknown regions.

Coverage paths (CPs) have been utilized as global paths in some recent exploration works.
Coverage path planning, a well-studied problem over past decades, involves finding a route that traverses every point of a specific area or volume of interest while avoiding obstacles \cite{galceran2013survey}.
Conventional 2D approaches employ predefined patterns to cover subspaces \rev{separated} by boustrophedon \cite{choset1998coverage}, trapezoidal \cite{oksanen2009coverage}, morse-based \cite{choset2000exact} or grid-based \cite{gabriely2002spiral} decomposition to handle obstacles.
For 3D \rev{environments} with higher structural complexity, recent works solve the problem using a \rev{Traveling Salesman Problem (TSP)} formulation \cite{englot2012sampling, bircher2016three, cao2020hierarchical}.
However, these methods are not well-suited for \rev{the} exploration task due to the high computational complexity.
These single-time computation methods operate on a predetermined map, while \rev{the} exploration task necessitates real-time and incremental coverage path planning when the map is updated.

Recently, several CP-guided hierarchical exploration approaches have been proposed, introducing the coverage path of the entire unexplored space as a more reasonable global guidance.
Zhao \etal \cite{zhao2023tdle} propose a LiDAR exploration system that arranges the visitation order of the uniformly and dynamically divided subregions serving as global guidance.
Kan \etal \cite{kan2020online} alternatively \rev{employs} a hex decomposition for coverage path planning as well as generating circular and straight-line paths to explore the current subregion.
However, these two methods work only for 2D spaces and are not applicable to aerial robot exploration in 3D environments.
Song \etal \cite{song2018surface, song2020online} propose a global coverage planning algorithm that partitions the entire map into sectors, and a local inspection planning algorithm that samples informative viewpoints of low-confidence surfaces.
However, the algorithm is tailored for accurate 3D modeling tasks, performing rather slow flight speed and long exploration duration.
Zhou \etal \cite{zhou2023racer} extends \cite{zhou2021fuel} by integrating a global coverage path of hgrids \cite{ericson2004real}, which \rev{guides} the swarm agents to explore different unknown regions with balanced workloads.
However, hgrids divide the space uniformly without considering environmental obstacles, which can potentially lead to unreasonable coverage paths.
Moreover, these approaches utilizing coverage path fail to effectively integrate the global guidance intention into local planning, causing exploration movements to deviate from the intended direction occasionally.

\section{System Overview}
\label{sec:system_overview}

\subsection{\rev{Problem Definition}}
\label{subsec:problem_def}
The problem considered in this paper is exploring an unknown and bounded 3D space $V \subset \mathbb{R}^3$ using a fully autonomous aerial robot and constructing a complete volumetric map of the accessible space $V_\text{acc} \subset V$.
\rev{The volumetric map consists of a 3D grid of voxels, where each voxel corresponds to one of three states: free, occupied, or unknown.}
The sensor used for localization and exploration is a \rev{forward}-looking stereo camera with limited FoV\rev{, which is modeled as a viewing frustum defined by its horizontal FoV, vertical FoV, and sensing depth}.
The exploration planner should be able to provide safe and feasible trajectories, followed by which, the aerial robot perceives information covering $V_\text{acc}$ with minimal exploration duration.

\subsection{\rev{Proposed Framework Overview}}
\label{subsec:overall_framework}
An overview of the proposed exploration framework is depicted in Fig.~\ref{fig:system_overview}, consisting of two components: exploration planning \rev{preprocessing} (Sec.~\ref{sec:preliminaries}) and CP-guided hierarchical exploration planning (Sec.~\ref{sec:cp_planning}).
Upon updating the voxel map with \rev{the} latest received sensor measurements, exploration planning \rev{preprocessing is} performed promptly within the bounding box of \rev{the} updated map.
\rev{The preprocessing phase involves an incremental space decomposition (Sec.~\ref{subsec:space_decom}), which enables the construction of a connectivity graph (Sec.~\ref{subsec:connectivity_graph}) to boost the efﬁciency of coverage path planning.
Additionally, \rev{the} frontier is identified with \rev{a} careful selection of viewpoint representatives (Sec.~\ref{subsec:frontier}).
Subsequently, the hierarchical exploration planner first conducts coverage path planning (Sec.~\ref{subsec:cp_construct}), providing global guidance for the local planner.
Then, the local path planning optimizes the frontier visitation order while maintaining consistency with the intention of the coverage path (Sec.~\ref{subsec:CP_guided_planning}) and generates executable trajectories (Sec.~\ref{subsec:traj_gen}).}
The \rev{preprocessing} and exploration motion are consistently updated and replanned based on the latest sensor measurements.
The exploration concludes when no \rev{frontier remains} in the map.

\subsection{\rev{Notation}}
\label{subsec:notation}
For clarity and ease of reference, key notation definitions are presented below, which are represented in the world frame.

\nomenclature[01]{$V$}{\rev{The 3D space in the exploration task bounding box.}}

\nomenclature[02]{$V_\text{acc}$}{\rev{The accessible space within $V$.}}

\nomenclature[03]{$\Gamma$}{\rev{The list of cells.}}

\nomenclature[03]{$\gamma_i$}{\rev{An instance of a cell.}}

\nomenclature[04]{$\mathcal{Z}$}{\rev{The list of zones.}}

\nomenclature[04]{$z_i$}{\rev{An instance of a zone.}}

\nomenclature[05]{$\mathcal{B}_t$}{\rev{The bounding box of the updated map at time $t$.}}

\nomenclature[06]{$\mathcal{G}$}{\rev{The connectivity graph.}}

\nomenclature[07]{$\mathcal{G}_f$}{\rev{The free subgraph of the connectivity graph.}}

\nomenclature[08]{$\mathcal{G}_u$}{\rev{The unknown subgraph of the connectivity graph.}}

\nomenclature[09]{$E_p$}{\rev{The portal edges of the connectivity graph.}}

\nomenclature[10]{$\vect{p}_c$}{\rev{The current position.}}

\nomenclature[11]{$\matr{C}_\text{cp}$}{\rev{The cost matrix used when solving the coverage path planning problem.}}

\nomenclature[12]{$\bar{\Lambda}$}{\rev{The coverage path produced from the coverage path planning process.}}

\nomenclature[13]{$\Lambda$}{\rev{The reduced coverage path used in the local path planning.}}

\nomenclature[14]{$\matr{C}_\text{sop}$}{\rev{The cost matrix used when solving the local path planning problem.}}

\printnomenclature

\section{Exploration Planning \rev{Preprocessing}}
\label{sec:preliminaries} 
At the beginning of each exploration planning iteration, several \rev{preprocessing} steps are performed online to facilitate efficient coverage path planning.
Whenever the map is updated, \rev{the entire predefined exploration space is partitioned by a coarse-to-fine space decomposition.}
Additionally, a connectivity graph capturing the underlying environment topology is incrementally constructed.
\rev{Frontiers} are also identified within the map update regions with frontier viewpoint representatives carefully chosen.

\subsection{Connectivity-Aware Space Decomposition}
\label{subsec:space_decom}

\begin{figure}[t]
	\centering
  \includegraphics[width=0.9\columnwidth]{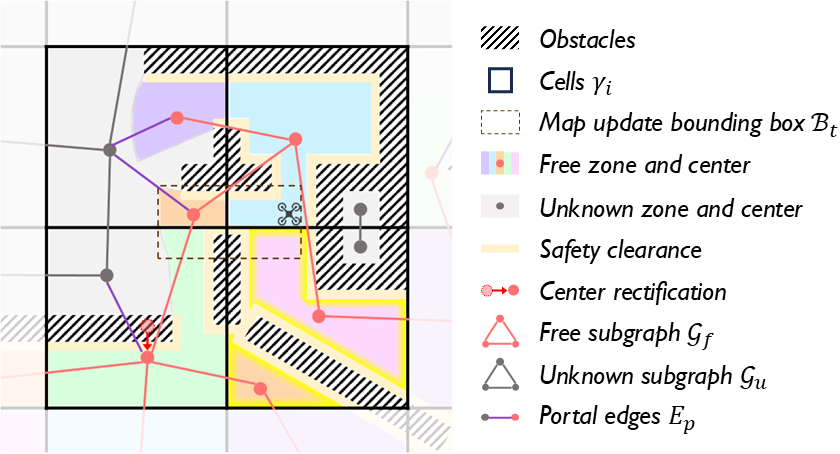}  
  \vspace{-0.2cm}
  \caption{\label{fig:space_decom} A snapshot illustrating the results of space decomposition and connectivity graph construction during the exploration. The surrounding transparent cells are not intersected with \rev{the} bounding box of \rev{the} updated map and kept unchanged. The two zones highlighted in yellow are connected by an inaccessible narrow corridor and consequently grouped into seperate zones. The middle-right rectangle is an inaccessible hollow and excluded from exploration planning utilizing the connectivity graph.}
  \vspace{-0.6cm}
\end{figure}

The entire exploration space is constantly decomposed based on the voxel connectivity given the \rev{latest} occupancy information, serving as elementary task units for coverage path planning.
Before the exploration begins, an initial coarse decomposition partitions the entire exploration space into uniform \textit{cells} $\Gamma$ with size proportional to the onboard camera's FoV.
\rev{In our case, the cell size $s_\text{cell}$ is set equal to the maximum sensing depth.}
Whenever the map is updated during the exploration, the cells intersected with the bounding box of \rev{the} updated map $\mathcal{B}_t$ are further decomposed into \rev{disjoint \textit{zones}} $\mathcal{Z}$ as shown in Fig.~\ref{fig:space_decom}.
Unlike naive approaches that adopt uniform decomposition, our method meticulously \rev{separates} disconnected regions based on the latest occupancy information.
This fine decomposition is formulated as a 3D extension of the Connected \rev{Components} Labeling (CCL) problem \cite{gonzalez2009digital}, a 2D image processing technique described below.

\begin{lemma}[\bfseries Connected Components Labeling] \label{lemma:ccl}
    The connected components labeling algorithm identifies individual components in an image based on pixel connectivity and assigns unique labels to connected pixels as separate objects.
\end{lemma}

The CCL algorithm operates on individual cells, treating the voxels in the cell as elements to be labeled, analogous to pixels in a 2D image.
Apart from the standard voxel states of occupied and unknown, free voxels are further classified as either \textit{safe-free} or \textit{unsafe-free}, depending on whether they violate the minimum safety clearance distance $d_{\text{min}}$.
During the labeling process, unknown and safe-free voxels are labeled while occupied and unsafe-free voxels are omitted.
Neighboring voxels are considered connected and assigned the same label if and only if they are both safe-free or both unknown.
Voxels with the same label are grouped together into a \textit{free zone} or an \textit{unknown zone} based on their occupancy status.
By \rev{omitting} unsafe-free voxels, regions connected by inaccessible narrow corridors are grouped into separate zones, such as the two highlighted zones in Fig.~\ref{fig:space_decom}.
Then the initial center of each zone is computed as the average position of all its voxels.
However, since convexity is not guaranteed by connectivity, the initial center may fall into obstacles or other zones.
In such cases, the center is rectified into the position of the nearest voxel within its zone, as demonstrated in Fig.~\ref{fig:space_decom}.

\begin{algorithm}[t!]
  \caption{\rev{Incremental Connectivity Graph Update}}
  \label{alg:connectivity_graph}
  \KwIn{cell list $\Gamma$, zone list $\mathcal{Z}$, map update bounding box $\mathcal{B}_t$, \rev{connectivity graph $\mathcal{G}$}}
  \KwOut{\rev{updated connectivity graph $\mathcal{G}$}}
  \BlankLine

  \rev{$\mathcal{G}$.getSubgraph($\mathcal{G}_f$, $\mathcal{G}_u$, $E_p$)}
  
  \ForEach{\textnormal{$\gamma_i \in \Gamma$ \rev{with} $\gamma_i \cap \rev{\mathcal{B}_t} \neq \emptyset$}}{
    $\mathcal{G}$.clearVerticesAndEdges($\gamma_i$)

    \ForEach{\textnormal{$z \in \gamma_i.\mathcal{Z}$}}{
      $\mathcal{G}$.addVertex($z.\vect{c}$) 
    }

    \ForEach{\textnormal{neighborhood $\gamma_j$ of $\gamma_i$}}{
      \ForEach{\textnormal{$z_m \in \gamma_i.\mathcal{Z}$, $z_n \in \gamma_j.\mathcal{Z}$}}{
        \If{\textnormal{$z_m$ and $z_n$ both free}}{
          $p \leftarrow$ restrictedA*($z_m.\vect{c}, z_n.\vect{c}, \gamma_i \cup \gamma_j$)

          \If{\textnormal{$p$ exists}}{
            $\mathcal{G}_f$.addEdge($z_m.\vect{c}, z_n.\vect{c}, p$) 
          } 
        }
        \ElseIf{\textnormal{$z_m$ and $z_n$ both unknown}}{
          $p \leftarrow$ restrictedA*($z_m.\vect{c}, z_n.\vect{c}, \gamma_i \cup \gamma_j$)

          \If{\textnormal{$p$ exists}}{
            $\mathcal{G}_u$.addEdge($z_m.\vect{c}, z_n.\vect{c}, p$) 
          } 
        }
      }
    } 
    
    \ForEach{\textnormal{free $z_f \in \gamma_i.\mathcal{Z}$ and unknown $z_u \in \gamma_i.\mathcal{Z}$}}{
      $p \leftarrow$ restrictedA*($z_f.\vect{c}, z_u.\vect{c}, \gamma_i$)

      \If{\textnormal{$p$ exists}}{
        $E_p$.addEdge($z_f.\vect{c}, z_u.\vect{c}, p$) 
      }
    }
  }

\end{algorithm}
\setlength{\textfloatsep}{0pt}

\vspace*{-0.4cm}
\subsection{Incremental Connectivity Graph Construction}
\label{subsec:connectivity_graph}
Based on the identified zones, a connectivity graph $\mathcal{G}$ is incrementally constructed, where the zone centers serve as vertices and the connectivity between zones serves as edges.
Note that the zone-level connectivity here is different from the inter-voxel connectivity mentioned in Sec.~\ref{subsec:space_decom}.
As illustrated in Fig.~\ref{fig:space_decom}, the connectivity graph consists of two subgraphs: $\mathcal{G}_f = (V_f, E_f)$ and $\mathcal{G}_u = (V_u, E_u)$ capturing free and unknown regions respectively, along with portal edges $E_p$ connecting them, i.e.,
$\mathcal{G} = (V_f \cup V_u, E_f \cup E_u \cup E_p)$

\begin{figure}[t]
	\centering
  \includegraphics[width=0.9\columnwidth]{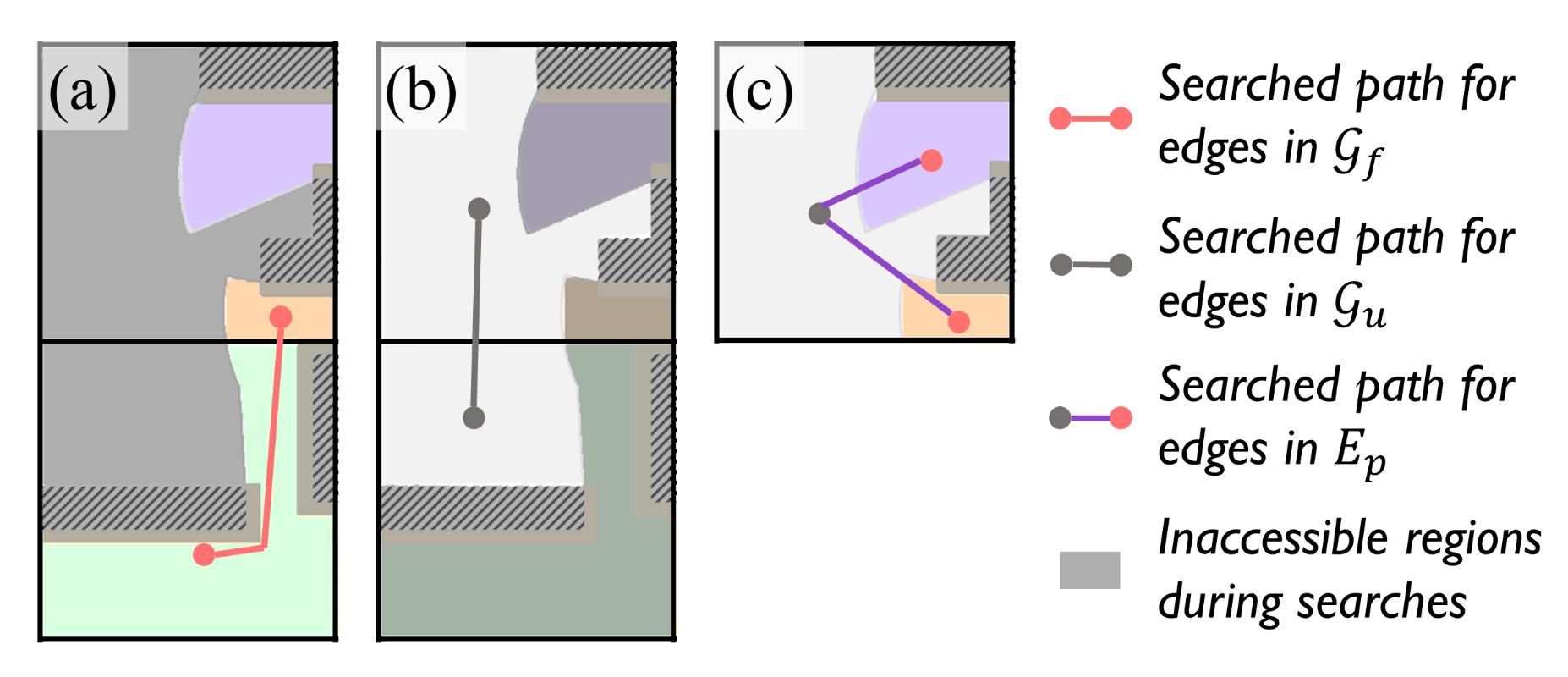}  
  \vspace{-0.4cm}
  \caption{\label{fig:A*} Examples for the three types of restricted A* performed when constructing \rev{(a) the free subgraph $\mathcal{G}_f$, (b) the unknown subgraph $\mathcal{G}_u$ and (c) the portal edges $E_p$ respectively}.}
  \vspace{-0.2cm}
\end{figure}

The connectivity graph construction process is detailed in Alg.~\ref{alg:connectivity_graph}. 
For each cell $\gamma_i$ involved with the bounding box $\mathcal{B}_t$, the centers of free and unknown zones \rev{contribute} to vertices $V_f$ and $V_u$ respectively.
The edges are constructed based on three types of restricted A* searches, as indicated in \rev{lines 9, 13, 17} of Alg.~\ref{alg:connectivity_graph} and exemplified in Fig.~\ref{fig:A*}.
For edges $E_f$ in the free subgraph $\mathcal{G}_f$, connectivity is evaluated pairwise between each free vertex $z_m$ in $\gamma_i$ and $z_n$ in \rev{each} six-neighboring cell $\gamma_j$.
This evaluation involves a restricted A* search \rev{between zone centers $z_m.\vect{c}$ and $z_n.\vect{c}$} in the free space which is constrained within the region enclosing the two cells (Fig.~\ref{fig:A*}(a)).
If a path is found, an edge connecting $z_m$ and $z_n$ is added to $E_f$, with edge weight defined as path length.
Similarly, the edges $E_u$ in the unknown subgraph $\mathcal{G}_u$ are updated through restricted A* searches between unknown vertices in unknown space (Fig.~\ref{fig:A*}(b)).
Regarding the portal edges $E_{p}$, which typically emerge in partially explored cells, the restricted A* searches constrained in only cell $\gamma_i$ are performed for each pair of free and unknown vertices within it (Fig.~\ref{fig:A*}(c)).
During the above A* searches, the lengths of path segments passing through unknown space are multiplied by a constant penalty factor $a_\text{penal}$ to account for the uncertainty of unknown space.

After the incremental construction, if the connectivity graph $\mathcal{G}$ contains multiple disjoint components, this indicates the presence of inaccessible hollows in the environment, such as the middle-right rectangle shown in Fig.~\ref{fig:space_decom}.
The existence of disjoint components can be identified by applying the CCL algorithm (\rev{Lemma~\ref{lemma:ccl}}) to the connectivity graph. 
In such cases, any isolated subgraphs composed solely of unknown zones are removed and excluded from exploration planning.

The primary expense of connectivity graph construction arises from the A* searches conducted when creating the edges.
Fortunately, since these searches are confined within small local regions, the incremental connectivity graph requires negligible construction time while bringing significant benefits for subsequent coverage path planning and trajectory generation.
Compared to a time-consuming direct search on voxels, the connectivity graph offers an efficient protocol to approximate traversal distances between positions, especially in large-scale environments.

\subsection{Frontier Extraction and Viewpoints Sampling}
\label{subsec:frontier}
\begin{figure}[t]
	\centering
  \subfigtopskip=0pt
	\subfigbottomskip=2pt
	\subfigcapskip=-3pt   
  \subfigure{\includegraphics[width=0.9\columnwidth]{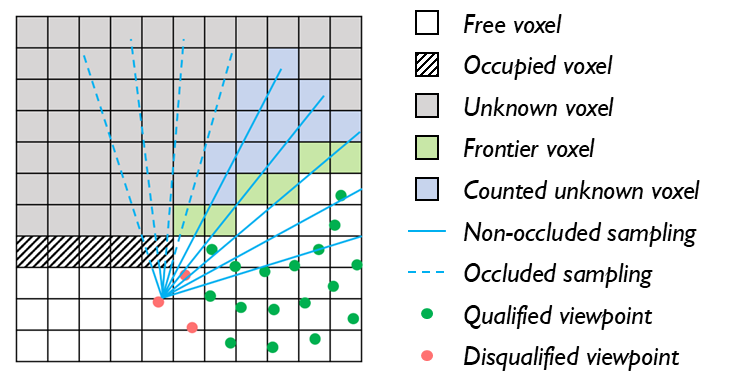}} \\     
  \subfigure{\includegraphics[width=0.9\columnwidth]{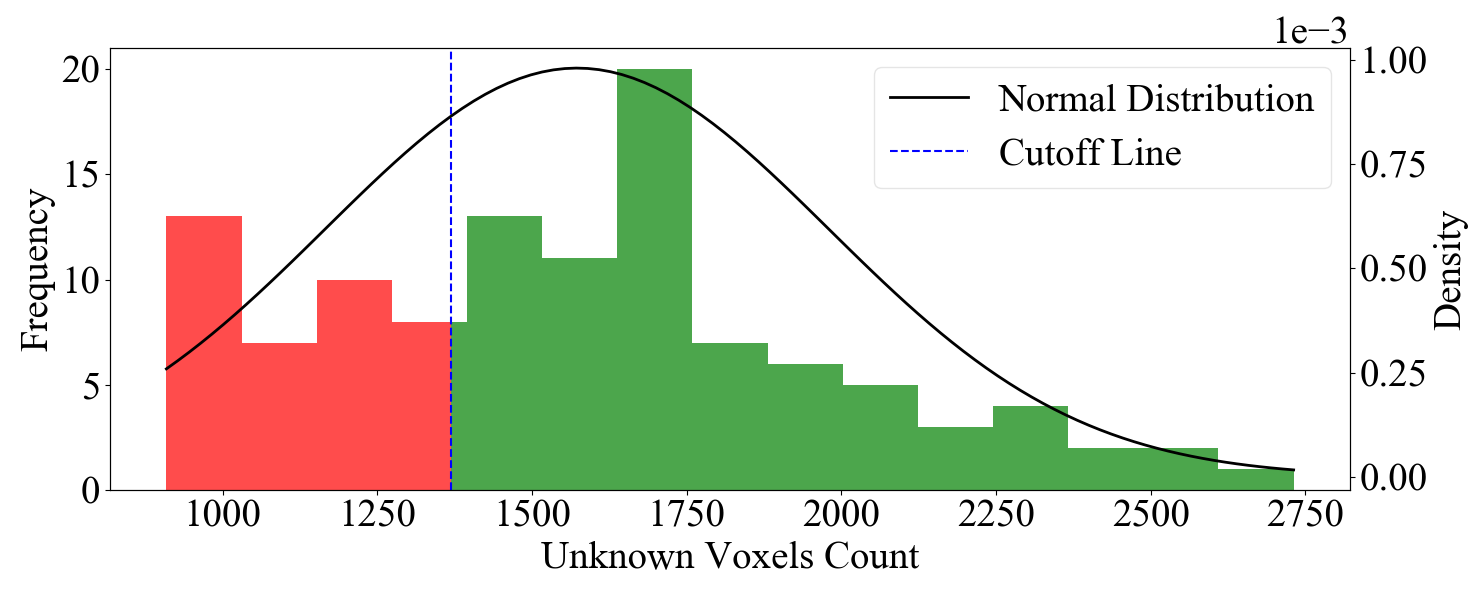}}
  \vspace{-0.4cm}
  \caption{\label{fig:vp_filter} The top image illustrates the viewpoint qualification assessment, where the assessed viewpoint is disqualified due to its observation of an insufficient number of unknown voxels. \rev{The blue lines represent raycast samplings within camera FoV used for counting unknown voxels}. The bottom image shows an example of the histogram and the distribution of the numbers of unknown voxels \rev{at} all viewpoints, where the qualification cutoff line $\mu - z\sigma$ is highlighted in a dotted blue line.}
  \vspace{-0.6cm}
\end{figure}

Frontiers are defined as the boundary regions between known-free and unknown space \cite{yamauchi1997frontier}.
Information regarding frontiers and viewpoints is required to establish targets for later exploration planning.
To begin with, frontier clusters of appropriate size are extracted and viewpoints are uniformly sampled around following \cite{zhou2021fuel}.
Additionally, a qualification assessment is applied to each viewpoint candidate, as depicted in Fig.~\ref{fig:vp_filter}. 
This is accomplished by counting the number of unknown voxels intersected by the truncated rays extending from the viewpoint to a maximum sensor depth at several sampling directions within \rev{the} sensor's FoV. 
Viewpoints holding a number of unknown voxels lower than the cutoff line $\mu - z\sigma$ are disqualified.
Here $\mu$ and $\sigma$ represent the mean and standard deviation of \rev{the number of unknown voxels at all viewpoints}, and the standard score $z$ is a preset constant.
These disqualified viewpoints offer a limited observation of unknown space, providing little benefit for efficient exploration.
The viewpoint covering most frontier voxels among the qualified viewpoints is selected as the \textit{viewpoint representative}.
The viewpoint representative, along with its corresponding frontier, is assigned to the zone in which it is located.
This approach ensures that the viewpoint representative not only covers a significant number of frontier voxels but also an acceptable number of unknown voxels, highlighting its capability to efficiently discover unexplored regions beyond the frontier.


\section{CP-Guided Hierarchical Exploration Planning}
\label{sec:cp_planning}

The hierarchical exploration planning begins with constructing a global coverage path over zones utilizing the connectivity graph.
Instead of greedily selecting the next viewpoint target, our local planner incorporates the intention of \rev{the} coverage path \rev{guidance} more consistently and explores surrounding frontiers in a flexible and globally optimized order minimizing navigation time.

\subsection{Coverage Path Planning}
\label{subsec:cp_construct}
\begin{figure*}[t]
	\centering
  \includegraphics[width= 1.8\columnwidth]{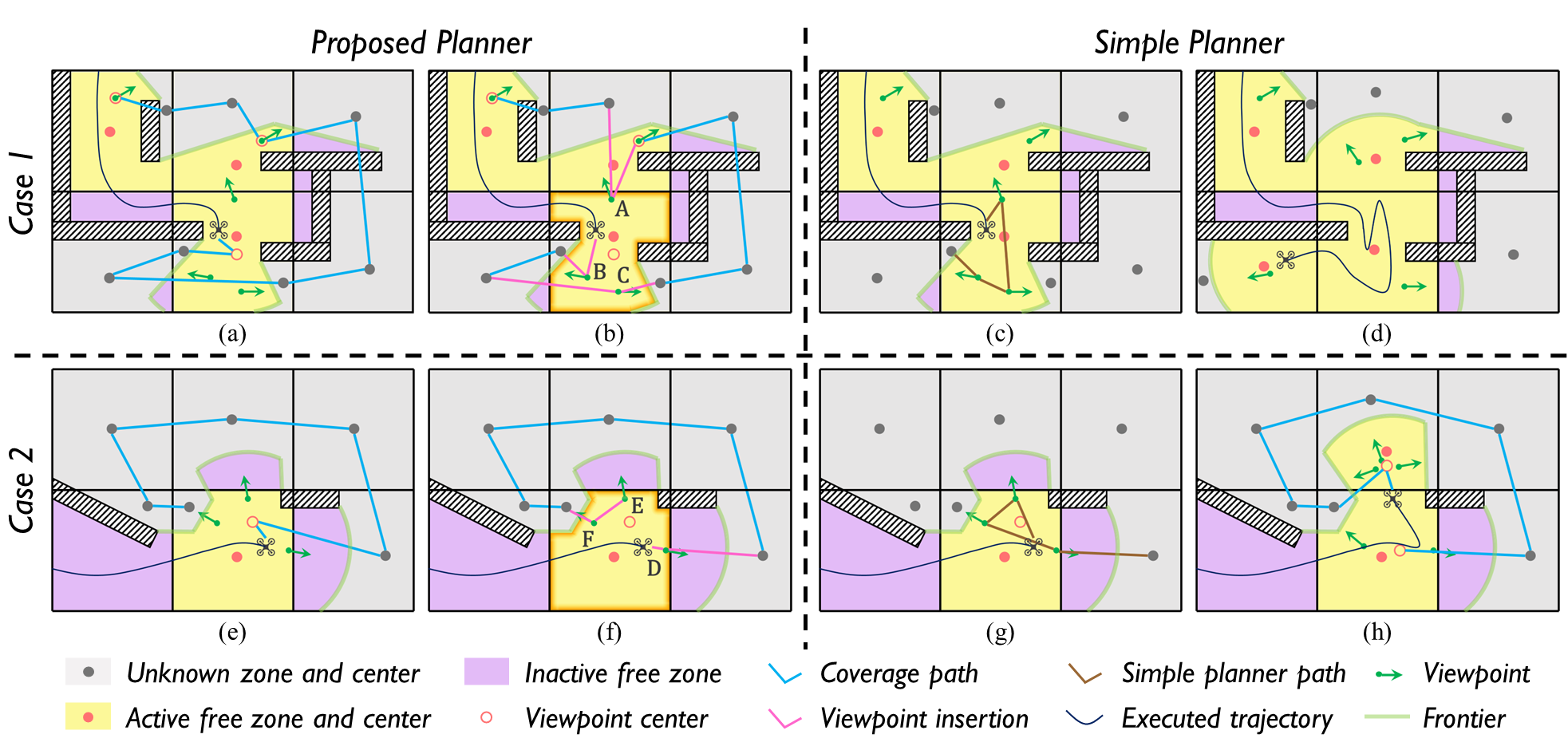}   
  \vspace{-0.4cm}
  \caption{\label{fig:cp_planning_8fig} Two cases are shown in the top and bottom rows respectively. The left two columns illustrate the proposed hierarchical exploration planner. In (a)(e), a coverage path is first generated, shown as \rev{a} blue polyline. During local path planning in (b)(f), the viewpoint representatives belonging to the highlighted zone are collected and inserted into the coverage path, shown as pink segments. The proposed planner generates efficient local paths consistently following the intention of global guidance.
  The right two columns \rev{depict} problems produced by a simple local planner. In (c)(d), it produces a \rev{circuitous} route reducing exploration efficiency. In (g)(h), it takes a detour deviating from intended global \rev{guidance}.}
  \vspace{-0.3cm}
\end{figure*}

We aim for a coverage path that includes all zones awaiting agent inspection with minimal traversal time, as depicted in Fig.~\ref{fig:cp_planning_8fig}(a).
Unlike previous methods considering only unknown space \cite{zhou2023racer,song2020online}, we also account for \textit{active free zones}, which are defined as free zones containing at least one frontier viewpoint representative. 
This better models the exploration process where the agent first reaches frontier viewpoints in active free zones before pushing frontiers into unknown regions.
For active free zones, \textit{viewpoint centers} are computed as the average positions of all viewpoint representatives within the zone and rectified as in Sec.~\ref{subsec:space_decom} if necessary.
Distinguished from the zone center, which is related to zone geometry, the viewpoint center considers frontiers distribution and represents a more accurate position that the agent needs to arrive when visiting the zone.

Coverage path planning is formulated as an \rev{Asymmetric} Traveling Salesman Problem (ATSP) inspired by \cite{zhou2021fuel}.
We solve for an open-loop tour of \textit{target positions}, starting from the current position $\vect{p}_c$ and passing through viewpoint centers of all active free zones as well as zone centers of all unknown zones.
To ensure the solution \rev{minimizes} the actual exploration duration, we dedicate effort to constructing a cost matrix that evaluates the traversal time between target positions more accurately. 
This is achieved by employing a more precise traversal time model, without resorting to \rev{empirical} heuristic cost designs.

To compute the cost matrix $\matr{C}_\text{cp}$ for the ATSP, a path $\mathcal{P}$ is first searched for each pair of target positions $(\vect{p}_0, \vect{p}_n)$ using a hybrid method.
For a short-distance search when $||\vect{p}_0-\vect{p}_n||_2 < d_\text{thr}$, the path is computed with a voxel-based A* search that treats unknown voxels as free but applies a multiplicative penalty $a_\text{penal}$ to account for their uncertain status.
For a long-distance search otherwise, the A* search is performed on the connectivity graph between the two graph vertices nearest to $\vect{p}_0$ and $\vect{p}_n$.
While this only provides a rough path, achieving high accuracy is unnecessary over a long distance. 
Together, this hybrid approach efficiently solves both short and long-distance path searching, enabling rapid cost matrix computation.

Given the searched path $\mathcal{P} = \{\vect{p}_0, \vect{p}_1, \cdots, \vect{p}_n\}$ for target positions $\vect{p}_0$ and $\vect{p}_n$, the traversal time for $\mathcal{P}$ is estimated in a segment-wise manner. 
For segment $\vect{p}_i \vect{p}_{i+1}$ with length $l_i$, the approximate traversal time $t_i$ can be computed using a linear uniformly accelerated motion model:
\begin{equation} \label{eq:cp_ti}
  t_i = \dfrac{l_i}{v_m} + \dfrac{(v_m - |\hat{v}_i|)^2}{2v_ma_m} + 2\dfrac{|\hat{v}_i|}{a_m} H(-\hat{v}_i),
\end{equation}
where $v_m$ and $a_m$ are maximum velocity and acceleration. $H(\cdot)$ is the Heaviside step function, which takes a value of zero for negative arguments and one otherwise. 
$\hat{v}_i$ is the projection scalar of the \rev{velocity $\vect{v}_i$ at $\vect{p}_i$} along the segment $\vect{p}_i\vect{p}_{i+1}$,
\begin{equation}
  \hat{v}_i = \frac{\vect{v}_i \cdot (\vect{p}_{i+1} - \vect{p}_i)}{||\vect{p}_{i+1} - \vect{p}_i||}
\end{equation}
where the \rev{velocity $\vect{v}_i$ at $\vect{p}_i$} is defined as $\vect{v}_0 = \vect{v}_1,$
\begin{equation}
  \vect{v}_i = v_m \cdot \dfrac{\vect{p}_i - \vect{p}_{i-1}}{||\vect{p}_i - \vect{p}_{i-1}||}, \quad i \geq 1.
  \label{eq:hat_vi}
\end{equation}

Using \rev{(\ref{eq:cp_ti})-(\ref{eq:hat_vi})}, the traversal time between $\vect{p}_0$ and $\vect{p}_n$ is computed as $\Sigma_{i=0}^{n-1} t_i$. The entries of the cost matrix $\matr{C}_\text{cp}$, which includes traversal times between all pairs of target positions, are computed in this fashion.
The exceptions are the entries corresponding to the cost from zone centers or viewpoint centers to the current position $\vect{p}_c$, which are assigned zero.
This means returning to $\vect{p}_c$ incurs no cost.
Consequently, every closed-loop tour always includes an open-loop solution that shares the same cost value and starts from $\vect{p}_c$, which satisfies our objectives.
This cost evaluation approach accounts for quadrotor kinematics, resulting in higher costs being assigned to longer and more tortuous paths.
As a result, the obtained cost matrix provides more accurate estimations of traversal time that better reflect the actual executions \rev{in comparison with previous empirical heuristic cost designs}.

\vspace{-0.4cm}
\subsection{CP-Guided Local Path Planning}
\label{subsec:CP_guided_planning}
The coverage path $\bar{\Lambda} = \{\vect{p}_c, \bar{z}_1, \bar{z}_2, \cdots, \bar{z}_n\}$ produced above offers a promising order of zone visitation\rev{, where $\bar{z}_i$ are unknown and active free zones considered during coverage path planning}.
Nonetheless, the agent still requires an effective order for visiting frontiers during exploration.
\rev{A \textit{simple planner}} might extract frontiers within the first zone along the coverage path and compute a shortest path visiting these frontiers locally, as shown in Fig. 6(c) and (g).
However, this locally optimized strategy may result in \rev{circuitous} routes with sharp turns, reducing exploration efficiency (Fig.~\ref{fig:cp_planning_8fig}(d)). 
More importantly, it can lead to suboptimal detours that deviate from the intention of global guidance, making the global guidance less meaningful (Fig.~\ref{fig:cp_planning_8fig}(h)).
In contrast, our planner produces local paths that consistently \rev{follow} the guidance of the global coverage path, while allowing the possibility of intentionally leaving certain frontier viewpoints to be dropped by later, as illustrated in Fig.~\ref{fig:cp_planning_8fig}(b)(f).

Specifically, the frontier viewpoint representatives held by zone $\bar{z}_1$ and zone $\bar{z}_c$ are collected into a set $\mathcal{V}$, where $\bar{z}_c$ denotes the zone $\vect{p}_c$ belongs to.
Note that $\bar{z}_c$ and $\bar{z}_1$ may refer to the same zone, which happens when $\vect{p}_c$ is located in zone $\bar{z}_1$.
As illustrated in Fig.~\ref{fig:cp_planning_8fig}(b)(f), these viewpoint representatives in $\mathcal{V}$ are then inserted into the reduced coverage path ordered sequence
$\Lambda = \bar{\Lambda} \backslash \{\bar{z}_1, \bar{z}_c\} = \{\vect{p}_c, z_1, \cdots, z_{n^\prime}\}$
with the objective of minimizing traversal time.
This can be formulated as a Sequential Ordering Problem (SOP) \cite{hernadvolgyi2004solving}, which is a variation of ATSP with precedence constraints.

\begin{lemma} [\bfseries Sequential Ordering Problem] \label{lemma:sop}
  Given a directed graph $G=(V,E)$, edge weight $w : E \to \mathbb{R}$ and a set of precedence constraints $P \subseteq V \times V$, the Sequential Ordering Problem \rev{searches} for a minimum cost permutation of vertices from $s \in V$ to $t \in V$ that satisfies the precedence constraints.
\end{lemma}

In our case, we define the graph vertices as $V_\text{sop} = \Lambda \cup \mathcal{V}$ and the precedence constraints set as
\begin{equation}
    P_\text{sop} = \{(\vect{p}_c, x_i) | x_i \in V_\text{sop}\backslash\{\vect{p}_c\}\} \cup \{ (z_i, z_j) | z_i, z_j \in \Lambda, i < j\},
  \end{equation}
where precedence constraint $(a,b)$ indicates vertex $a$ must be visited before vertex $b$.
The cost matrix defining the problem, which again we wish to calculate accurately, can be expressed as
\begin{equation}
  \matr{C}_\text{sop} = 
  \begin{bmatrix}
    \matr{C}_{z,z} & \matr{C}_{z,v} \\
    \matr{C}_{v,z} & \matr{C}_{v,v}
  \end{bmatrix}.
\end{equation}

$\matr{C}_{z,z}$ is an \rev{asymmetric} matrix which represents costs between reduced coverage path elements in $\Lambda$.
Its entry values can be filled by reusing the information from cost matrix $\matr{C}_\text{cp}$ described in Sec.~\ref{subsec:cp_construct} without extra computational overhead:
\begin{equation}
  \matr{C}_{z,z}(i,j) = 
  \begin{cases}
    \matr{C}_\text{cp}(i^\prime,j^\prime) & \text{if $i < j$}, \\
    -1 & \text{if $i > j$}, \\
    0 & \text{otherwise},
  \end{cases}
\end{equation}
where $\matr{C}_\text{cp}(i^\prime,j^\prime)$ is the corresponding entry for elements $i,j$, and the $-1$ represents the precedence constraints in $P_\text{sop}$.

$\matr{C}_{v,v}$ is a \rev{symmetric} matrix that records the cost values between frontier viewpoint representatives in $\mathcal{V}$.
The entry for viewpoint $v_i, v_j$ with yaw angles $y_i, y_j$ are calculated as
\begin{equation} \label{eq:c_vv}
  \matr{C}_{v,v}(i,j) = \matr{C}_{v,v}(j,i) = \max\{t_{pos}, t_{yaw}\},
\end{equation}
where the positional time cost $t_{pos} = \sum t_i$ is computed using \rev{(\ref{eq:cp_ti})} and the time $t_{yaw}$ required for yaw rotation between two viewpoints is similarly estimated using a linear uniformly accelerated motion model.

Similarly, the entries of $\matr{C}_{v,z}$ and $\matr{C}_{z, v}$, which store costs between viewpoint representatives and reduced coverage path elements, can be calculated by \rev{(\ref{eq:c_vv})}.
Note that to compute $t_{yaw}$, the heading directions of reduced coverage path elements $z_i$ are artificially defined as \rev{$\vect{y}(z_i) = z_i.\vect{c} - z_{i-1}.\vect{c}$} with \rev{$z_0.\vect{c} \vcentcolon = \vect{p}_c$}.
To ensure the current position $\vect{p}_c$ is always at the beginning of the path, the values in the first column of matrix $\matr{C}_{v,z}$ are set to $-1$ as precedence constraints.

Fig.~\ref{fig:cp_planning_8fig}(b)(f) exemplify local planning results for frontier visitation order in two scenarios.
In Fig.~\ref{fig:cp_planning_8fig}(b), viewpoint representative A is intentionally postponed for later visitation, and viewpoint representative C is planned to be visited after the agent exits the bottom-left zone. 
In Fig.~\ref{fig:cp_planning_8fig}(f), viewpoint \rev{representatives} E and F are arranged to be visited in the last.
These decisions produce an exploration motion that is better aligned with the global coverage path. 
This flexibility enables efficient exploration of the surrounding frontiers in \rev{a} globally optimized manner while maintaining consistency with the intention of \rev{the} coverage path.

\subsection{Local Refinement and Trajectory Generation}
\label{subsec:traj_gen}
Once the frontier visitation order is determined, a local viewpoint refinement is performed to obtain the optimal combination among all viewpoints similar to \cite{zhou2021fuel}, since only viewpoint \rev{representatives} were considered in the previous stage.
Given an ordered list of viewpoint targets, navigation paths are searched within free space using a hybrid method.
Voxel-based A* search is employed for nearby targets, whereas graph A* search is applied on the free subgraph $\mathcal{G}_f$ of the connectivity graph for distant targets.
Based on the path segments, a minimum-time B-spline trajectory is then generated using \cite{zhou2019robust}, ensuring constraints on the smoothness, safety and quadrotor dynamics.

\section{Exploration Planner Evaluation Environment}
\label{sec:epee}

In this section, we introduce the exploration planner evaluation environment that facilitates fair and comprehensive simulation experiments in the subsequent benchmark experiments (Sec.~\ref{sec:benchmark}) and ablation studies (Sec.~\ref{sec:ablation}).
Most existing works often validate the exploration planners on a narrow selection of specific maps, typically inherited from previous studies.
However, this can result in algorithms overly specialized for those particular maps.
Therefore, we develop a lightweight software-in-the-loop \textit{exploration planner evaluation environment} for fair and comprehensive simulation evaluation of fast autonomous exploration planners using \rev{an} aerial robot.
This environment allows for \rev{the} comparison and assessment of exploration planners across a variety of testing scenarios with different characteristics using an identical quadrotor simulator.

\subsection{Exploration Planner Evaluation Environment}
\label{subsec:exp_env}
\begin{figure}[t]
	\centering
  \includegraphics[width=0.95\columnwidth]{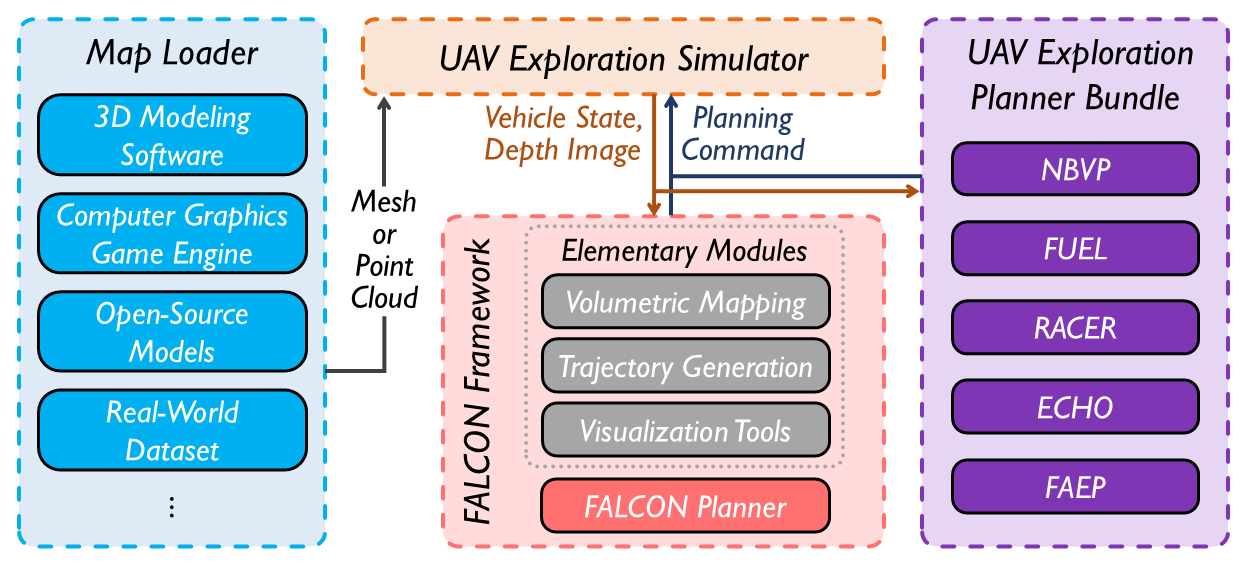}  
  \vspace{-0.4cm}
  \caption{\label{fig:epee_system} The system setup of the exploration planner evaluation environment developed for autonomous exploration with aerial robot.}
  \vspace{-0.2cm}
\end{figure}

\rev{The simulation environment was developed based on the ROS Noetic\footnote{https://wiki.ros.org/noetic} framework and the depth information is directly rendered by simulation using Open3D\footnote{https://www.open3d.org/} without artificial noise.}
The system setup of the exploration planner evaluation environment for the task of autonomous exploration with aerial robot is depicted in Fig.~\ref{fig:epee_system}.
It consists of the following key components:
\begin{enumerate}
  \item A map loader that supports map inputs generated from various sources, such as 3D modeling \rev{software}, game engines, open-source models\rev{,} and real-world datasets;
  \item An aerial robot simulator that models quadrotor dynamics and visual sensor measurements.
  \item Several elementary modules for high-level exploration algorithms, which include volumetric mapping\cite{zhang2022exploration}, collision-free trajectory generation\cite{zhou2020raptor}\rev{,} and visualization tools;
  \item A plug-and-play bundle of state-of-the-art exploration planner representatives \cite{bircher2016receding, zhou2021fuel, zhou2023racer, yu2023echo, zhao2023autonomous} for testing and evaluation.
\end{enumerate}

\begin{figure}[t]
	\centering
  \subfigtopskip=0pt
	\subfigbottomskip=2pt
	\subfigcapskip=-3pt
  \subfigure{\includegraphics[width=0.3\columnwidth]{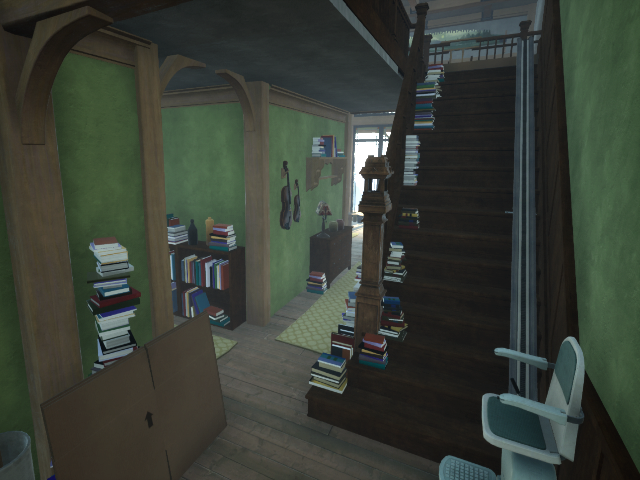}} \hskip -3pt      
  \subfigure{\includegraphics[width=0.3\columnwidth]{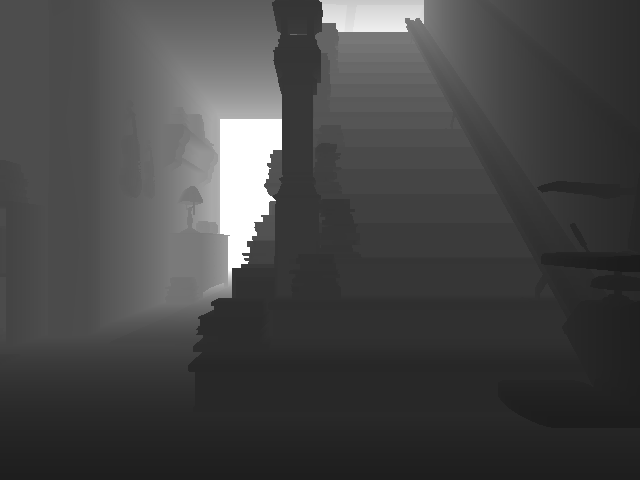}} \hskip -3pt    
  \subfigure{\includegraphics[width=0.3\columnwidth]{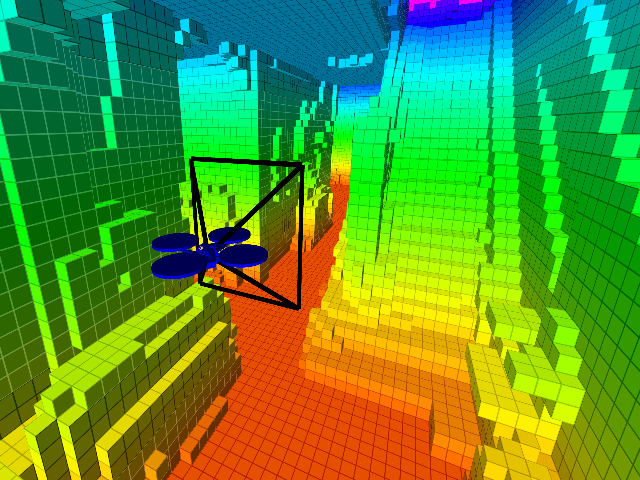}} \hskip -3pt \\ \vskip -1.5pt
  \subfigure{\includegraphics[width=0.3\columnwidth]{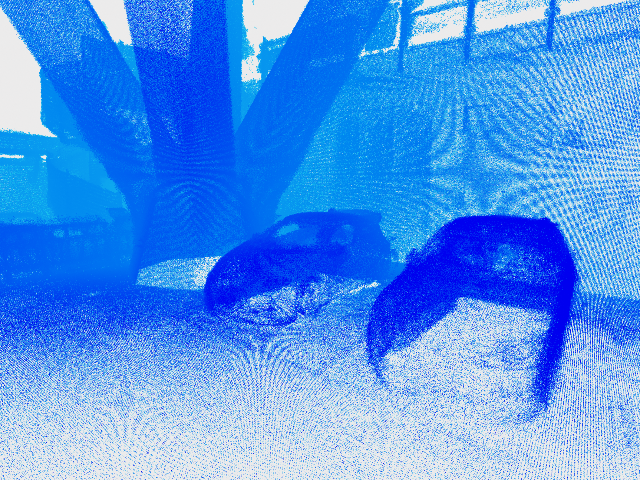}} \hskip -3pt      
  \subfigure{\includegraphics[width=0.3\columnwidth]{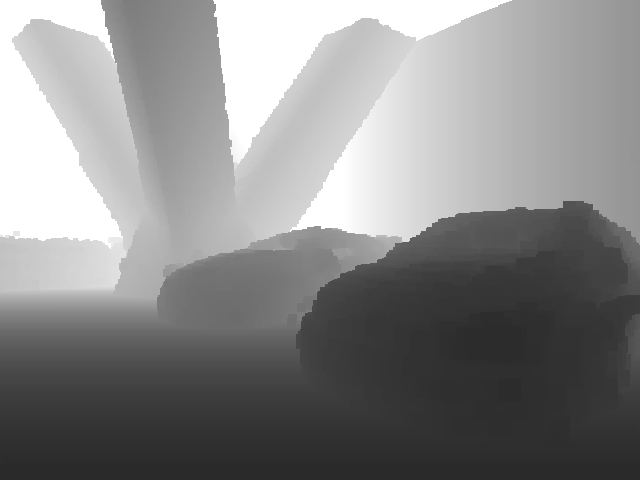}} \hskip -3pt    
  \subfigure{\includegraphics[width=0.3\columnwidth]{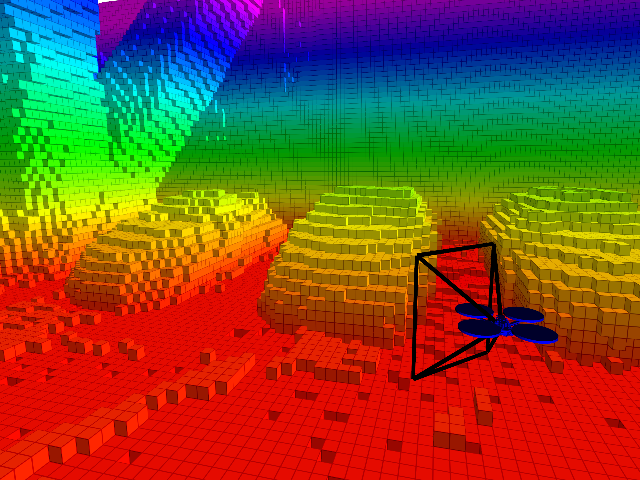}} \hskip -3pt 
  \vspace{-0.2cm}
  \caption{\label{fig:simulator_render} The first column displays snapshots of the original scenarios and the second column shows the real-time depth images rendered by the simulator. The third column showcases a quadrotor conducting an exploration task in the simulator, along with the volumetric maps constructed on the fly.}
  \vspace{-1.4cm}
\end{figure}

Different from a recent autonomous exploration development environment \cite{cao2022autonomous} for LiDAR-based exploration with \rev{a} ground vehicle, our environment is designed for aerial robot exploration with visual sensors.
More importantly, our environment not only provides a diverse testing map dataset but also supports customization of testing scenarios generated from various sources.
This includes 3D modeling \rev{software} (\rev{e.g.,} SolidWorks$\footnote{\url{https://www.solidworks.com/}}$, OpenSCAD$\footnote{\url{https://openscad.org/}}$), 3D computer graphics game engines (\rev{e.g.,} Unreal Engine$\footnote{\url{https://www.unrealengine.com/}}$, Unity$\footnote{\url{https://unity.com/}}$), as well as open-source models and real-world datasets (\rev{e.g.,} Gazebo models$\footnote{\url{https://app.gazebosim.org/}}$, FusionPortable\cite{jiao2022fusionportable}).
Given the map input and planning command, the simulator is able to simulate quadrotor dynamics and render real-time depth images as sensor measurements.
This simulator builds upon the one utilized in several previous works \cite{zhou2020raptor,zhou2021fuel,zhang2022exploration,zhou2023racer}, which was limited to accepting only point cloud inputs.
In our improved version, we have introduced the capability of importing a map in either mesh or point cloud representation. 
As demonstrated in Fig.~\ref{fig:simulator_render}, the top row shows a mesh map \textit{Edith Finch}$\footnote{\url{https://www.unrealengine.com/marketplace/en-US/product/ef-house}}$ provided by the Unreal Engine and the bottom row shows a point cloud map \textit{complex parking garage} released by MARSIM\cite{kong2023marsim}.
This evaluation environment facilitates comprehensive comparisons of \rev{existing} exploration planners across diverse testing maps.
As an example, we incorporate several state-of-the-art exploration planners \cite{bircher2016receding, zhou2021fuel, zhou2023racer, yu2023echo, zhao2023autonomous} and the proposed planner into this environment, offering a plug-and-play benchmark bundle for further research.
It also serves as a mature platform that benefits the research community in developing high-level autonomous exploration algorithms for aerial robots.

\subsection{Evaluation Scenarios}
We provide diverse testing scenarios for comprehensive simulation experiments, where the scenario diversity is measured by \textit{accessibility} and \textit{complexity}. 
Accessibility is defined as the volumetric percentage of accessible regions of the entire exploration space within the bounding box. 

\begin{table}[t]
  \caption{Characteristics of Testing Scenarios}
  \begin{tabular}{M{0.9cm}M{0.9cm}M{0.9cm}M{0.9cm}M{1.15cm}M{1.55cm}}
    \toprule\toprule
    \textbf{Scene} & \begin{tabular}[c]{@{}c@{}}\textbf{Access-}\\\textbf{ibility}\end{tabular} & \begin{tabular}[c]{@{}c@{}}\textbf{Complex}\\\textbf{-ity}\end{tabular} & \begin{tabular}[c]{@{}c@{}}\textbf{Dimen-}\\\textbf{sion}\end{tabular} & \textbf{\rev{Size (m)}} & \textbf{Source} \\
    \midrule
    \begin{tabular}[c]{@{}c@{}}Classical\\Office\end{tabular} & \begin{tabular}[c]{@{}c@{}}High\\99.55\%\end{tabular} & \begin{tabular}[c]{@{}c@{}}Low\\0.078\end{tabular} & 2.5D$^\dagger$ & 30x15x2 & \begin{tabular}[c]{@{}c@{}}Pointcloud\\Generator\cite{zhou2021fuel}\end{tabular} \\ [0.4cm]
    \begin{tabular}[c]{@{}c@{}}Complex\\Office\end{tabular} & \begin{tabular}[c]{@{}c@{}}High\\92.42\%\end{tabular} & \begin{tabular}[c]{@{}c@{}}High\\0.656\end{tabular} & 2.5D$^\dagger$ & 30x30x2 & SolidWorks \\ [0.4cm]
    \begin{tabular}[c]{@{}c@{}}Octa\\Maze\end{tabular} & \begin{tabular}[c]{@{}c@{}}Mid\\86.61\%\end{tabular}   & \begin{tabular}[c]{@{}c@{}}Low\\0.061\end{tabular} & 2.5D$^\dagger$ & 35x35x2 & OpenSCAD \\ [0.4cm]
    \begin{tabular}[c]{@{}c@{}}DARPA\\Tunnel\end{tabular} & \begin{tabular}[c]{@{}c@{}}Low\\28.19\%\end{tabular} & \begin{tabular}[c]{@{}c@{}}Low\\0.024\end{tabular} & 2.5D & 42x20x2 & \begin{tabular}[c]{@{}c@{}}Real-World\\Dataset\tablefootnote{\url{https://github.com/subtchallenge/systems_tunnel_ground_truth}}\end{tabular} \\ [0.4cm]
    \begin{tabular}[c]{@{}c@{}}Duplex\\Office\end{tabular} & \begin{tabular}[c]{@{}c@{}}Mid\\89.30\%\end{tabular} & \begin{tabular}[c]{@{}c@{}}High\\0.791\end{tabular} & 3D & 20x20x4 & SolidWorks \\ [0.4cm]
    \begin{tabular}[c]{@{}c@{}}Power\\Plant\end{tabular} & \begin{tabular}[c]{@{}c@{}}Mid\\72.60\%\end{tabular} & \begin{tabular}[c]{@{}c@{}}Low\\0.006\end{tabular} & 3D & 30x15x15 & \begin{tabular}[c]{@{}c@{}}Gazebo\\Models\end{tabular} \\
    \toprule\toprule
  \end{tabular} \\
  \footnotesize{$^\dagger$ 2.5D maps are an upgrade of 2D maps by incorporating height information. We \rev{consider} a map as 2.5D if it can be visually displayed on a 2D plane while accurately expressing the environment.}\\
  \label{tab:exp_scenarios}
  \vspace{-0.4cm}
\end{table}

\begin{figure}[t]
  \centering
  \subfigtopskip=0pt
  \subfigbottomskip=1pt
  \subfigure[Classical Office]{\includegraphics[width=0.3\columnwidth]{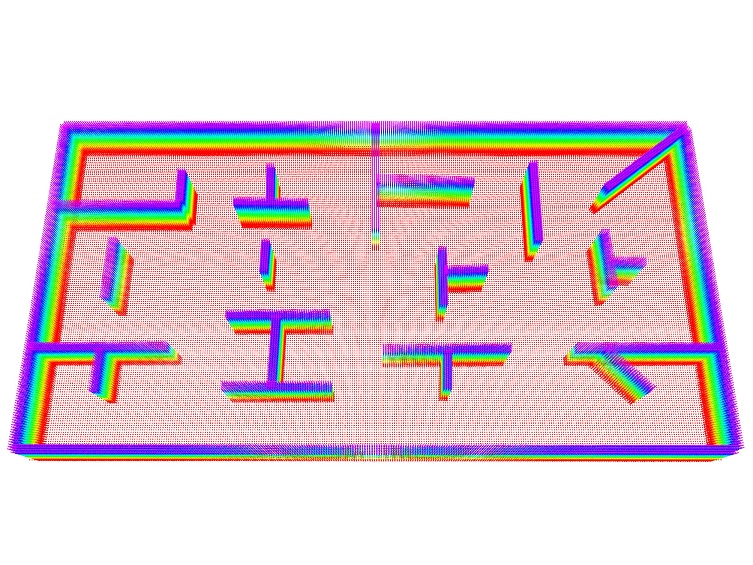}} \hskip -2pt  
  \subfigure[Complex Office]{\includegraphics[width=0.3\columnwidth]{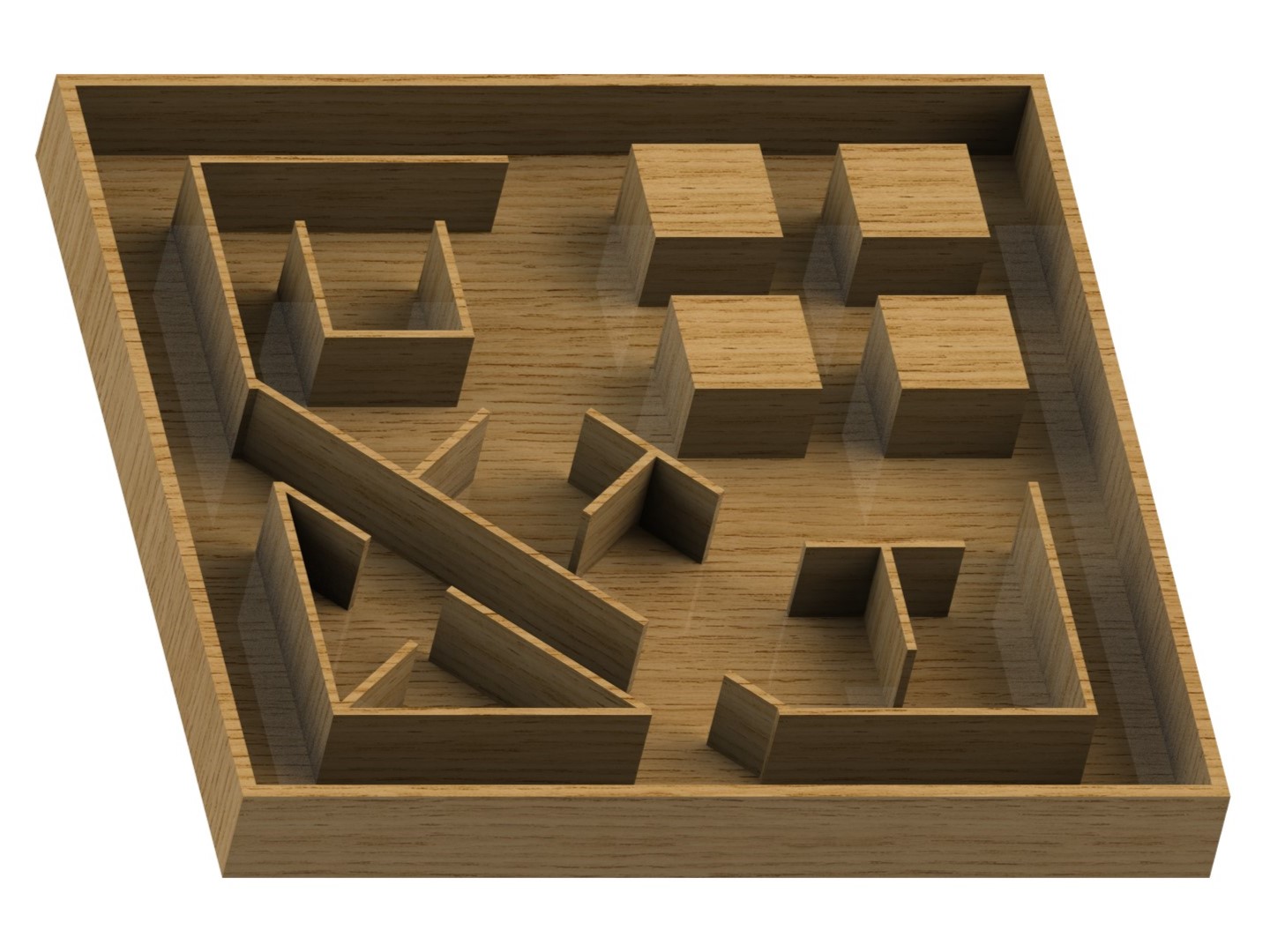}} \hskip -2pt      
  \subfigure[Octa Maze]{\includegraphics[width=0.3\columnwidth]{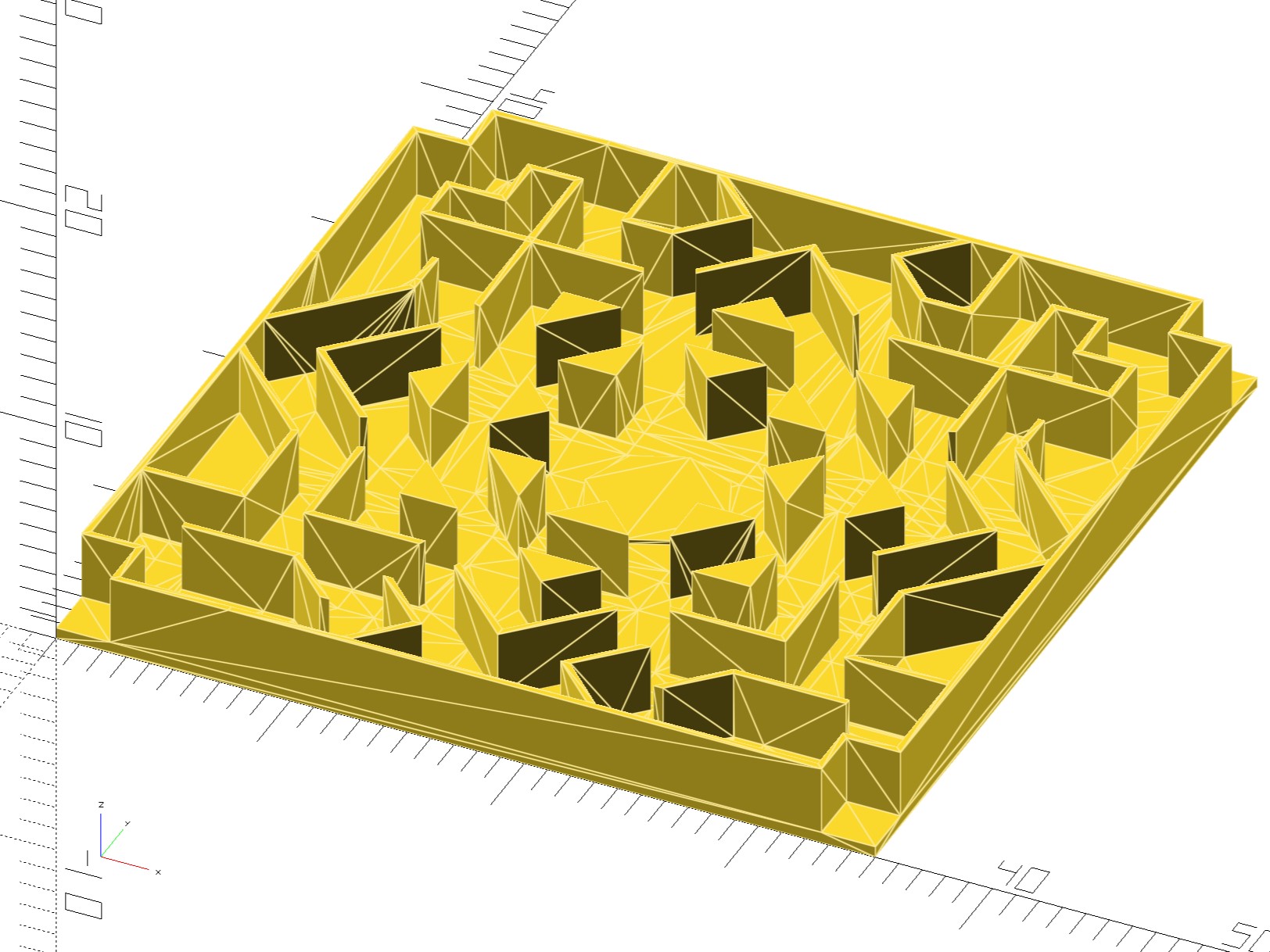}} \hskip -2pt \\
  \subfigure[DARPA Tunnel]{\includegraphics[width=0.3\columnwidth]{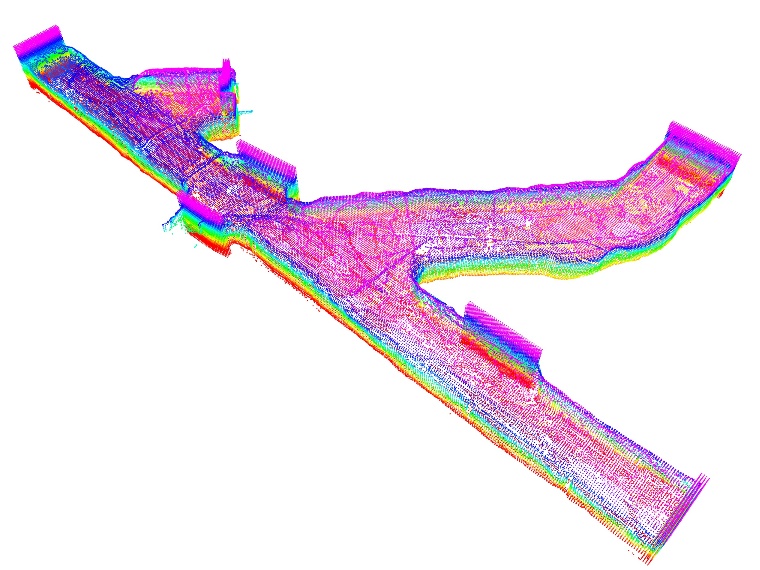}} \hskip -2pt      
  \subfigure[Duplex Ofﬁce]{\includegraphics[width=0.3\columnwidth]{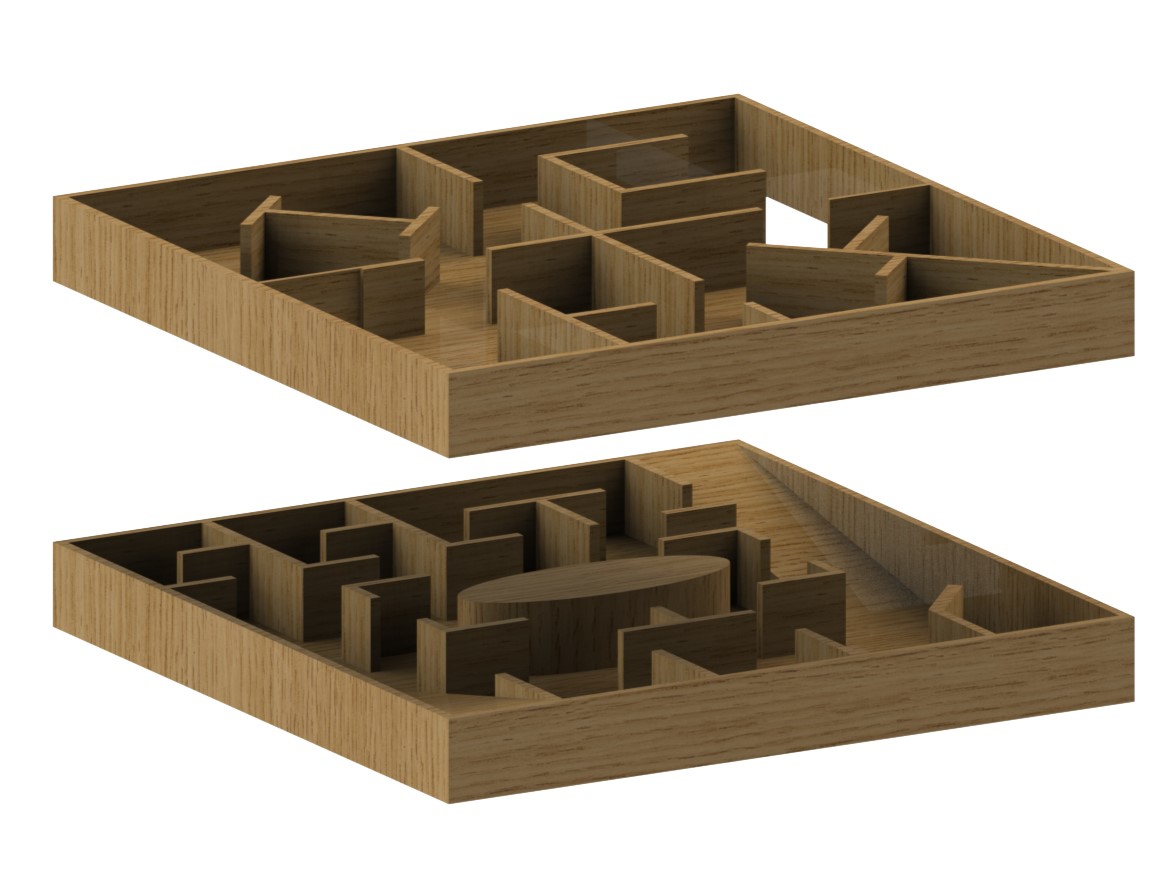}} \hskip -2pt
  \subfigure[Power Plant]{\includegraphics[width=0.3\columnwidth]{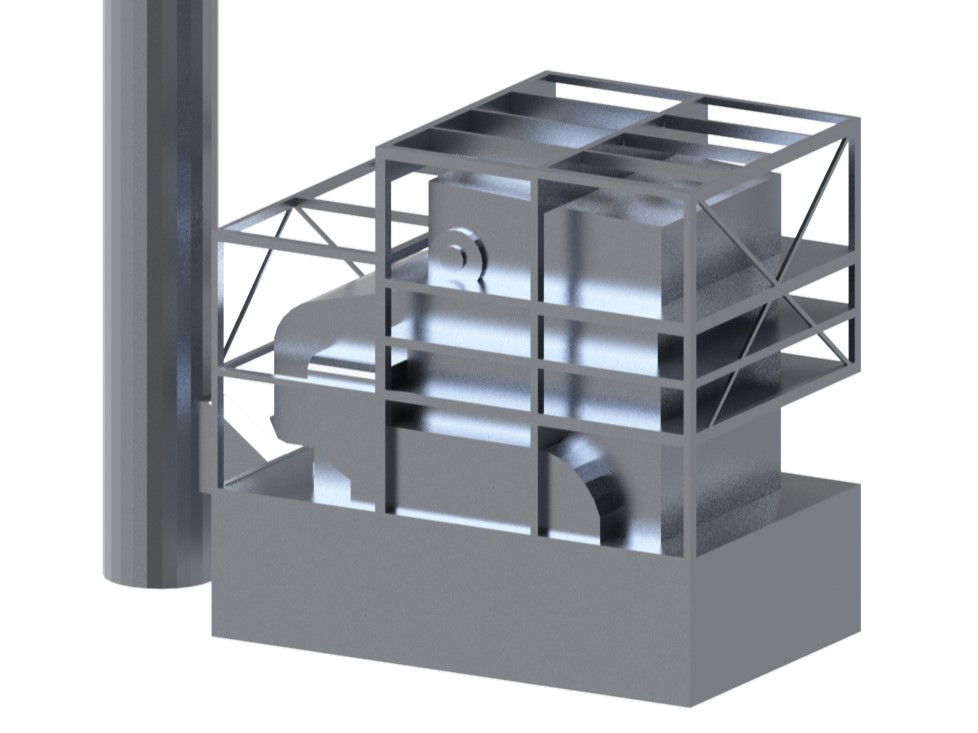}}
  \caption{\label{fig:exp_scenarios} An overview of the testing scenarios used for simulation experiments.}
  \vspace{-1.2cm}
\end{figure}

\begin{equation}
    \text{Accessibility} = \frac{V_\text{acc}}{V}.
\end{equation}
Complexity computation begins by sampling $n$ pairs of points, where the Euclidean distances between the pairs are uniformly distributed. 
For each pair of points, the ratio of the shortest free path length to the Euclidean distance is computed.
Complexity is quantified by the variance-to-mean ratio (VMR) of these $n$ ratios, which reflects the dispersion of their distribution.
A more dispersed distribution indicates a more unstructured arrangement of obstacles in the scenario and higher complexity.

\begin{equation}
  \begin{aligned}
    \text{Complexity} &= \frac{\sigma^2}{\mu} = \frac{\sum_{i=1}^n (r_i - \frac{1}{n}\sum_{i=1}^n r_i)^2}{\sum_{i=1}^n r_i} \\
    r_i &= \frac{L(p_i, q_i)}{||p_i-q_i||_2}, i = 1, 2, \cdots, n
  \end{aligned}
\end{equation}

For subsequent simulation experiments, we provide six scenarios with different levels of accessibility and complexity.
These scenarios include the \textit{Classical Office}, \textit{Complex Office}, \textit{Octa Maze}, \textit{DARPA Tunnel}, \textit{Duplex Office}, \textit{Power Plant}.
The characteristics of these \rev{scenarios} are summarized in \rev{Table~\ref{tab:exp_scenarios}} and overviews are depicted in Fig.~\ref{fig:exp_scenarios}.

\section{Benchmark and Analysis}
\label{sec:benchmark}
In this section, we conduct extensive benchmark experiments in simulation to evaluate the proposed exploration planner \textbf{FALCON} comparing with state-of-the-art methods \cite{bircher2016receding, zhou2021fuel, zhou2023racer, yu2023echo, zhao2023autonomous}.
By analyzing the experimental results, we provide an in-depth evaluation of the relative strengths and limitations exhibited by these exploration planners according to the \textit{VECO} criteria.
All simulation experiments are conducted on a computer with an Intel Core i7-13700F, GeForce RTX \rev{4090 24G,} and 32GB memory.

\begin{table}[t]
  \caption{Parameters Setting for Exploration Planners}
  \begin{tabular}{M{1.2cm}p{2.1cm}p{1.0cm}M{0.8cm}M{1.8cm}}
    \toprule\toprule
    \textbf{Type} & \textbf{Parameter} & \textbf{Section} & \textbf{Notation} & \textbf{Value} \\
    \midrule
    \multirow{4}{*}{\begin{tabular}[c]{@{}c@{}}FALCON\\Params\end{tabular}} & Safety clearance & \ref{subsec:space_decom} & $d_\text{min}$ & $0.7$ m \\
                                                                            & \rev{Uniform cell size} & \rev{\ref{subsec:space_decom}} & \rev{$s_\text{cell}$} & \rev{$5.0$ m} \\
                                                                            & Unknown penalty & \ref{subsec:connectivity_graph},\ref{subsec:cp_construct} & $a_\text{penal}$ & $1.5$ \\
                                                                            & Standard score & \ref{subsec:frontier} & $z$ & $0.1$ \\
                                                                            & Distance threshold & \ref{subsec:cp_construct},\ref{subsec:traj_gen} & $d_\text{thr}$ & $10.0$ m\\ \cmidrule{1-5}
    \multirow{6}{*}{\begin{tabular}[c]{@{}c@{}}Benchmark\\Common\\Params\end{tabular}} & \multicolumn{2}{l}{Max linear velocity} & $v_m$ & $2.0$ m$\slash$s\\
                                                                            & \multicolumn{2}{l}{Max linear acceleration} & $a_m$ & $3.0$ m$\slash$s$^2$\\
                                                                            & \multicolumn{2}{l}{Max angular velocity} & $\dot{\xi}_m$ & $1.57$ rad$\slash$s\\
                                                                            & \multicolumn{2}{l}{Max angular acceleration} & $\ddot{\xi}_m$ & $1.57$ rad$\slash$s$^2$ \\
                                                                            & \multicolumn{2}{l}{Sensor FoV} & - & $[80\times60]$ deg \\
                                                                            & \multicolumn{2}{l}{Sensing depth} & - & $5.0$ m \\
    \toprule\toprule
  \end{tabular}
  \label{tab:params}
  \vspace{-2.6cm}
\end{table}

\subsection{Implementation Details}
\label{subsec:implement_detail}
\subsubsection{\textbf{FALCON} Configurations}
For the algorithms, \rev{the} ATSP mentioned in Sec.~\ref{subsec:cp_construct} is solved using the LKH solver \cite{helsgaun2000effective}. 
The solver of \rev{the} CCL problem in \rev{Lemma~\ref{lemma:ccl}} is self-implemented according to \cite{abubaker2007one}.
An open-source solver$\footnote{\url{https://github.com/rod409/SOP}}$ is employed for the SOP problem in \rev{Lemma~\ref{lemma:sop}}.
The voxel map update module follows \cite{zhang2022exploration}.
The parameter configuration of the proposed planner is listed in \rev{Table~\ref{tab:params}}, which is used in all the experiments in Sec.~\ref{sec:benchmark}~-~\ref{sec:real_exp}.
Exploration replanning is triggered by multiple events, including update to frontiers, detection of unsafe trajectories, and expiration of a predefined time interval set to $3$s. 
The exploration planner operates on a single core for fair benchmark comparisons. 
However, various planner modules can be easily parallelized, such as the calculation of the coverage path cost matrix $\matr{C}_\text{cp}$.

\begin{table*}[t]
  \centering
  \caption{Exploration Statistics}
  \vspace{-0.2cm}
    \begin{tabular}{M{0.8cm}M{1.5cm}M{0.5cm}M{0.5cm}M{0.5cm}M{0.5cm}M{0.5cm}M{0.5cm}M{0.5cm}M{0.5cm}M{0.5cm}M{0.5cm}M{0.5cm}M{0.5cm}M{0.5cm}M{0.5cm}M{0.5cm}M{0.5cm}}
    \toprule\toprule
    \multirow{2}{*}{\textbf{Scene}} & \multirow{2}{*}{\textbf{Method}} & \multicolumn{4}{c}{\textbf{Exploration Time (s)}}         
                                                                        & \multicolumn{4}{c}{\textbf{Flight Distance (m)}}          
                                                                        & \multicolumn{4}{c}{\textbf{Coverage (m$^3$)}}          
                                                                        & \multicolumn{4}{c}{\textbf{Avg Velocity (m/s$^2$)}} \\
                                    &                                  & \textbf{Avg} & \textbf{Std} & \textbf{Max} & \textbf{Min}
                                                                        & \textbf{Avg} & \textbf{Std} & \textbf{Max} & \textbf{Min}
                                                                        & \textbf{Avg} & \textbf{Std} & \textbf{Max} & \textbf{Min}
                                                                        & \textbf{Avg} & \textbf{Std} & \textbf{Max} & \textbf{Min} \\
    \midrule
    \multirow{6}{*}{\begin{tabular}[c]{@{}c@{}}Classical\\Office\end{tabular}} 
      & NBVP\cite{bircher2016receding}   & 741.6 & 83.12 & 839.4 & 609.7 & 338.9 & 31.40 & 373.6 & 295.8 & 757.2 & 28.55 & 795.6 & 719.1 & 0.46 & 0.02 & 0.49 & 0.44 \\
      & FUEL\cite{zhou2021fuel}          & 128.2 & 4.03 & 133.6 & 122.1 & 207.4 & 6.80 & 216.9 & 195.6 & 863.0 & 1.13 & 864.6 & 860.6 & 1.62 & 0.05 & 1.68 & 1.51 \\
      & RACER\cite{zhou2023racer}        & 114.0 & 5.80 & 120.3 & 101.9 & 176.4 & 10.95 & 192.5 & 149.3 & 880.9 & 1.44 & 882.7 & 878.1 & 1.55 & 0.04 & 1.62 & 1.46 \\
      & ECHO\cite{yu2023echo}            & 126.2 & 6.01 & 139.3 & 119.5 & 195.3 & 10.07 & 210.6 & 178.4 & 879.6 & 1.20 & 881.6 & 877.8 & 1.55 & 0.05 & 1.61 & 1.46 \\
      & FAEP\cite{zhao2023autonomous}    & 115.1 & 4.96 & 126.4 & 108.0 & 167.4 & 14.34 & 193.9 & 152.6 & 881.0 & 1.89 & 884.0 & 878.6 & 1.40 & 0.08 & 1.52 & 1.28 \\
      & Proposed                         & \textbf{98.0} & 4.66 & 104.0 & 91.1 & \textbf{159.9} & 9.06 & 176.1 & 149.1 & \textbf{883.8} & 0.99 & 884.8 & 881.9 & \textbf{1.63} & 0.05 & 1.70 & 1.55  \\
    \midrule
    \multirow{6}{*}{\begin{tabular}[c]{@{}c@{}}Complex\\Office\end{tabular}} 
      & NBVP\cite{bircher2016receding}   & 900.0 & 0.0 & 900.0 & 900.0 & 391.2 & 6.69 & 400.4 & 382.9 & 1029.2 & 119.47 & 1156.1 & 778.7 & 0.43 & 0.01 & 0.44 & 0.43 \\
      & FUEL\cite{zhou2021fuel}          & 216.9 & 10.04 & 229.0 & 205.1 & 373.8 & 17.49 & 407.3 & 358.9 & 1617.5 & 1.18 & 1619.2 & 1616.1 & 1.72 & 0.06 & 1.78 & 1.63 \\
      & RACER\cite{zhou2023racer}        & 235.7 & 30.37 & 290.4 & 204.6 & 404.7 & 60.15 & 516.6 & 351.8 & 1649.0 & 1.96 & 1651.9 & 1646.4 & 1.71 & 0.04 & 1.78 & 1.64 \\
      & ECHO\cite{yu2023echo}            & 240.0 & 7.50 & 251.6 & 229.8 & 395.5 & 13.65 & 407.7 & 369.9 & 1636.4 & 1.87 & 1639.0 & 1634.4 & 1.65 & 0.03 & 1.68 & 1.61 \\
      & FAEP\cite{zhao2023autonomous}    & 191.8 & 7.72 & 203.7 & 182.4 & 317.8 & 16.64 & 346.0 & 295.2 & 1637.3 & 2.51 & 1641.6 & 1634.1 & 1.55 & 0.13 & 1.69 & 1.31 \\
      & Proposed                         & \textbf{153.4} & 6.22 & 164.7 & 146.6 & \textbf{274.8} & 12.80 & 297.5 & 259.1 & \textbf{1662.6} & 0.60 & 1663.6 & 1661.8 & \textbf{1.79} & 0.03 & 1.83 & 1.73 \\
    \midrule
    \multirow{6}{*}{\begin{tabular}[c]{@{}c@{}}Octa\\Maze\end{tabular}}         
    & NBVP\cite{bircher2016receding}   & 900.0 & 0.0 & 900.0 & 900.0 & 394.1 & 12.28 & 406.1 & 372.1 & 1187.9 & 35.28 & 1248.0 & 1140.2 & 0.44 & 0.01 & 0.45 & 0.41 \\
    & FUEL\cite{zhou2021fuel}          & 302.6 & 9.69 & 321.1 & 293.8 & 502.7 & 21.82 & 537.9 & 470.1 & 1933.8 & 2.14 & 1937.4 & 1931.0 & 1.66 & 0.03 & 1.71 & 1.60 \\
    & RACER\cite{zhou2023racer}        & 312.0 & 27.67 & 342.9 & 281.6 & 499.1 & 47.82 & 563.4 & 440.0 & 1977.1 & 3.87 & 1982.3 & 1972.2  & 1.60 & 0.04 & 1.64 & 1.56 \\
    & ECHO\cite{yu2023echo}            & 299.1 & 7.89 & 311.0 & 289.4 & 472.3 & 10.63 & 483.3 & 452.2 & 1966.9 & 2.19 & 1969.2 & 1964.1 & 1.58 & 0.04 & 1.63 & 1.52 \\
    & FAEP\cite{zhao2023autonomous}    & 281.0 & 6.69 & 288.6 & 270.3 & 421.4 & 19.41 & 456.3 & 398.9 & 1969.4 & 4.15 & 1975.3 & 1965.4 & 1.43 & 0.11 & 1.58 & 1.25 \\
    & Proposed                         & \textbf{216.7} & 3.97 & 221.5 & 211.8 & \textbf{377.9} & 7.69 & 384.4 & 365.5 & \textbf{2019.8} & 0.75 & 2020.4 & 2018.5 & \textbf{1.74} & 0.02 & 1.76 & 1.72 \\
    \midrule
    \multirow{6}{*}{\begin{tabular}[c]{@{}c@{}}DARPA\\Tunnel\end{tabular}}         
      & NBVP\cite{bircher2016receding}   & 650.8 & 87.83 & 734.0 & 500.3 & 212.8 & 25.48 & 235.0 & 172.6 & \textbf{462.5} & 3.03 & 466.0 & 458.6 & 0.33 & 0.01 & 0.35 & 0.32\\
      & FUEL\cite{zhou2021fuel}          & 67.4 & 6.09 & 75.4 & 58.2 & 110.7 & 6.84 & 119.5 & 101.6 & 433.0 & 3.58 & 437.2 & 427.8 & 1.64 & 0.08 & 1.81 & 1.58 \\
      & RACER\cite{zhou2023racer}        & 62.0 & 5.14 & 67.5 & 52.4 & 104.6 & 7.02 & 113.0 & 92.1 & 427.7 & 2.15 & 429.7 & 423.5 & 1.69 & 0.04 & 1.76 & 1.65\\
      & ECHO\cite{yu2023echo}            & 58.2 & 2.83 & 62.6 & 54.2 & 102.3 & 4.34 & 108.3 & 95.4 & 442.3 & 1.32 & 443.9 & 440.1 & 1.76 & 0.02 & 1.79 & 1.73 \\
      & FAEP\cite{zhao2023autonomous}    & 59.9 & 3.49 & 65.0 & 54.5 & 96.0 & 4.32 & 99.6 & 88.5 & 424.1 & 4.48 & 430.4 & 416.6 & 1.53 & 0.11 & 1.62 & 1.31\\
      & Proposed                         & \textbf{49.7} & 2.75 & 52.8 & 44.9 & \textbf{92.2} & 4.84 & 98.5 & 84.5 & 444.8 & 1.97 & 447.1 & 441.2 & \textbf{1.84} & 0.02 & 1.87 & 1.81\\
    \midrule
    \multirow{6}{*}{\begin{tabular}[c]{@{}c@{}}Duplex\\Office\end{tabular}} 
      & NBVP\cite{bircher2016receding}   & 900.0 & 0.00 & 900.0 & 900.0 & 407.9 & 70.92 & 503.5 & 316.6 & 1266.2 & 148.88 & 1395.0 & 984.6 & 0.45 & 0.08 & 0.56 & 0.35 \\
      & FUEL\cite{zhou2021fuel}          & 216.7 & 6.95 & 229.2 & 209.0 & 353.7 & 16.38 & 371.1 & 333.0 & 1356.6 & 3.33 & 1360.5 & 1351.8 & 1.63 & 0.05 & 1.69 & 1.57 \\
      & RACER\cite{zhou2023racer}        & 250.4 & 4.32 & 254.2 & 242.2 & 389.0 & 6.36 & 398.0 & 378.5 & 1380.5 & 2.22 & 1383.0 & 1377.8 & 1.55 & 0.02 & 1.59 & 1.53 \\
      & ECHO\cite{yu2023echo}            & 232.5 & 11.94 & 244.2 & 211.7 & 371.9 & 20.43 & 395.0 & 334.4 & 1372.9 & 14.35 & 1384.8 & 1344.7 & 1.60 & 0.02 & 1.62 & 1.57 \\
      & FAEP\cite{zhao2023autonomous}    & 202.7 & 3.66 & 208.9 & 198.8 & 306.7 & 15.52 & 328.0 & 288.7 & 1371.5 & 2.69 & 1374.4 & 1367.7 & 1.44 & 0.07 & 1.55 & 1.36 \\
      & Proposed                         & \textbf{178.1} & 8.19 & 186.6 & 163.4 & \textbf{298.5} & 17.37 & 321.5 & 272.0 & \textbf{1399.2} & 2.23 & 1402.6 & 1396.4 & \textbf{1.67} & 0.04 & 1.72 & 1.63 \\
    \midrule
    \multirow{6}{*}{\begin{tabular}[c]{@{}c@{}}Power\\Plant\end{tabular}}         
      & NBVP\cite{bircher2016receding}   & 900.0 & 0.00 & 900.0 & 900.0 & 384.8 & 11.89 & 403.2 & 369.5 & 2583.5 & 215.94 & 2863.3 & 2294.9 & 0.43 & 0.01 & 0.45 & 0.41 \\
      & FUEL\cite{zhou2021fuel}          & 378.4 & 18.08 & 402.6 & 356.6 & 631.2 & 32.49 & 678.0 & 586.4 & 4156.5 & 3.12 & 4160.8 & 4151.2 & 1.68 & 0.02 & 1.72 & 1.65 \\
      & RACER\cite{zhou2023racer}        & 424.1 & 30.57 & 473.8 & 388.7 & 660.6 & 27.80 & 702.2 & 624.9 & 4141.4 & 6.01 & 4149.3 & 4135.2 & 1.56 & 0.06 & 1.63 & 1.48 \\
      & ECHO\cite{yu2023echo}            & 441.2 & 19.57 & 456.9 & 403.1 & 677.3 & 35.61 & 716.3 & 613.7 & 4154.8 & 8.68 & 4164.2 & 4140.9 & 1.53 & 0.02 & 1.57 & 1.50 \\
      & FAEP\cite{zhao2023autonomous}    & 333.2 & 12.52 & 344.4 & 313.6 & 541.9 & 16.13 & 561.6 & 520.5 & 4195.5 & 8.76 & 4209.5 & 4185.2 & 1.34 & 0.14 & 1.52 & 1.19 \\
      & Proposed                         & \textbf{291.9} & 6.56 & 298.4 & 280.1 & \textbf{495.7} & 16.64 & 511.4 & 464.0 & \textbf{4212.8} & 4.08 & 4220.4 & 4209.6 & \textbf{1.70} & 0.02 & 1.73 & 1.66 \\
    \toprule\toprule
  \end{tabular}
  \label{tab:result}
\end{table*}

\begin{figure*}[t!]
	\centering
  \subfigtopskip=0pt
	\subfigbottomskip=2pt
	\subfigcapskip=-3pt
  \subfigure[NBVP]{\includegraphics[width=0.3\columnwidth]{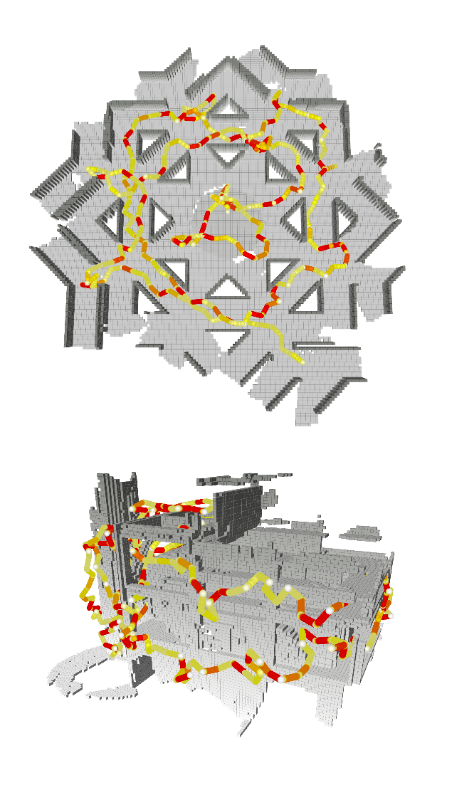}} \hskip -2pt  
  \subfigure[FUEL]{\includegraphics[width=0.3\columnwidth]{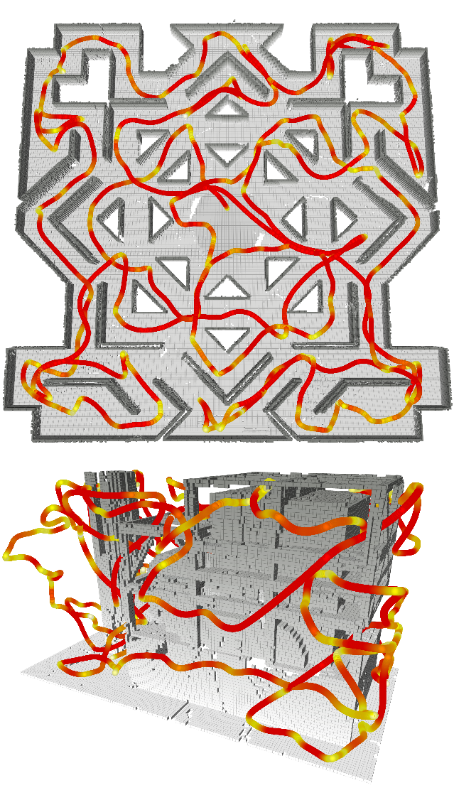}} \hskip -2pt 
  \subfigure[RACER]{\includegraphics[width=0.3\columnwidth]{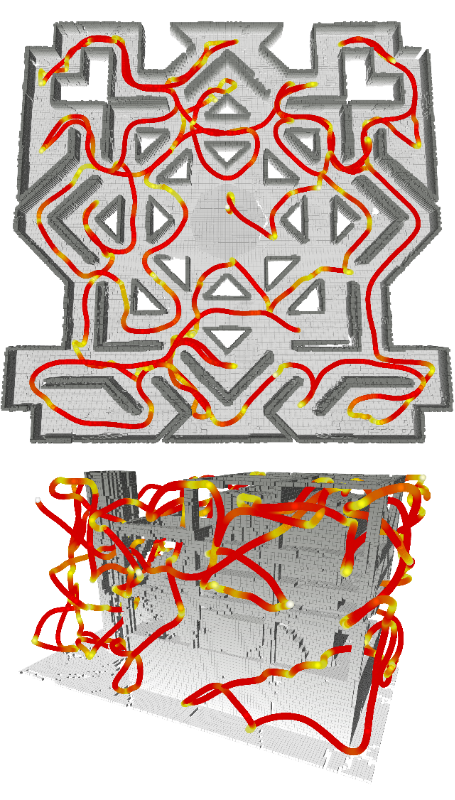}} \hskip -2pt
  \subfigure[ECHO]{\includegraphics[width=0.3\columnwidth]{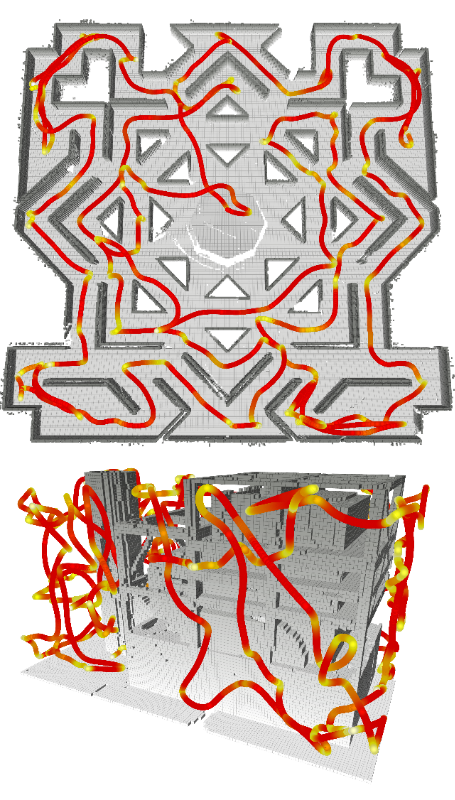}} \hskip -2pt
  \subfigure[FAEP]{\includegraphics[width=0.3\columnwidth]{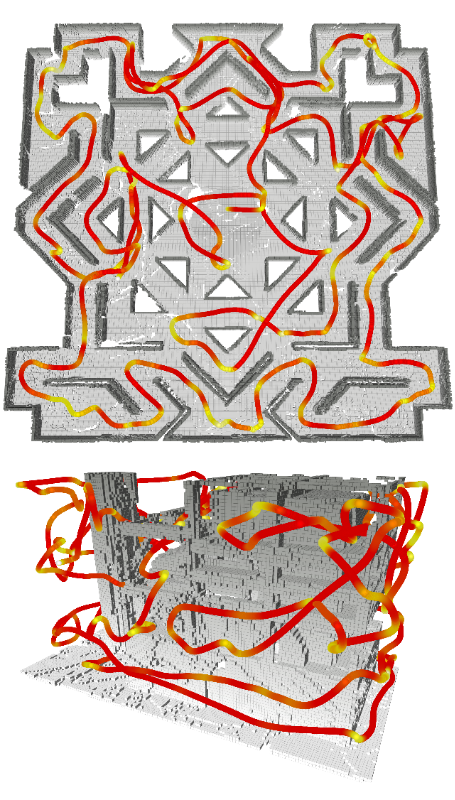}} \hskip -2pt
  \subfigure[Proposed]{\includegraphics[width=0.3\columnwidth]{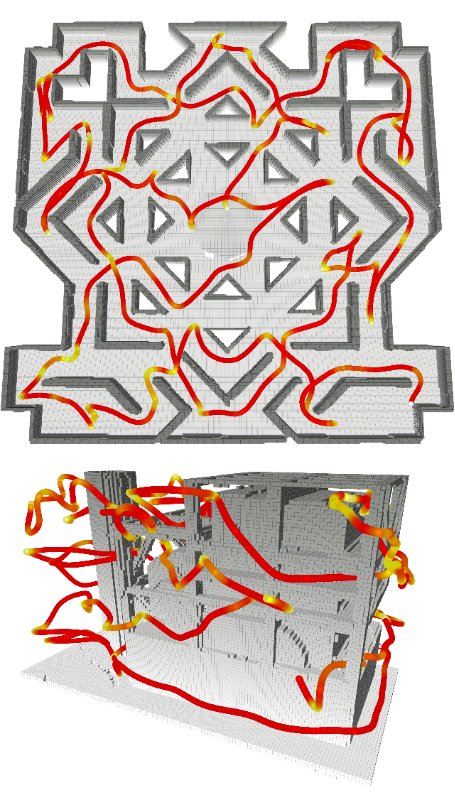}} \hskip -2pt
  \subfigure{\includegraphics[width=0.09\columnwidth]{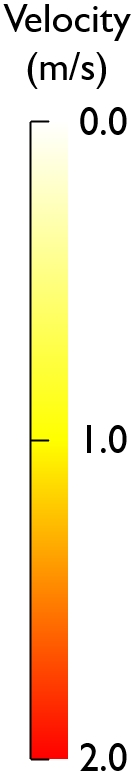}} \hskip -2pt
  \caption{\label{fig:sim_exe_traj} The final executed trajectories from all the six exploration planners in \textit{Octa Maze} and \textit{Power Plant}.}
  \vspace{-0.4cm}
\end{figure*}

\begin{figure*}[t]
	\centering
  \subfigtopskip=0pt
	\subfigbottomskip=2pt
	\subfigcapskip=-3pt 
  \subfigure{\includegraphics[width=1.5\columnwidth]{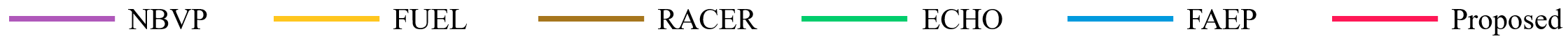}}   
  \setcounter{subfigure}{0}
  \subfigure[Classical Office]{\includegraphics[width=0.6\columnwidth]{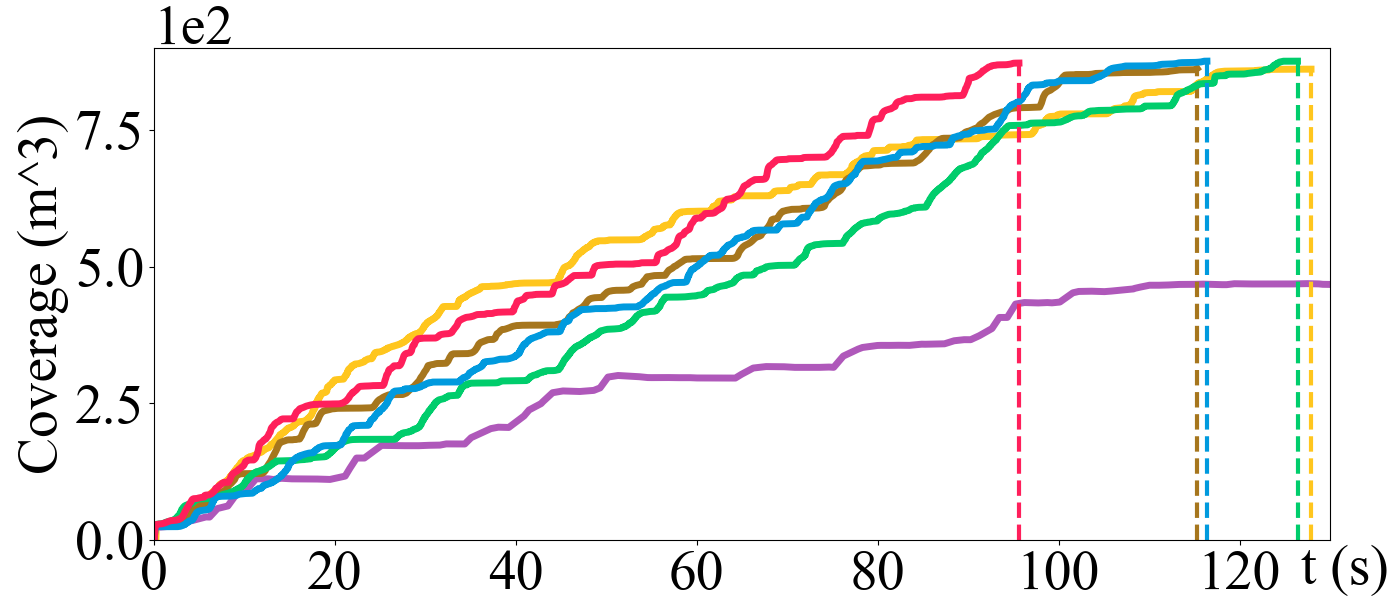}}
  \subfigure[Complex Office]{\includegraphics[width=0.6\columnwidth]{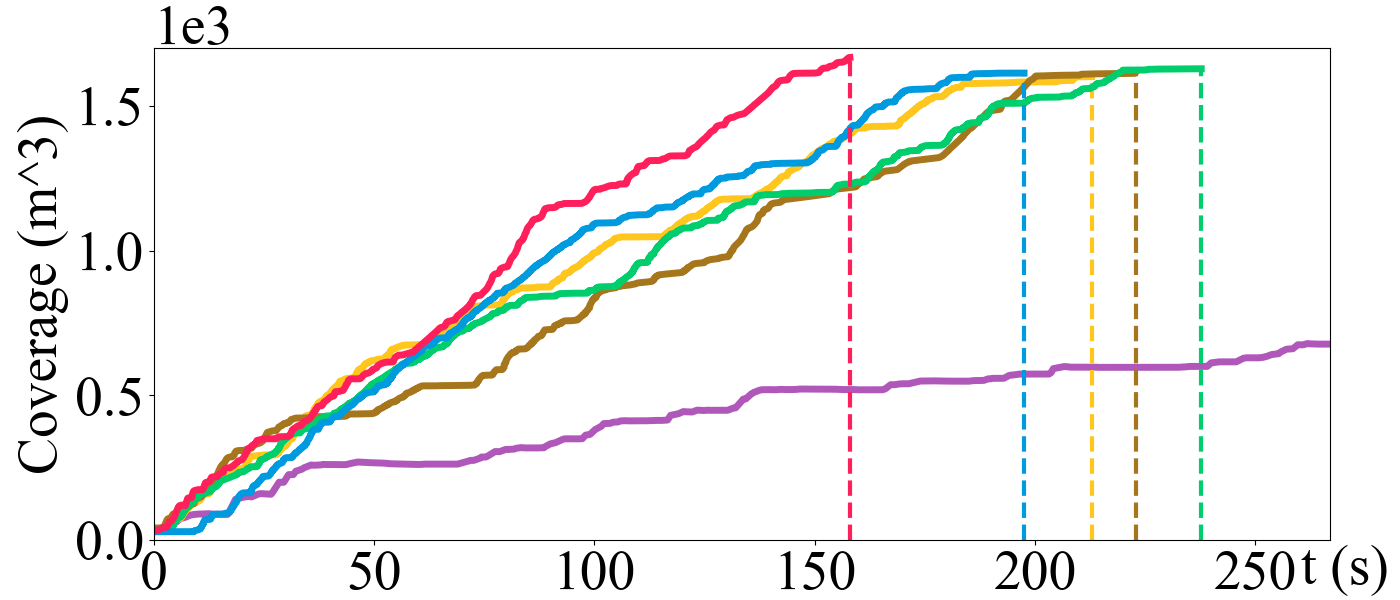}}   
  \subfigure[Octa Maze]{\includegraphics[width=0.6\columnwidth]{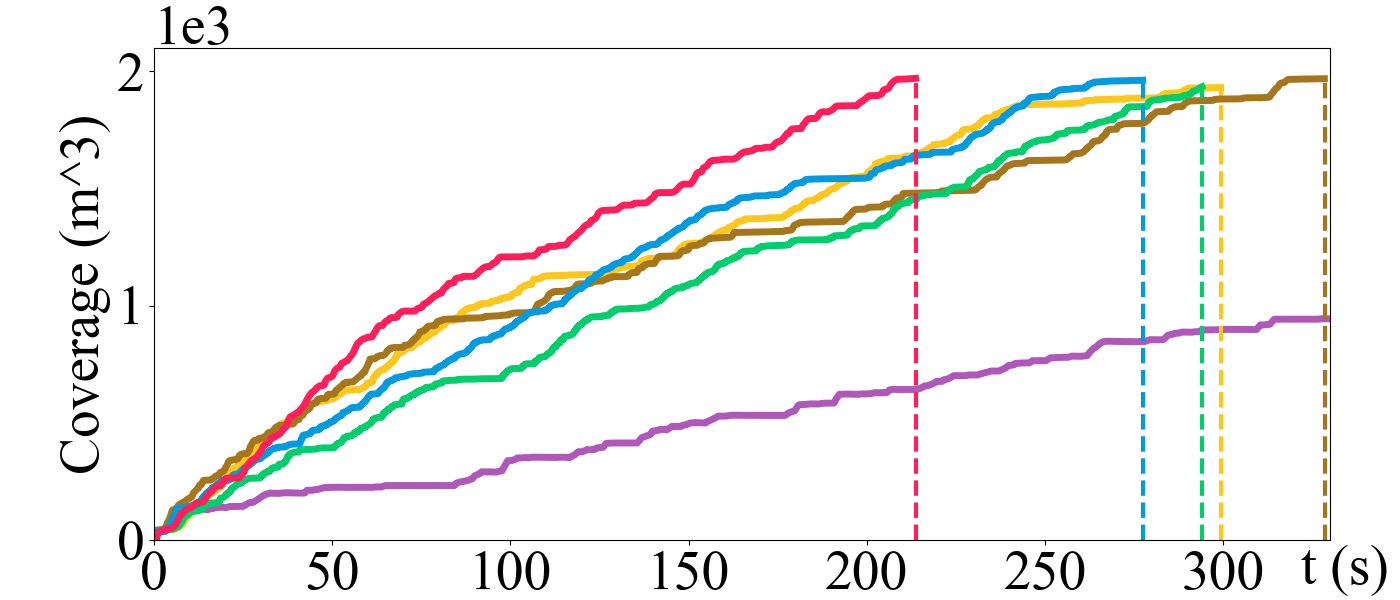}}   
  \subfigure[DARPA Tunnel]{\includegraphics[width=0.6\columnwidth]{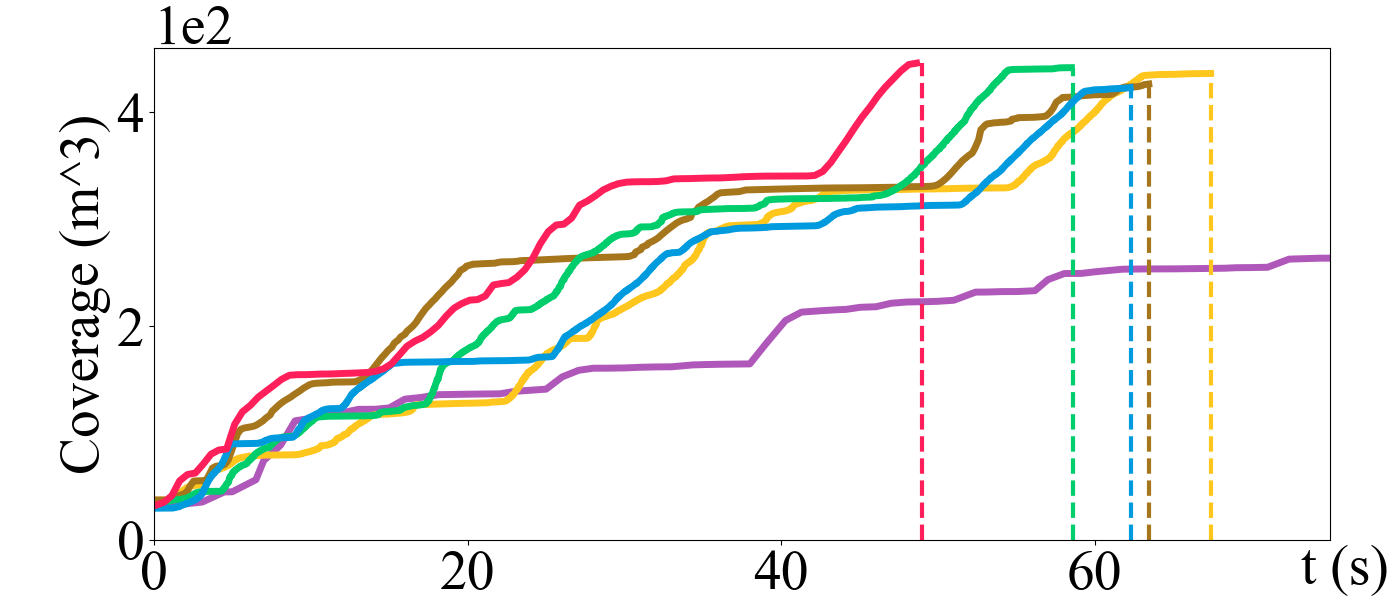}}   
  \subfigure[Duplex Office]{\includegraphics[width=0.6\columnwidth]{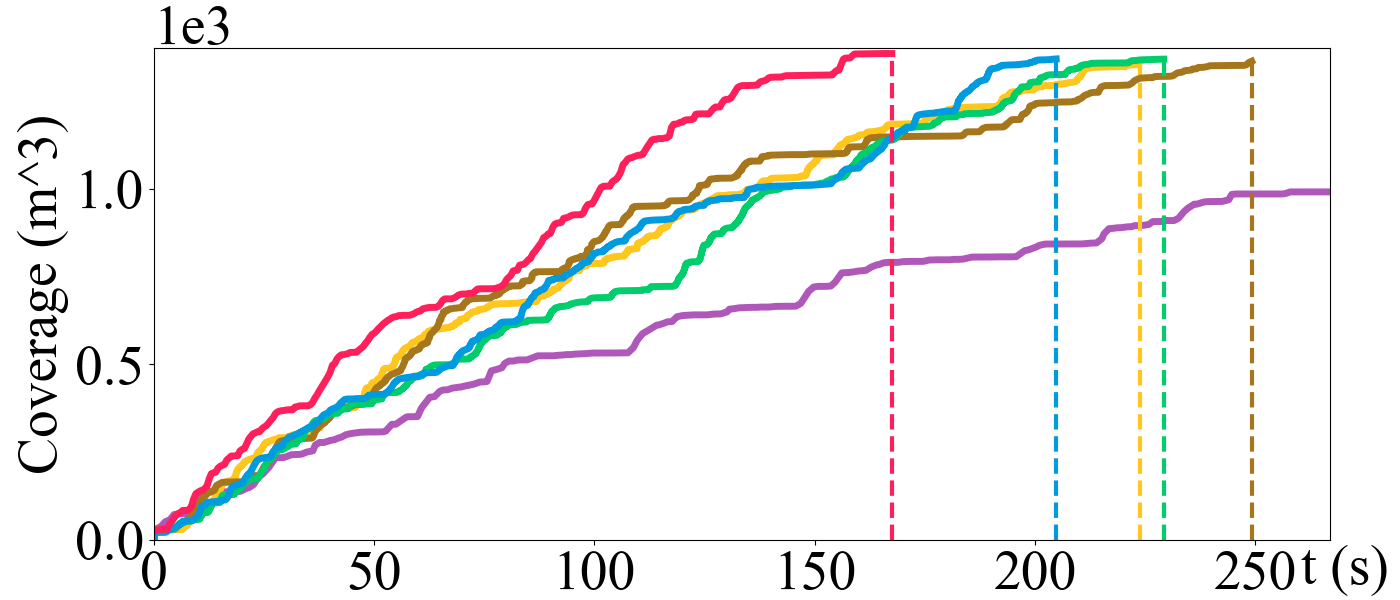}}   
  \subfigure[Power Plant]{\includegraphics[width=0.6\columnwidth]{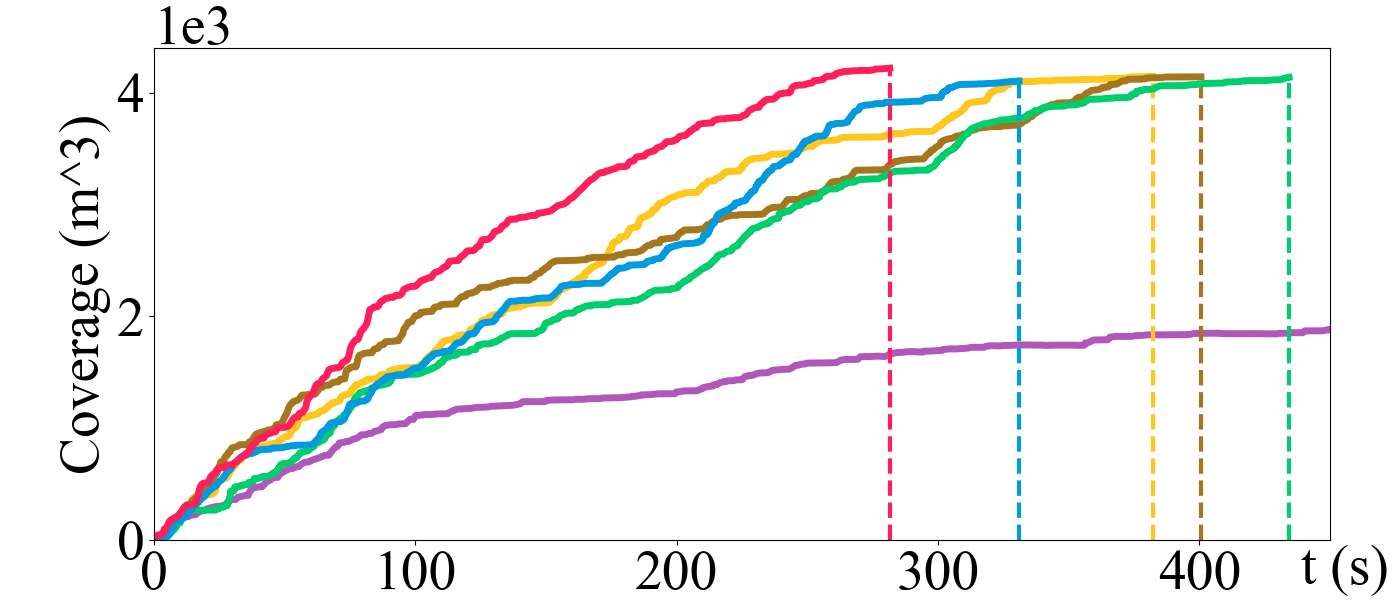}}   
  \caption{\label{fig:sim_progress_plot} The exploration progress of all the five \rev{state-of-the-art} benchmarks and the proposed exploration planner \textbf{FALCON} in all the six testing scenarios.}
  \vspace{-0.2cm}
\end{figure*}

\begin{figure*}[t]
	\centering
  \includegraphics[width= 1.8\columnwidth]{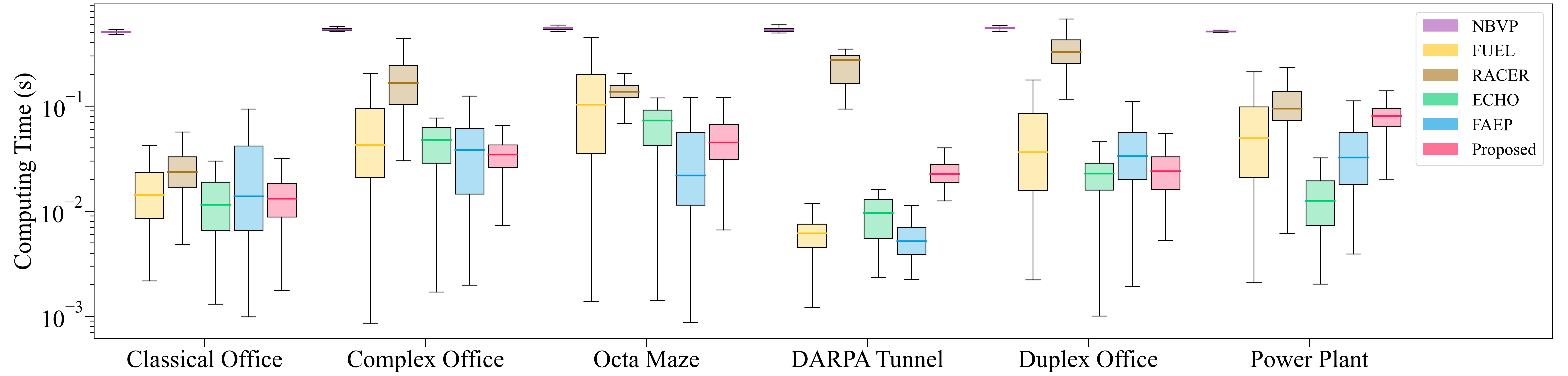}  
  \vspace{-0.2cm}
  \caption{\label{fig:exp_computing_times} The box plot \rev{with logarithmic scale} of computation time for a single planning iteration across all six exploration planners in the six testing scenarios.}
  \vspace{-0.4cm}
\end{figure*}

\subsubsection{Benchmark Candidates}
The \rev{benchmark} comparison evaluates \textbf{FALCON} against a series of \rev{state-of-the-art} visual-based exploration planners that prioritize rapid exploration completion.
The planner bundle includes NBVP\cite{bircher2016receding}, FUEL\cite{zhou2021fuel}, RACER\cite{zhou2023racer}, ECHO\cite{yu2023echo} and FAEP\cite{zhao2023autonomous}, listed in ascending order of publication dates.
NBVP is a sampling-based exploration planner that employs a receding horizon strategy to determine the \rev{next-best view} based on the amount of unmapped space that can be explored.
FUEL is a more \rev{sophisticated} hybrid planner that generates a global path of frontier visitation order and then refines a local set of viewpoint samples around the frontiers.
Instead of \rev{a} frontier visitation order, RACER opts for a coverage path of \rev{unexplored} regions as the global guidance.
As RACER is a multi-robot exploration planner, we adapt it for a single-robot version in this benchmark comparison.
ECHO utilizes a delicate heuristic evaluation function to determine the next frontier target, without a global tour.
Since there is no open-source code available for ECHO, we use our implementation.
FAEP improves upon FUEL by adding frontier-level factors to cost formulation and introducing an adaptive yaw planning module.
For all these methods, the mapping modules inherited from their source code are utilized.
Note that the impact of their differences on the exploration should be limited, as they typically relate to the reconstruction near obstacle surfaces whereas the exploration planners focus on the boundary regions between known and unknown spaces.
In the benchmark experiments, the linear and angular dynamics limitations as well as the sensor model are set identical for all methods, as shown in \rev{Table~\ref{tab:params}}. 
\rev{All benchmark experiments are subject to a time limit of $900$ seconds, beyond which they are terminated prematurely and left incomplete.}

\subsection{Simulation Benchmark Results And Analysis}
\label{subsec:sim_benchmark}

All the five state-of-the-art benchmarks and the proposed exploration planner are comprehensively evaluated across all six scenarios.
For each planner in each scenario, the statistics from $10$ runs are summarized in \rev{Table~\ref{tab:result}} and the final exploration trajectories in \textit{Octa Maze} and \textit{Power Plant} are visualized in Fig.~\ref{fig:sim_exe_traj}. 
The exploration progresses are \rev{plotted} in Fig.~\ref{fig:sim_progress_plot} and the box plots of the computation time of a single planning iteration are shown in Fig.~\ref{fig:exp_computing_times}.
The complete exploration experiments are presented in the supplementary video due to limited space.

The analysis of the benchmark comparison results is conducted according to the \textit{VECO} criteria that an ideal exploration planner should satisfy, as mentioned in Sec.~\ref{sec:intro}.
In the subsequent discussion, we further elaborate on the \textit{VECO} criteria and analyze the benchmark comparison results accordingly.

\subsubsection{\textbf{V}ersatility} \label{subsubsec:versatility}
\textit{An exploration planner should exhibit effective performance across diverse environments, regardless \rev{of} the level of accessibility of the entire space or the complexity of obstacles present.}

For the versatility criterion, we analyze the performance of each planner based on the accessibility and complexity characteristics of the testing \rev{scenarios} as shown in \rev{Table~\ref{tab:exp_scenarios}}.
NBVP only accomplishes exploration in small and simple scenarios like \textit{Classical Office} and \textit{DARPA Tunnel}, exhibiting limited versatility.
ECHO has satisfactory performance in some of the 2.5D scenarios but shows relatively slow performance in scenarios like \textit{Complex Office} and \textit{Power Plant}, primarily due to its carefully designed cost function which does not generalize well.
FUEL performs stably across scenarios with different levels of accessibility and complexity but its \rev{exploration efficiency} ranks low in all testing scenarios.
FAEP demonstrates good versatility and ranks within the top three in all six testing scenarios.
RACER is the runner-up in \textit{Classical Office}, but its performance deteriorates in high-complexity environments like \textit{Complex Office} and low-accessibility environments like \textit{DARPA Tunnel}.
This is because of high \rev{computational} complexity in \rev{a} high-complexity environment and constant misleading by inaccessible regions in \rev{a} low-accessibility environment.
The proposed planner \textbf{FALCON} consistently outperforms all the benchmarks in all the scenarios, exhibiting excellent versatility across different environments.

\begin{table*}[t]
  \centering
  \caption{General Ablation Studies Statistics in Complex Office}
  \vspace{-0.2cm}
  \begin{tabular}{ccccccccccccccccc}
    \toprule\toprule
    \multirow{2}{*}{\textbf{Method}} & \multicolumn{4}{c}{\textbf{Exploration Time (s)}} & \multicolumn{4}{c}{\textbf{Flight Distance (m)}} & \multicolumn{4}{c}{\textbf{Coverage (m$^3$)}} & \multicolumn{4}{c}{\textbf{Avg Velocity (m/s$^2$)}} \\
                                    & \textbf{Avg} & \textbf{Std} & \textbf{Max} & \textbf{Min}
                                    & \textbf{Avg} & \textbf{Std} & \textbf{Max} & \textbf{Min}
                                    & \textbf{Avg} & \textbf{Std} & \textbf{Max} & \textbf{Min}
                                    & \textbf{Avg} & \textbf{Std} & \textbf{Max} & \textbf{Min} \\
    \midrule
    \textbf{\texttt{UniDec}}                           & 164.4 & 6.86 & 172.1 & 152.6 & 297.2 & 15.78 & 316.5 & 270.7 & 1651.2 & 0.83 & 1651.9 & 1649.6 & \textbf{1.80} & 0.04 & 1.89 & 1.77 \\
    \textbf{\texttt{NoCG}}                           & 189.9 & 5.91 & 197.2 & 180.6 & 320.4 & 12.45 & 332.4 & 302.6 & 1653.1 & 1.54 & 1655.0 & 1651.2 & 1.68 & 0.04 & 1.73 & 1.62 \\
    $\textbf{\texttt{CP}}_\textbf{\texttt{AF}}$                         & 169.8 & 8.30 & 181.1 & 160.8 & 288.3 & 16.98 & 308.8 & 266.7 & \textbf{1663.1} & 0.32 & 1663.5 & 1662.6 & 1.69 & 0.02 & 1.72 & 1.66 \\
    $\textbf{\texttt{CP}}_\textbf{\texttt{UN}}$                         & 171.4 & 5.57 & 178.6 & 163.7 & 298.6 & 9.93 & 311.8 & 285.3 & 1662.2 & 1.63 & 1664.5 & 1659.9 & 1.74 & 0.02 & 1.77 & 1.71 \\
    \textbf{\texttt{NoSOP}}                          & 171.7 & 9.39 & 185.7 & 162.8 & 300.4 & 16.95 & 321.9 & 283.1 & 1650.8 & 0.88 & 1651.6 & 1649.2 & 1.75 & 0.02 & 1.77 & 1.71 \\
    Proposed                         & \textbf{153.4} & 6.22 & 164.7 & 146.6 & \textbf{274.8} & 12.80 & 297.5 & 259.1 & 1662.6 & 0.60 & 1663.6 & 1661.8 & 1.79 & 0.03 & 1.83 & 1.73 \\
    \toprule\toprule
  \end{tabular}
  \vspace{-0.4cm}
  \label{tab:ablation}
\end{table*} 

\subsubsection{\rev{\textbf{E}xploration Efficiency}} \label{subsubsec:exploration_efficiency}
\textit{An exploration planner should generate motions that minimize the exploration duration, covering accessible space as quickly as possible.}

For the \rev{exploration efficiency} criteria, each planner is assessed based on the average exploration time in \rev{Table~\ref{tab:result}}.
NBVP needs rather long exploration time and even fails to complete tasks in large and complex scenarios.
This is due to the heavy computation required for the next-best-view selection, which results in slow flight speed and even pauses during exploration, as shown in Fig.~\ref{fig:sim_exe_traj}(a).
ECHO neglects global optimality, leading to \rev{back-and-forth} exploration motions, as shown in Fig.~\ref{fig:sim_exe_traj}(d).
Both FUEL and FAEP demonstrate good performance by utilizing global guidance of frontier visitation order, and FAEP refines FUEL by adaptive yaw planning, yielding quite good results in all scenarios.
Instead of planning only in free space as the previous four methods do, RACER and the proposed planner \textbf{FALCON} generate global paths considering unknown regions.
However, RACER frequently produces back-and-forth motions reducing exploration efficiency, as shown in Fig.\ref{fig:sim_exe_traj}(c).
This is because its global guidance is not well followed, especially in high-complexity environments.
Fig.\ref{fig:sim_progress_plot} reveals that all the benchmarking planners have a long-tail issue to different degrees.
It indicates that some corners are not thoroughly explored during the exploration, requiring the agent to return to them at the end to complete the exploration.
The proposed exploration planner \textbf{FALCON} mitigates this long-tail problem and outperforms all the benchmarks in all scenarios, achieving the shortest exploration time.
In \textit{Duplex Office}, \textbf{FALCON} is 13.81\% faster than the runner-up, and in the more challenging scenario \textit{Octa Maze}, it is 29.67\% faster.
\textbf{FALCON}'s superior performance is achieved by a reasonable global coverage path guidance and a flexible and globally optimized local planner that consistently incorporates the global guidance's intention.

\subsubsection{\textbf{C}ompleteness} \label{subsubsec:completness}
\textit{When an exploration planner reports completion, the entire space should have been thoroughly explored and reconstructed, with accessible space covered as much as possible.}

For the \rev{completeness} criteria, the planners are compared based on the average coverage in \rev{Table~\ref{tab:result}} and the reconstruction quality shown in Fig.~\ref{fig:sim_exe_traj}.
In general, the coverage statistic for all methods is satisfactory, \rev{except} for NBVP, which performs well only in small and simple scenarios such as \textit{DARPA Tunnel}.
However, as shown in Fig.~\ref{fig:sim_exe_traj} (b-e), the reconstructed volumetric maps generated by the benchmarks commonly have some small corners or walls missing, among which FAEP has the worst reconstruction quality.
The \rev{completeness} performance is influenced by multiple factors, including volumetric mapping methods, frontier discard policy\rev{,} and exploration motion continuity.
The \rev{proposed} planner \textbf{FALCON} achieves a good overall performance in terms of both coverage \rev{volume} and reconstruction quality, as demonstrated in Fig.~\ref{fig:sim_exe_traj} (f).

\subsubsection{\rev{Resp\textbf{O}nsiveness}} \label{subsubsec:responsiveness}
\textit{An exploration planner should be computationally efficient, capable of promptly responding and generating exploration motions in real-time when new sensor data arrives.}

For the \rev{responsiveness} criteria, the planners are evaluated according to the computation time of a single planning iteration shown in Fig.~\ref{fig:exp_computing_times}.
\rev{NBVP} is the most computationally expensive planner due to its sampling-based nature.
RACER has high computation time in high-complexity \rev{environments} like \textit{Duplex Office} where the number of fine decomposition cells increases exponentially, and in low-accessibility \rev{environments} like \textit{DARPA Tunnel} where the planner needs to constantly consider inaccessible regions.
FUEL and FAEP have similar \rev{responsiveness} performance as their computational complexity \rev{is} related to the number of frontiers. 
Hence, their computation time shows \rev{a} relatively large standard deviation, as they compute quickly at the start and end of exploration and slower during the middle when the number of frontiers is large.
ECHO demonstrates relatively good computational efficiency by omitting global planning and solely evaluating the scalar heuristic function for each frontier.
Thanks to the connectivity graph boosted cost \rev{evaluation}, the proposed planner \textbf{FALCON} exhibits good \rev{responsiveness}, with an acceptable median and low standard deviation \rev{in one-iteration computation time}, while capable of generating motions with high \rev{exploration efficiency}.

\section{\rev{Ablation Studies}}
\label{sec:ablation}

The proposed method is further validated through a series of ablation studies in simulation. 
To justify the effectiveness of each proposed module, the ablation experiments quantitatively evaluate the complete exploration planner proposed comparing against baseline variants.

\subsection{Ablation Settings and General Experiments}
\label{subsec:abla_general}
According to the proposed modules, we design the following baseline variants for comparison:
\begin{itemize}
  \item \textbf{\texttt{UniDec}}: This variant of the space decomposition module (Sec.~\ref{subsec:space_decom}) uniformly decomposes the exploration space into cells with a fixed size, rather than employing the proposed connectivity-aware approach.
  \item \textbf{\texttt{NoCG}}: This variant for the connectivity graph (CG) module (Sec.~\ref{subsec:connectivity_graph}) omits the connectivity graph and directly uses voxel-based A* for path search between any two positions.
  \item $\textbf{\texttt{CP}}_\textbf{\texttt{AF}}$ and $\textbf{\texttt{CP}}_\textbf{\texttt{UN}}$: These are two variants of the coverage path (CP) planning module (Sec.~\ref{subsec:cp_construct}) that incorporate only active free (AF) zones and only unknown (UN) zones respectively as coverage path elements.
  \item \textbf{\texttt{NoSOP}}: This variant for the CP-guided local path planning module (Sec.~\ref{subsec:CP_guided_planning}) adopts a simple local planning strategy with local horizon, which considers only the next cell in the CP.
\end{itemize}

We compare these baseline variants with the proposed method in terms of general exploration efficiency in the \textit{Complex Office}.
The configurations are the same as in \rev{Table~\ref{tab:params}} and the statistics for $5$ runs are presented in \rev{Table~\ref{tab:ablation}}. 
The results demonstrate that the full proposed method, with all modules functioning, achieves the best performance with shortest exploration duration and flight distance.
This highlights the effectiveness and contribution of each proposed module towards the overall exploration efficiency.
In the following Sec.~\ref{subsec:abla_space_decomp}~-~\ref{subsec:abla_sop}, the efficacy of each module will be further discussed and analyzed in \rev{detail}.

\subsection{Ablation Study of Space Decomposition}
\label{subsec:abla_space_decomp}
\begin{figure}[t]
	\centering
  \vspace{-0.2cm}
  \includegraphics[width=0.8\columnwidth]{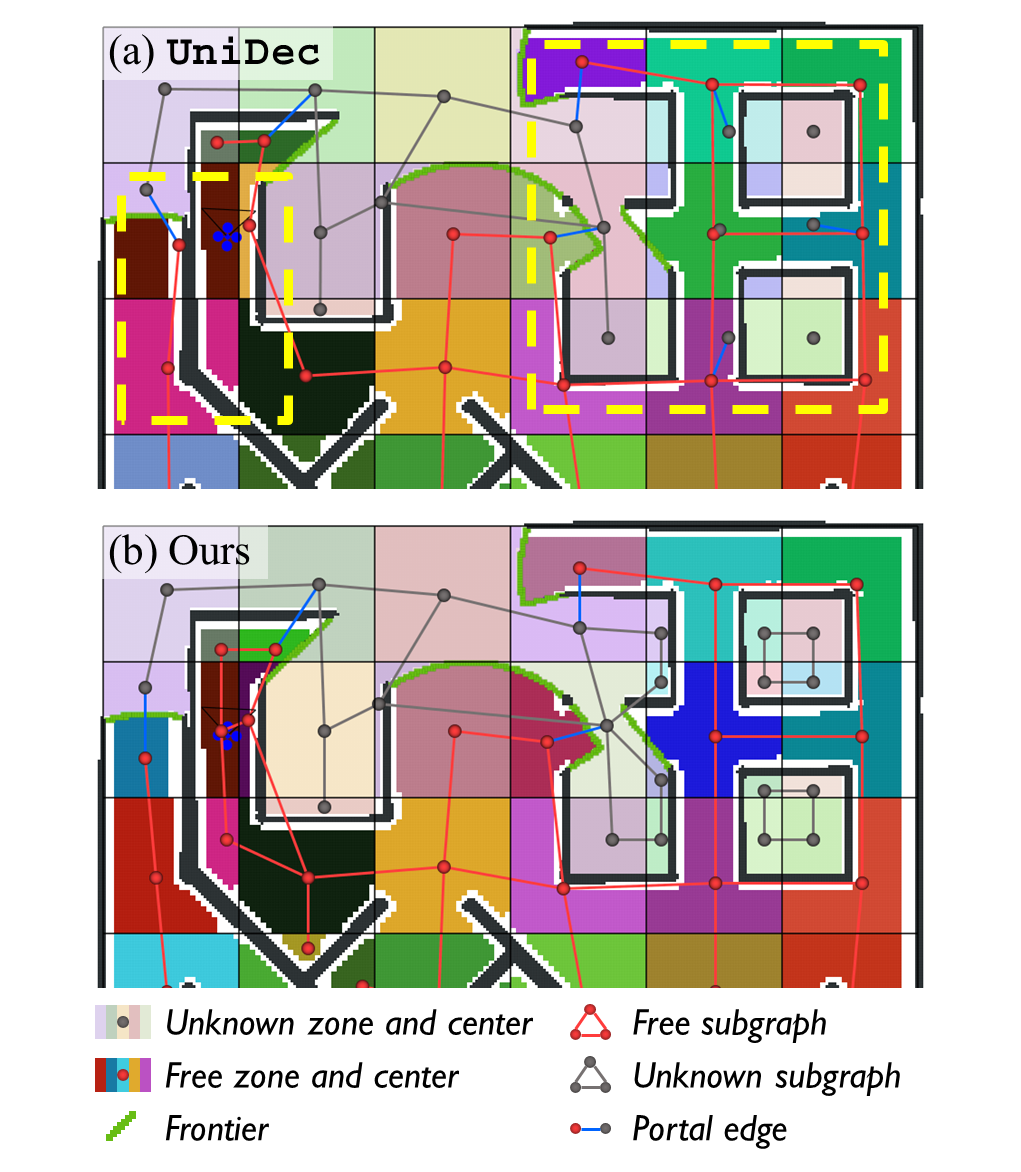}  
  \vspace{-0.4cm}
  \caption{\label{fig:abla_space_decomp} A comparison of space decomposition results and the connectivity graph built upon between the proposed method and the \textbf{\texttt{UniDec}} in the \textit{Complex Office}.}
  \vspace{-0.8cm}
\end{figure}

The \textbf{\texttt{UniDec}} variant replaces the proposed connectivity-aware space decomposition with a simple uniform decomposition approach.
Specifically, voxels within a uniform cell are clustered based on their occupancy status.
Then one free centroid and one unknown \rev{centroid} are computed as the average positions of the voxels within each cluster, as shown in Fig.~\ref{fig:abla_space_decomp}(a).
These centroids are later used as the elements during connectivity graph construction and coverage path planning.

We compare the space decomposition results and the connectivity graph produced by the proposed method versus \textbf{\texttt{UniDec}} during an exploration.
As shown in Fig.~\ref{fig:abla_space_decomp}(a), it can be observed that \textbf{\texttt{UniDec}} generates unreasonable centroids and connectivity graph in some areas, which cannot represent the environmental structure.
For example, in the right highlighted region, the unknown centroids are placed within free space, and in the left highlighted rectangle, the segregated free spaces are treated as one entity and assigned a single centroid.
This may result in problematic coverage paths, consistently misleading the exploration planner towards unreachable regions.
In contrast, as depicted in Fig.~\ref{fig:abla_space_decomp}(b), the proposed connectivity-aware decomposition yields a more reasonable space decomposition and connectivity graph that better \rev{captures} the actual topology of the environment.
The overall exploration duration of \textbf{\texttt{UniDec}} is $7.2\%$ longer than the proposed method, as shown in \rev{Table~\ref{tab:ablation}}.

\begin{table}[t]
  \caption{Ablation Study of Connectivity Graph\\on Computational Efficiency}
  \begin{tabular}{M{0.9cm}M{0.6cm}M{0.7cm}M{0.7cm}M{1.3cm}M{1.3cm}M{0.6cm}}
    \toprule\toprule
    \multirow{4}{*}{\textbf{Scene}} & \multirow{4}{*}{\textbf{Study}} & \multicolumn{5}{c}{\textbf{Average Computation Time (ms)}} \\
    \cmidrule{3-7}
     & & \begin{tabular}[c]{@{}c@{}}\textbf{Space}\\\textbf{Decom.}\\\textbf{\ref{subsec:space_decom}}\end{tabular} & \begin{tabular}[c]{@{}c@{}}\textbf{Conn.}\\\textbf{Graph}\\\textbf{\ref{subsec:connectivity_graph}}\end{tabular} & \begin{tabular}[c]{@{}c@{}}\textbf{Coverage}\\\textbf{Path$^\dagger$}\\\textbf{\ref{subsec:cp_construct}}\end{tabular} & \begin{tabular}[c]{@{}c@{}}\textbf{Local}\\\textbf{Planning$^\dagger$}\\\textbf{\ref{subsec:CP_guided_planning}}\end{tabular} & \multirow{2}{*}{\textbf{Total$^\diamond$}} \\
    \midrule
    \multirow{2}{*}{\begin{tabular}[c]{@{}c@{}}Complex\\Office\end{tabular}} 
    & NoCG & 0.83 & 6.96 & 114.15+5.07 & 32.01+0.60 & 171.14 \\ \cmidrule{2-7}
    & Full & 0.73 & 6.33 & \textbf{10.88}+4.68 & \textbf{2.09}+0.51 & \textbf{37.18} \\
    \midrule
    \multirow{2}{*}{\begin{tabular}[c]{@{}c@{}}Power\\Plant\end{tabular}} 
    & NoCG & 3.72 & 11.96 & 563.99+42.74 & 128.92+8.71 & 760.28 \\ \cmidrule{2-7}
    & Full & 3.74 & 9.17 & \textbf{16.47}+38.2 & \textbf{10.38}+5.97 & \textbf{86.21} \\
    \toprule\toprule
  \end{tabular} \\
  \footnotesize{$^\dagger$ The computation time is seperately counted for \textit{cost matrix construction time} + \textit{solver resolution time}.}\\
  \footnotesize{$^\diamond$ The total computation time for each planning iteration, including time for other modules such as frontiers identification and trajectory generation.}\\
  \vspace{-0.6cm}
  \label{tab:avg_compute_time}
\end{table}

\begin{figure}[t]
	\centering
  \subfigtopskip=0pt
	\subfigbottomskip=2pt
	\subfigcapskip=-3pt
  \subfigure[Connectivity graph]{\includegraphics[width=0.32\columnwidth]{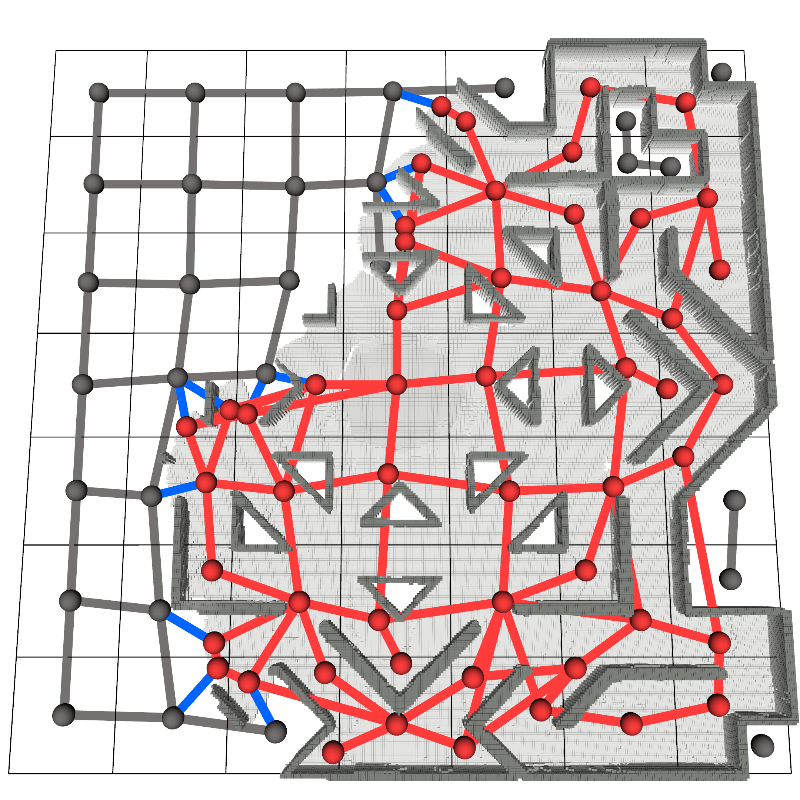}} \hskip -2pt
  \subfigure[Ours]{\includegraphics[width=0.32\columnwidth]{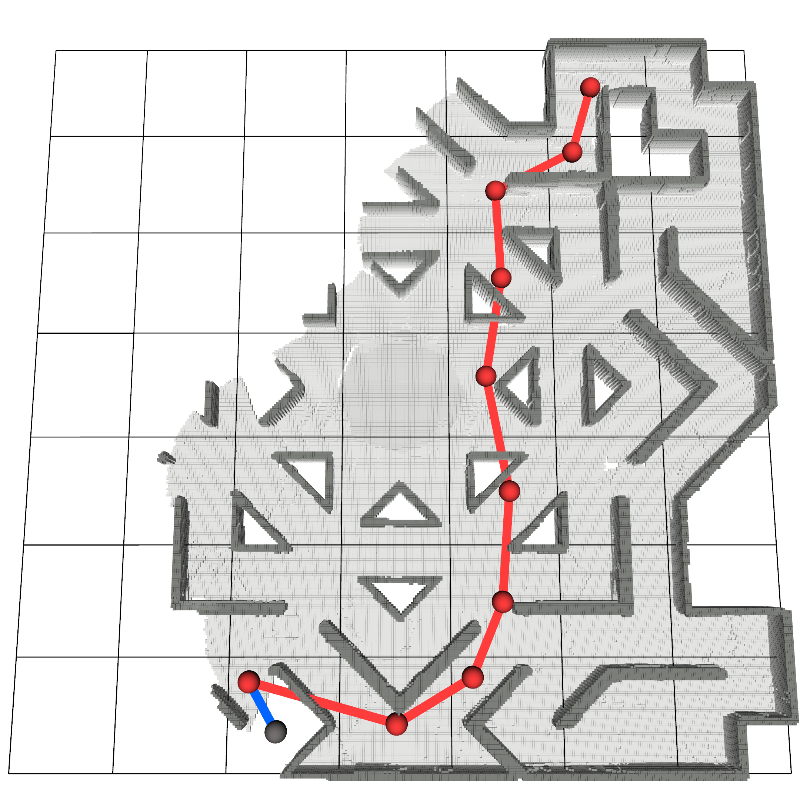}} \hskip -2pt      
  \subfigure[\textbf{\texttt{NoCG}}]{\includegraphics[width=0.32\columnwidth]{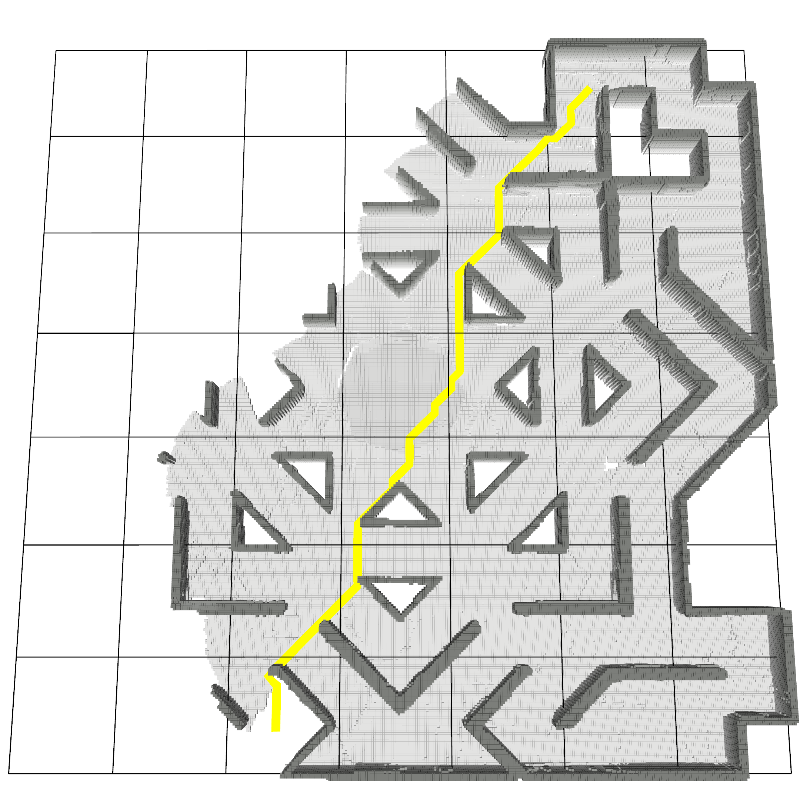}} \hskip -2pt      
  \caption{\label{fig:abla_cg_search} (a) shows the connectivity graph constructed by the proposed method during an \rev{exploration}. (b) and (c) display the path search results for cost evaluations from a graph-based A* search on the connectivity graph and a voxel-based A* search respectively.}
  \vspace{-1.2cm}
\end{figure}

\subsection{Ablation Study of Connectivity Graph}
\label{subsec:abla_cg}
The \textbf{\texttt{NoCG}} variant excludes the module of connectivity graph construction.
Consequently, distances between any two positions are computed through voxel-based A* searches when calculating the cost matrices $\matr{C}_\text{cp}$ and $\matr{C}_\text{sop}$ mentioned in Sec.~\ref{subsec:cp_construct} and \ref{subsec:CP_guided_planning} respectively.
In contrast, the proposed hybrid method leverages voxel-based A* search for short-distance targets below threshold $d_\text{thr}$ and graph-based A* search on connectivity graph otherwise.
Note that the path searches discussed in this section are only for the purpose of cost evaluation when constructing the matrices $\matr{C}_\text{cp}$ and $\matr{C}_\text{sop}$, which are not used in the actual path planning for execution.

\begin{figure}[t]
	\centering
  \includegraphics[width=0.85\columnwidth]{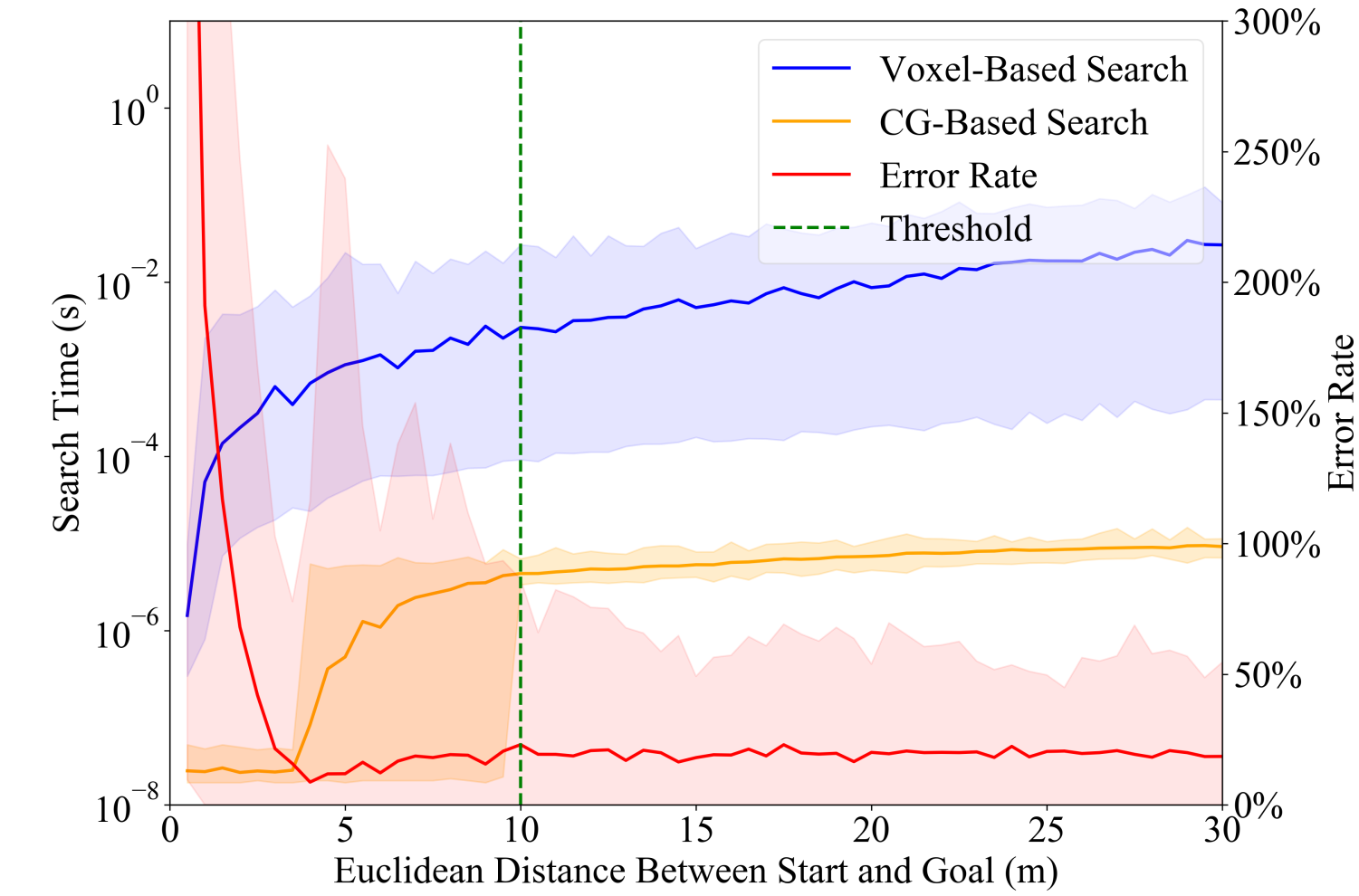}  
  \vspace{-0.2cm}
  \caption{\label{fig:abla_cg_plot} This figure plots the path search time of both voxel-based and graph-based searches for target pairs with different Euclidean distances sampled in \textit{Octa Maze}.
  It also plots the error rate of the graph-based search path length with respect to a voxel-based search.
  The threshold $d_\text{thr}$ utilized in our hybrid method is \rev{illustrated} as green dashed line.}
  \vspace{-1.8cm}
\end{figure}

\rev{Table~\ref{tab:avg_compute_time}} shows the average computation time of each module in a planning iteration for both \textbf{\texttt{NoCG}} and the full proposed method in \textit{Complex Office} and \textit{Power Plant} scenarios.
The statistics reveal that the cost matrix construction time of \textbf{\texttt{NoCG}} is over $10$ times higher compared to the proposed method, significantly increasing the total computation time.
As an example, Fig.~\ref{fig:abla_cg_search} showcases path searches between two long-distance targets performed by our hybrid method and \textbf{\texttt{NoCG}}.
With a real-time constructed connectivity graph (Fig.~\ref{fig:abla_cg_search}(a)), a $47.24$ m path (Fig.~\ref{fig:abla_cg_search}(b)) is searched on the graph in a negligible $0.04$ ms.
Without a connectivity graph, the voxel-based search conducted by \textbf{\texttt{NoCG}} takes $9.4$ ms to find a path with $37.58$ m length (Fig.~\ref{fig:abla_cg_search}(c)).
Although a voxel-based search provides a shorter path, it comes with a significantly higher computation cost.
This is further validated in Fig.~\ref{fig:abla_cg_plot}, which plots the error rate \rev{of searched path lengths}, calculated by $p_\text{graph} \slash p_\text{voxel} -1$, with both of the graph-based and voxel-based path searching time.
The plot reveals that graph-based search for short-distance targets may introduce highly inaccurate shortest path lengths while voxel-based search for long-distance targets may take much longer time up to two magnitudes.
Our hybrid method provides a balance between computational efficiency and pathfinding accuracy by utilizing voxel-based search for short-distance targets below threshold $d_\text{thr}$ and graph-based search for long-distance targets above threshold $d_\text{thr}$.
Considering the fact that this cost evaluation is performed repeatedly for each entry in the cost matrices $\matr{C}_\text{cp}$ and $\matr{C}_\text{sop}$,  our proposed hybrid cost evaluation method offers substantial computational savings.

The overall exploration efficiency in \rev{Table~\ref{tab:ablation}} indicates that \textbf{\texttt{NoCG}} diminishes the performance, resulting in $23.8\%$ longer exploration time compared to the full proposed method.
The long planning time is the primary reason for \textbf{\texttt{NoCG}}'s performance degradation, as it prevents the exploration planner from promptly responding to \rev{the} latest environment information.
In worst-case scenarios, the agent must stop and wait for planning completion.
By boosting the exploration planning \rev{responsiveness}, the connectivity graph is validated to result in more \rev{efficient} exploration motions with shorter durations.

\subsection{Ablation Study of Global Guidance}
\label{subsec:abla_cp}
We validate the effectiveness of our coverage path global guidance by comparing \rev{it} to two variants, $\textbf{\texttt{CP}}_\textbf{\texttt{AF}}$ and $\textbf{\texttt{CP}}_\textbf{\texttt{UN}}$. 
During global path planning, $\textbf{\texttt{CP}}_\textbf{\texttt{AF}}$ only considers active free zones, while $\textbf{\texttt{CP}}_\textbf{\texttt{UN}}$ only considers unknown zones.
The global paths produced by these two variants are similar to those generated by FUEL\cite{zhou2021fuel} and RACER\cite{zhou2023racer} respectively.
\rev{In contrast, the proposed method takes into account both active free and unknown zones as coverage path elements.}

\begin{figure}[t]
	\centering
  \includegraphics[width=0.75\columnwidth]{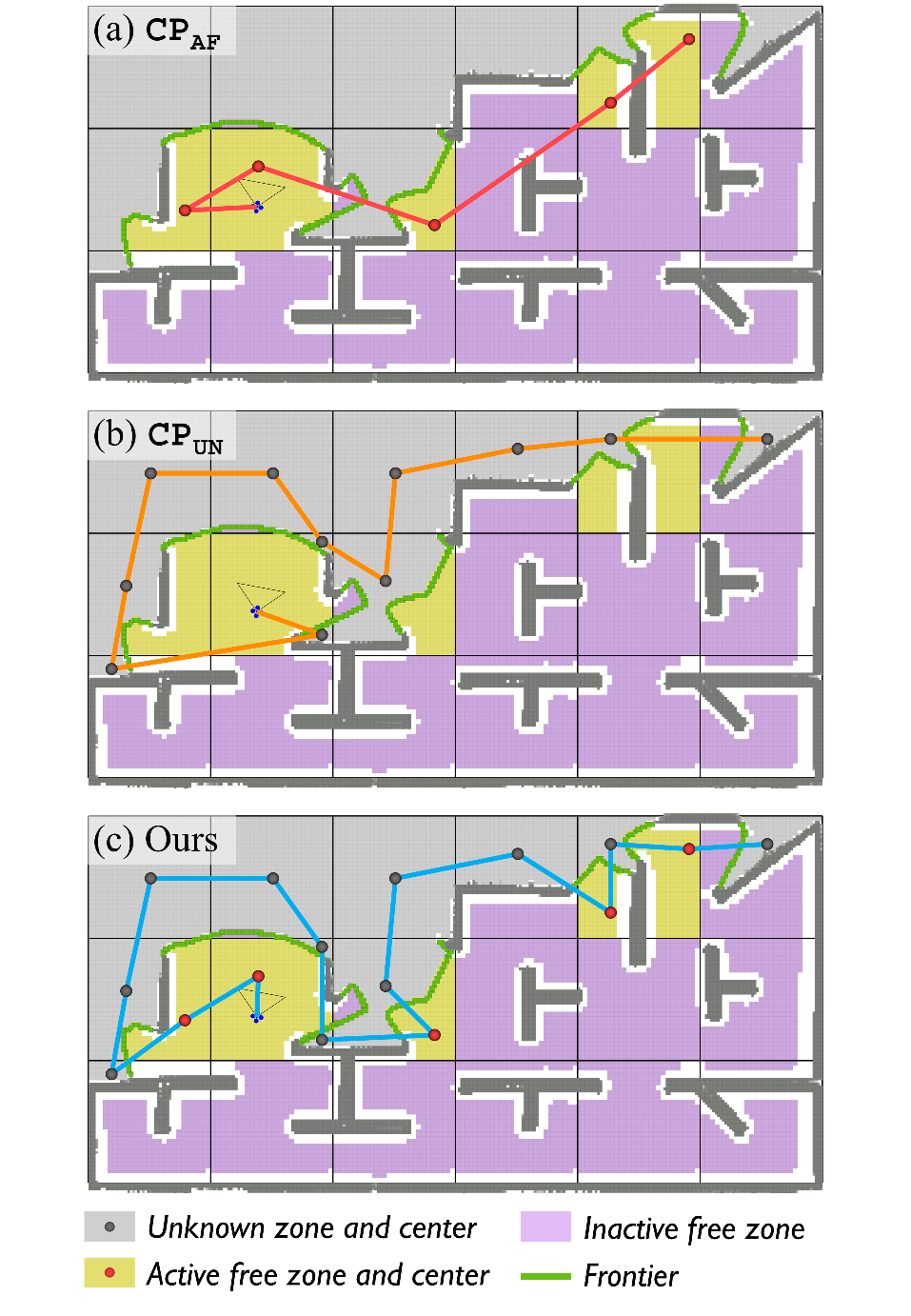}  
  \vspace{-0.4cm}
  \caption{\label{fig:abla_cp} Comparison of the global guidances generated by variants $\textbf{\texttt{CP}}_\textbf{\texttt{AF}}$, $\textbf{\texttt{CP}}_\textbf{\texttt{UN}}$ and the proposed method in the \textit{Classical Office}.}
  \vspace{-0.4cm}
\end{figure}

\begin{figure}[t]
	\centering
  \subfigtopskip=0pt
	\subfigbottomskip=2pt
	\subfigcapskip=-3pt
  \subfigure[\textbf{\texttt{NoSOP}}]{\includegraphics[width=0.35\columnwidth]{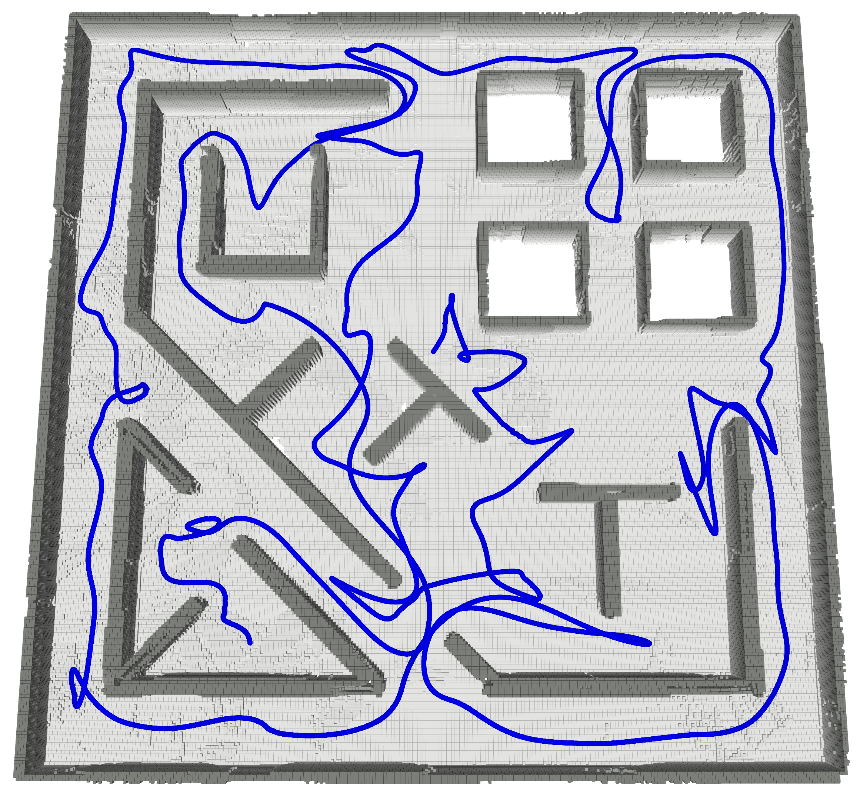}}
  \subfigure[Ours]{\includegraphics[width=0.35\columnwidth]{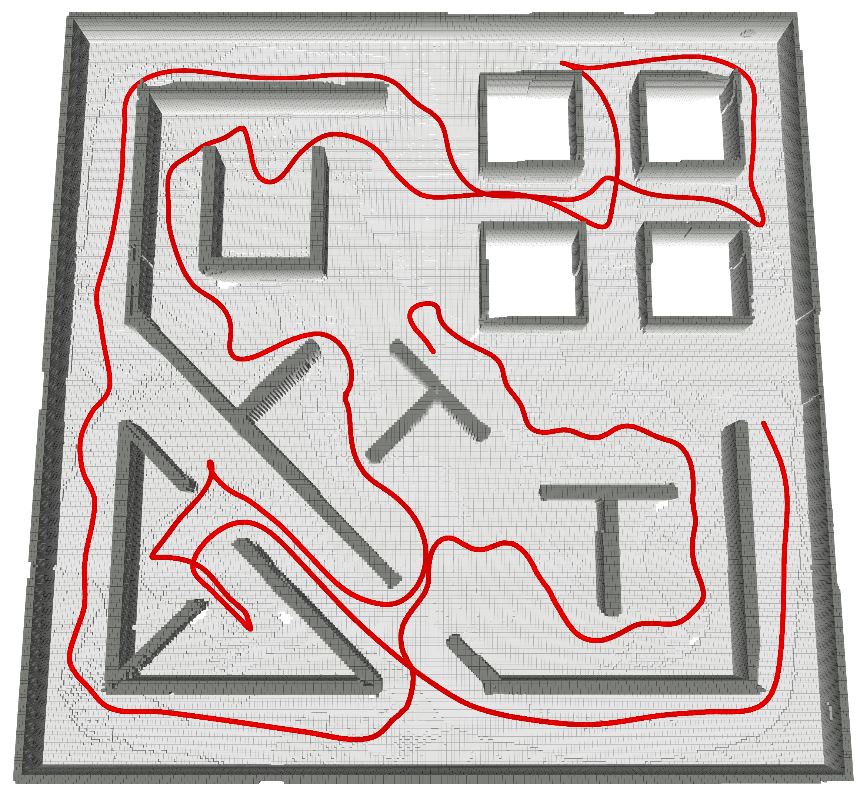}}
  \caption{\label{fig:abla_sop_traj} The exploration trajectories executed by \textbf{\texttt{NoSOP}} and the proposed planner in \textit{Complex Office} are illustrated in (a) and (b) respectively.}
  \vspace{-0.6cm}
\end{figure}

The global paths produced by $\textbf{\texttt{CP}}_\textbf{\texttt{AF}}$, $\textbf{\texttt{CP}}_\textbf{\texttt{UN}}$ and the proposed approach are illustrated in Fig.~\ref{fig:abla_cp}.
As the global path from $\textbf{\texttt{CP}}_\textbf{\texttt{AF}}$ \rev{focuses} solely on frontier regions, it fails to provide a coverage of the entire unexplored space, as shown in Fig.~\ref{fig:abla_cp}(a).
However, exploration involves not only visiting current frontiers but also covering the entire space.
This mismatch between \rev{the} optimization objective and the exploration mission leads to inefficient exploration with \rev{a} longer duration.
On the other hand, the global path from $\textbf{\texttt{CP}}_\textbf{\texttt{UN}}$ overlooks active free regions.
Although it covers the entire unexplored space, it sometimes generates suboptimal global guidances that \rev{lead} to unnecessarily long trajectories, as shown in Fig.~\ref{fig:abla_cp}(b).
This occurs because considering only unknown regions cannot \rev{model} the full exploration process, where the agent ﬁrst reaches frontiers before pushing them into unknown space.
In contrast, the proposed method better models this exploration process by planning coverage paths incorporating both active free and unknown zones, as shown in Fig.~\ref{fig:abla_cp}(c).
As shown in the statistics in \rev{Table~\ref{tab:ablation}}, $\textbf{\texttt{CP}}_\textbf{\texttt{AF}}$ and $\textbf{\texttt{CP}}_\textbf{\texttt{UN}}$ \rev{have} similar performance that \rev{completes} exploration using at least $10.7\%$ longer exploration time \rev{compared} to the proposed method.
The proposed coverage path including both unknown and active free zones consistently provides reasonable global guidance on \rev{the} visitation order of the entire space, achieving the best performance among the three approaches.

\subsection{Ablation Study of Local Path Planning}
\label{subsec:abla_sop}
Instead of the proposed local planner that utilizes SOP to insert viewpoints into \rev{the} coverage path, the variant \textbf{\texttt{NoSOP}} adopts a simple local planning strategy with \rev{a} local horizon.
It formulates the local planning as an ATSP problem for \rev{the} current position, all frontier viewpoints in the current cell, and \rev{the} center position of the next cell in global guidance.
The costs between positions are similarly calculated.

\begin{figure}[t] 
	\centering 
  \subfigtopskip=0pt
	\subfigbottomskip=2pt 
	\subfigcapskip=-3pt
  \subfigure[]{\includegraphics[width=0.75\columnwidth]{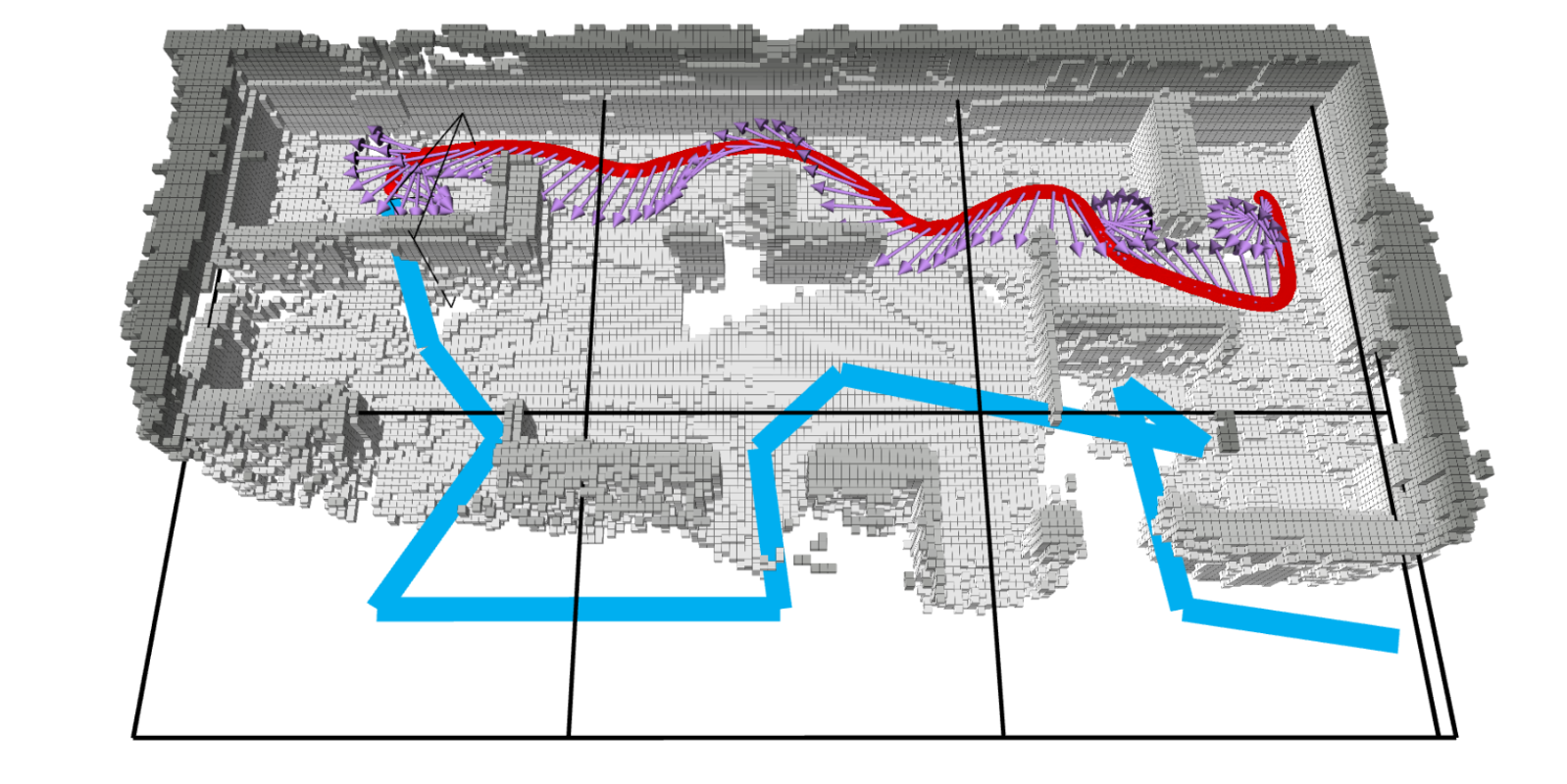}}     
  \subfigure[]{\includegraphics[width=0.75\columnwidth]{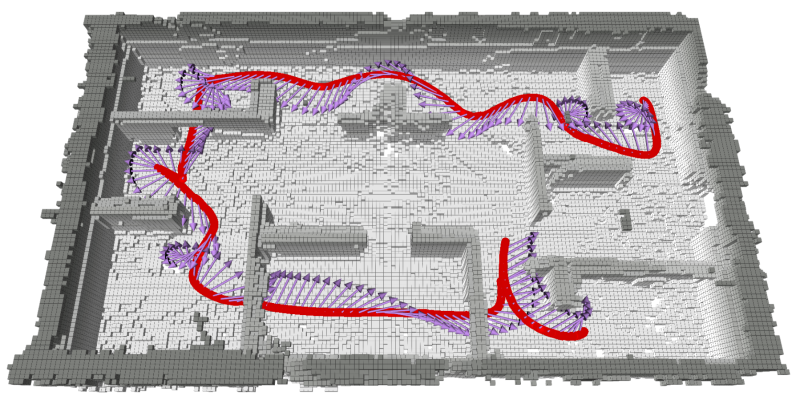}}     
  \subfigure[]{\includegraphics[width=0.75\columnwidth]{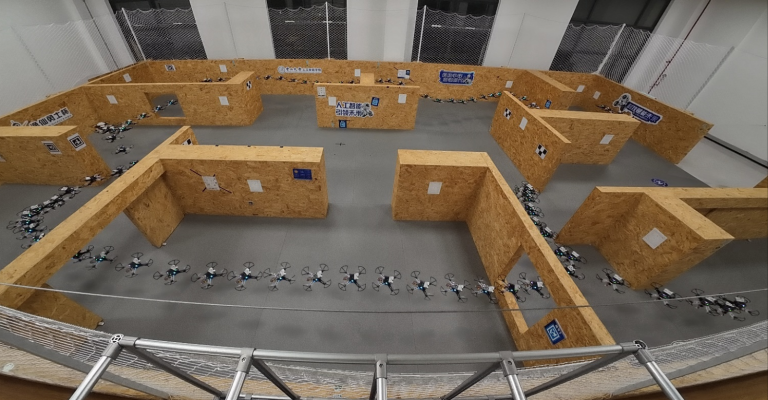}}     
  \vspace{-0.1cm}
  \caption{\label{fig:realworld_office} Real-world experiment in an office-like environment. (a) A snapshot of the exploration trajectory and coverage path planning result. (b) Reconstructed volumetric map and the final executed trajectory. (c) Composite image of the exploration experiment.}
  \vspace{-0.8cm}
\end{figure} 

As demonstrated in Fig.~\ref{fig:abla_sop_traj}, the exploration trajectory generated by \textbf{\texttt{NoSOP}} is more \rev{circuitous} and contains more revisitations compared with the proposed planner.
Statistics in \rev{Table~\ref{tab:ablation} show} that the variant \textbf{\texttt{NoSOP}} degrades the performance with $11.9\%$ longer exploration time compared with the proposed method.
Given the same global planning of \rev{the} coverage path, the performance decrease of \textbf{\texttt{NoSOP}} is mainly due to the inconsistency of local planning with the \rev{intention} of global guidance. 
\textbf{\texttt{NoSOP}} produces local \rev{paths only with local horizons}, dropping all the information on the global guidance except for the next cell to go.
This validates \rev{that} the proposed local planning module is able to consciously incorporate the intention of the CP guidance and produces efficient local paths for fast exploration.

\section{\rev{Real-World Experiments}}
\label{sec:real_exp}
We further conduct field experiments to validate \rev{the} practical feasibility of the proposed framework for fully autonomous exploration in three challenging real-world environments.
In all the real-world experiments, we utilize a customized lightweight quadrotor platform with a take-off weight of $850$ g.
The quadrotor is equipped with an NVIDIA Jetson Orin NX\footnote{\url{https://www.nvidia.com/en-us/autonomous-machines/embedded-systems/jetson-orin}} 16 GB, a NxtPx4 autopilot \cite{liu2024omninxt} and a forward-looking global-shutter stereo camera RealSense D435i\footnote{\url{https://www.intelrealsense.com/depth-camera-d435i}}.
The dynamics limitations of the quadrotor are set to $v_m = 1.0$ \rev{m/s}, $a_m = 1.0$ \rev{m/s$^2$}, $\dot{\xi}_m = 1.05$ \rev{rad/s}, $\ddot{\xi}_m = 1.05$ \rev{rad/s$^2$}.
The localization of the quadrotor is provided by a visual-inertial state estimator VINS-Fusion\cite{qin2019general}.
Both of the planning and localization modules were operated onboard without other external infrastructure.

\begin{figure}[t]
	\centering
  \subfigtopskip=0pt
	\subfigbottomskip=2pt
	\subfigcapskip=-3pt 
  \subfigure[]{\includegraphics[width=0.75\columnwidth]{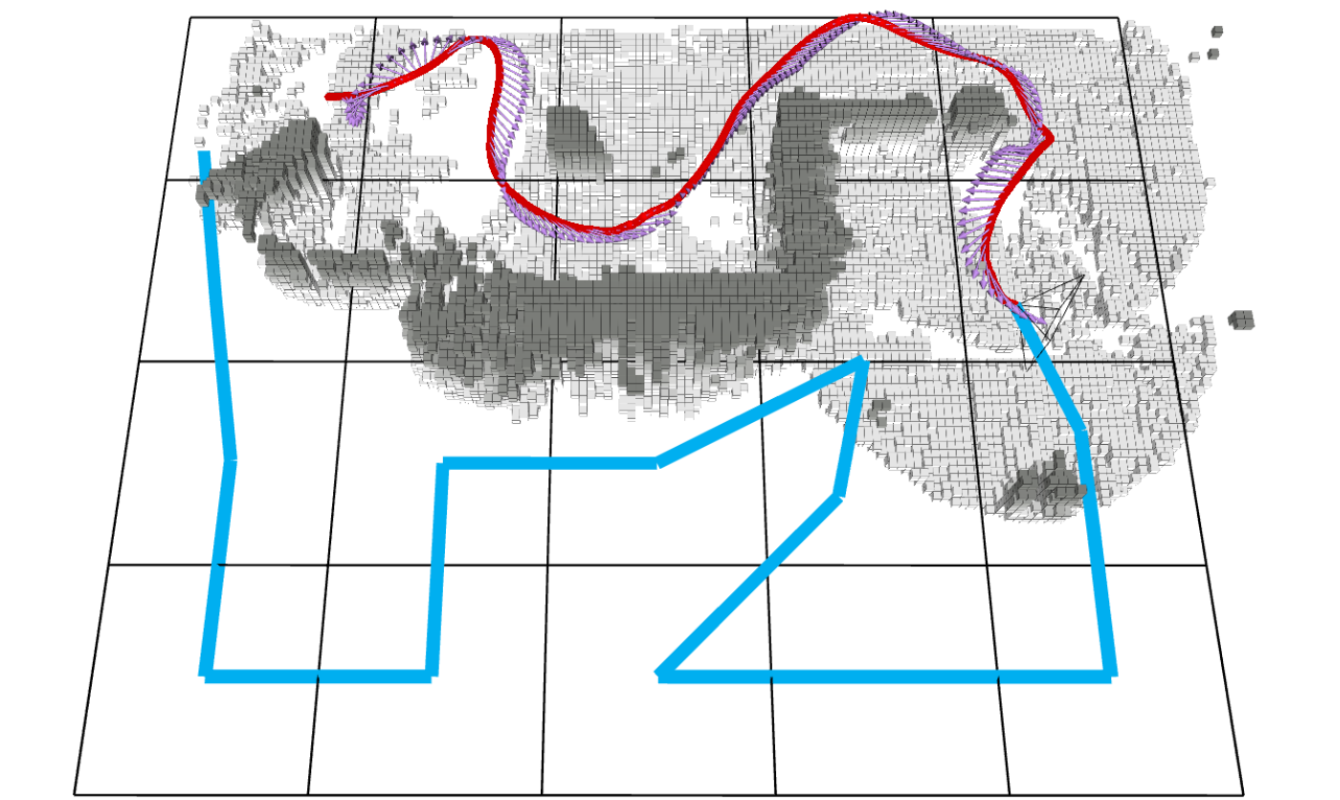}}     
  \subfigure[]{\includegraphics[width=0.75\columnwidth]{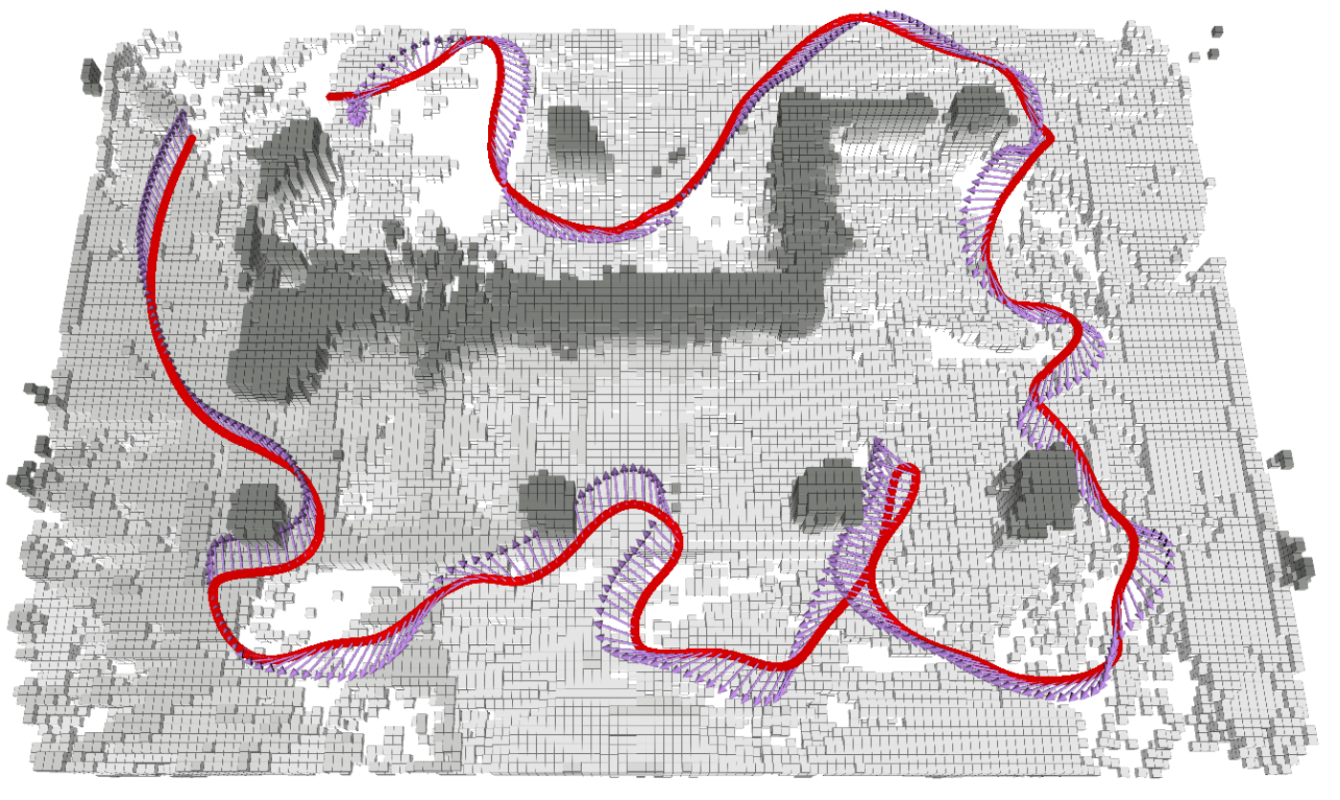}}     
  \subfigure[]{\includegraphics[width=0.75\columnwidth]{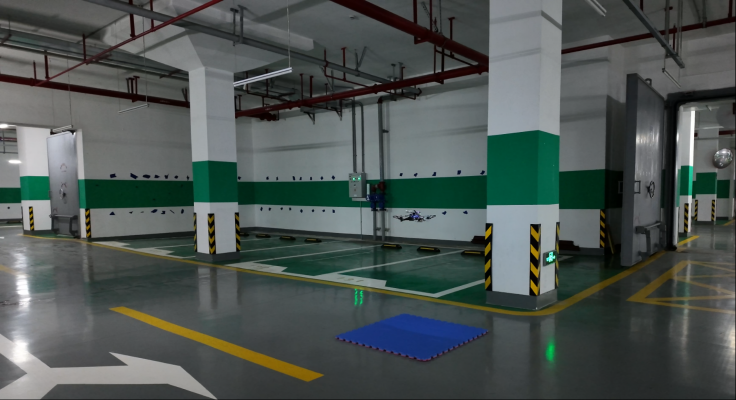}}     
  \vspace{-0.1cm}
  \caption{\label{fig:realworld_garage} \rev{A real-worl} experiment in \rev{a} parking lot. (a) Exploration trajectory and coverage path planning result during the exploration. (b) Final volumetric map and executed trajectory results. (c) Photo of the site.}
  \vspace{-0.8cm}
\end{figure}

We deploy our approach fully onboard for autonomous exploration experiments in three scenes. The first scene is an office-like environment with dimensions $15 \times 9 \times 1.2$ \rev{m$^3$,} as shown in Fig.~\ref{fig:realworld_office}(c).
The scene is partitioned into different rooms connected by windows and corridors.
The exploration task is completed in $46.3$ seconds, with a trajectory length of $37.7$ m, as demonstrated in Fig.~\ref{fig:realworld_office}(b).
A snapshot during the exploration is depicted in Fig.~\ref{fig:realworld_office}(a), showing the executed trajectory, online constructed volumetric map\rev{,} and the global coverage path (blue segments).
The second scene is a large-scale parking lot of $25 \times 20 \times 2$ \rev{m$^3$}, featuring pillars, walls\rev{,} and gates, as shown in Fig.~\ref{fig:realworld_garage}(c).
The exploration is finished in $95.9$ seconds, with a quadrotor movement distance of $88.3$ m.
A snapshot during the exploration and the results when the exploration finished \rev{is} shown in Fig.~\ref{fig:realworld_garage}(a) and Fig.~\ref{fig:realworld_garage}(b) respectively.
The challenging third scene is a more cluttered indoor environment with dimensions of $24 \times 6 \times 2$ \rev{m$^3$}, as shown in Fig.~\ref{fig:realworld_105}.
The obstacles are randomly placed in the scene, including boxes, tables, fans, bridges, etc.
The quadrotor accomplished exploring the space in $61.4$ seconds with a $53.9$ m trajectory.

In all three scenes, the proposed planner successfully achieves fast autonomous exploration with a limited sensor FoV quadrotor.
Notably, the executed trajectories involve very little back-and-forth movement or revisiting of previously explored regions, yielding remarkably efficient exploration motions with short durations and flight distances.
This superior performance is attributed to the capability of our exploration planner to generate reasonable coverage paths as global guidance and consistently adhere to this guidance during local planning.
Overall, the results of real-world experiments provide strong validation for the effectiveness and robustness of the proposed exploration planner in accomplishing fast autonomous exploration fully onboard in complex 3D environments. 
For more details, readers are referred to the supplementary video\footnote{\url{https://youtu.be/BGH5T2kPbWw}}, which provides complete exploration processes of three exploration trials in each of the three real-world environments.

\section{Conclusions}
\label{sec:conclude}
This paper presents \textbf{FALCON}, a fast autonomous exploration framework using coverage path guidance for aerial robots equipped with limited FoV sensors.
Whenever receiving the latest occupancy information, an incremental space decomposition and connectivity graph construction are performed to continually capture the underlying environment topology, facilitating efficient coverage path planning.
A hierarchical planner then generates a reasonable global coverage path spanning the entire unexplored space.
Then a local path is optimized to minimize traversal time and consciously incorporate the intention of \rev{the} coverage path guidance.
A lightweight exploration planner evaluation environment is developed to provide fair and comprehensive validation of autonomous exploration algorithms.
Through extensive benchmark experiments comparing with state-of-the-art methods, the proposed framework exhibits superior exploration efficiency, achieving $13.81\%\sim 29.67\%$ faster explorations.
Ablation studies and real-world experiments further validate the effectiveness of each proposed module and the capability \rev{to accomplish} fast autonomous exploration fully onboard in complex 3D environments.
The implementation has been made available for the benefit of the community.
In the future, we aim to take our exploration planning capabilities to the next level by expanding to multi-robot systems, which allows explorations on more complex environments with greater efficiency and flexibility.

\vspace{-0.2cm}

\addtolength{\textheight}{0.cm}   

\newlength{\bibitemsep}\setlength{\bibitemsep}{0.0\baselineskip}
\newlength{\bibparskip}\setlength{\bibparskip}{0.1pt}
\let\oldthebibliography\thebibliography
\renewcommand\thebibliography[1]{%
\oldthebibliography{#1}%
\setlength{\parskip}{\bibitemsep}%
\setlength{\itemsep}{\bibparskip}%
}

\bibliography{ref} 

\vspace{-1cm}

\begin{IEEEbiography}[{\includegraphics[width=1in,height=1.25in,clip,keepaspectratio]{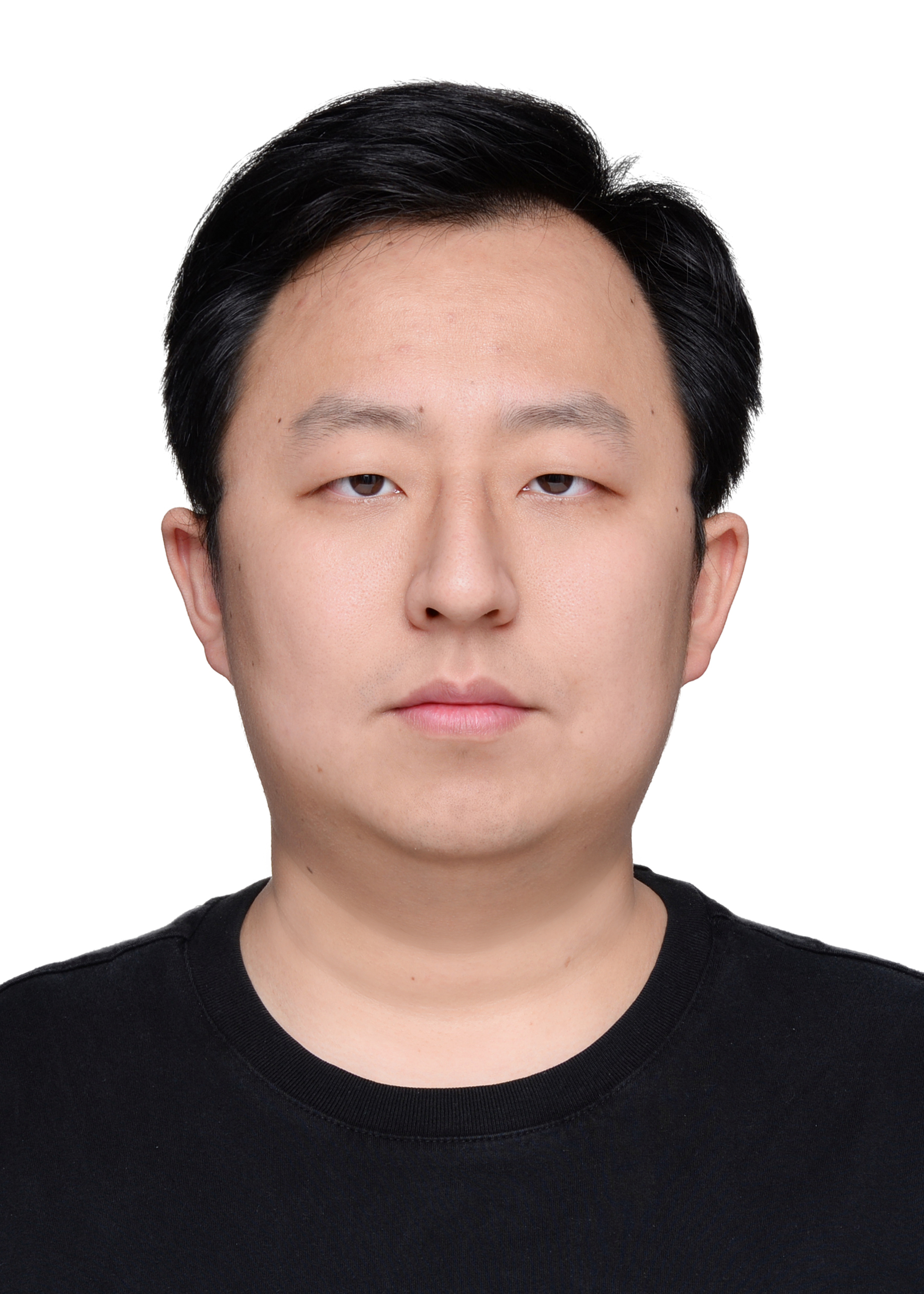}}]{Yichen Zhang}
    (Graduate Student Member, IEEE) received the B.Eng. degree in computer engineering in 2020 from the Hong Kong University of Science and Technology, Hong Kong, where he is currently working toward the Ph.D. degree in electronic and computer engineering under the supervision of Prof. Shaojie Shen. His research interests include autonomous exploration, aerial robot swarm, active SLAM, and motion planning.
\end{IEEEbiography}
\vspace{-1cm}
\begin{IEEEbiography}[{\includegraphics[width=1in,height=1.25in,clip,keepaspectratio]{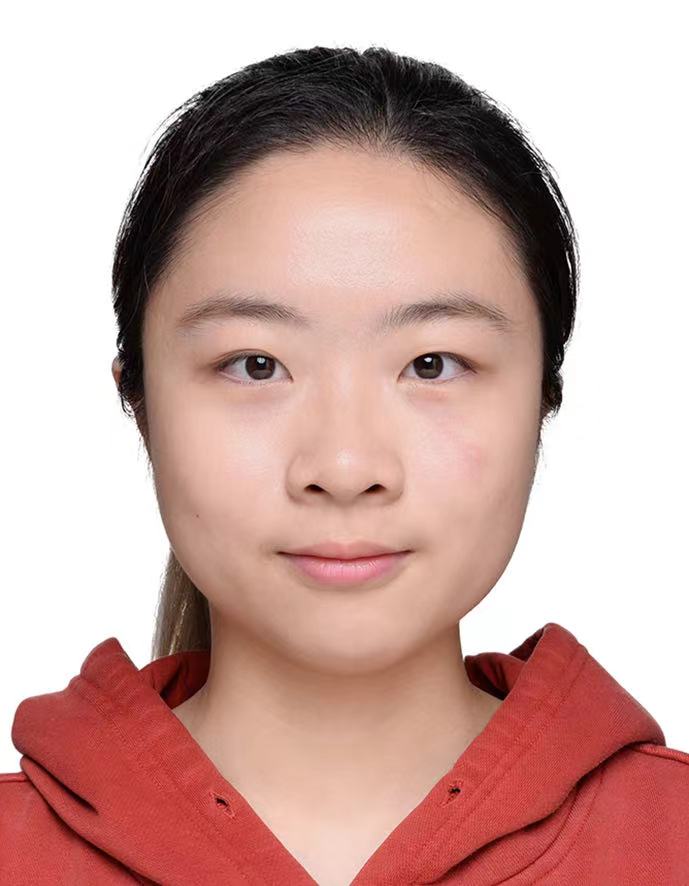}}]{Xinyi Chen}
    (Graduate Student Member, IEEE) received the B.Sc. degree in Mathematics, and in Computer Science from the Hong Kong University of Science and Technology, Hong Kong, in 2021. She is currently working towards the Ph.D. degree with the Hong Kong University of Science and Technology, Hong Kong, under the supervision of Prof. Shaojie Shen. Her research interests in robotics include motion planning, autonomous exploration, perception-aware planning, localization and mapping.
\end{IEEEbiography}
\vspace{-1cm}
\begin{IEEEbiography}[{\includegraphics[width=1in,height=1.25in,clip,keepaspectratio]{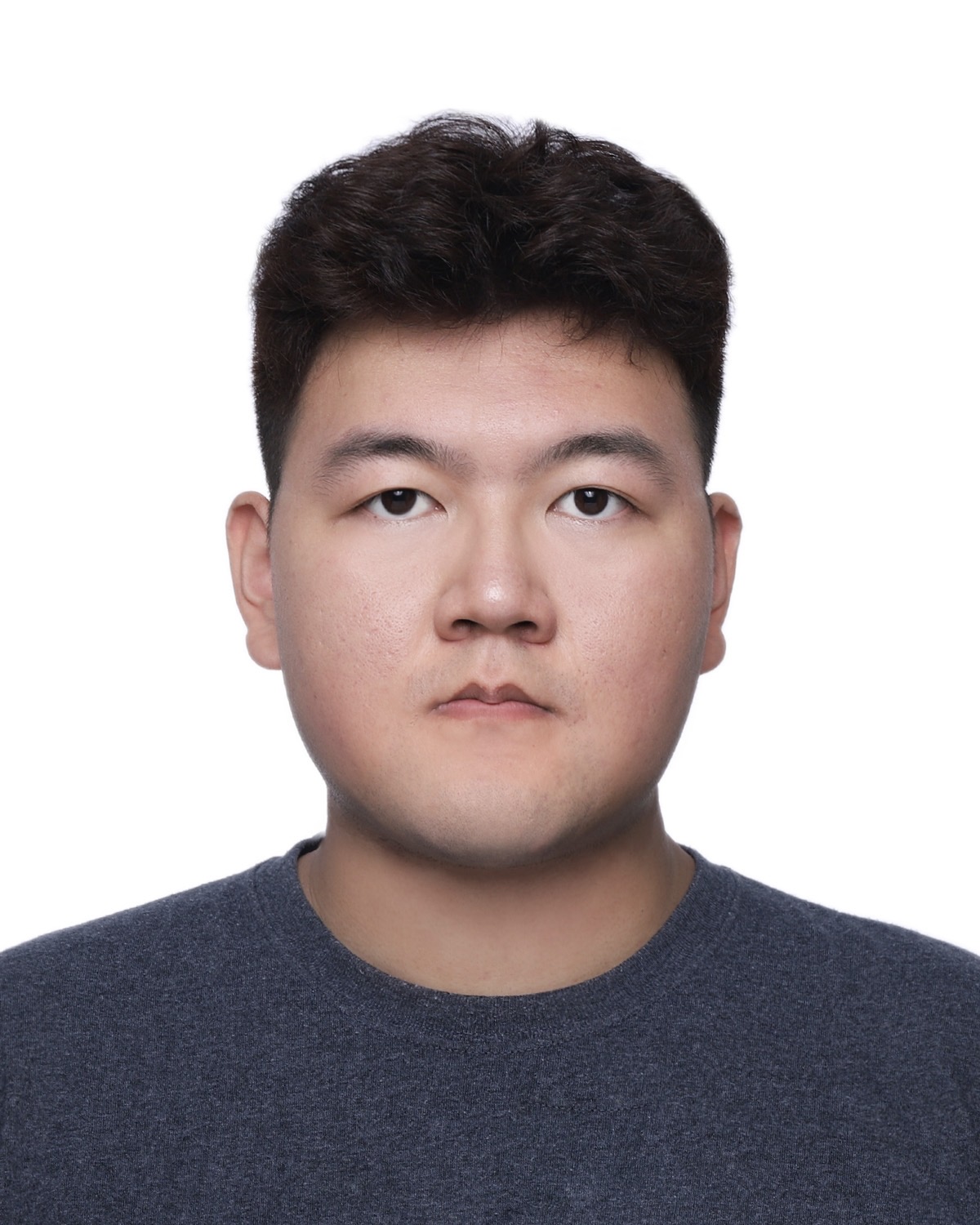}}]{Chen Feng}
    (Graduate Student Member, IEEE) received his B.Eng. degree in Mechatronics Engineering from the Harbin Institute of Technology, Harbin, China, in 2021. He is currently pursuing the Ph.D. degree in Electronic and Computer Engineering with the Aerial Robotics Group at the Hong Kong University of Science and Technology, Hong Kong, China. His research interests include intelligent scene perception and motion planning for aerial and mobile robots.
\end{IEEEbiography}
\vspace{-1cm}
\begin{IEEEbiography}[{\includegraphics[width=1in,height=1.25in,clip,keepaspectratio]{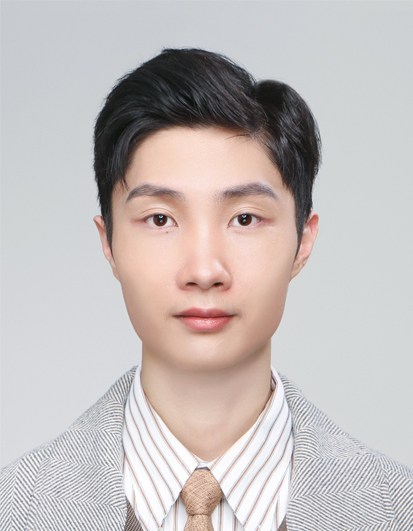}}]{Boyu Zhou}
    (Member, IEEE) received the Ph.D. degree in electronic and computer engineering from the Hong Kong University of Science and Technology, Hong Kong, China, in 2022. He is currently a tenure-track Assistant Professor (doctoral supervisor) at the Southern University of Science and Technology (SUSTech), where he serves as the Director of the Smart Autonomous Robotics (STAR) group. His research interests encompass aerial and mobile robots, motion planning, active perception, multi-robot systems, mobile manipulation, and vision-language navigation. He was the recipient of the IEEE TRO 2023 King-Sun Fu Best Paper Award, the IEEE RAL 2023 best paper award, and the IEEE ICRA 2024 Best UAV Paper Finalist. He is currently an Associate Editor for the IEEE International Conference on Robotics and Automation.
\end{IEEEbiography}
\vspace{-1cm}
\begin{IEEEbiography}[{\includegraphics[width=1in,height=1.25in,clip,keepaspectratio]{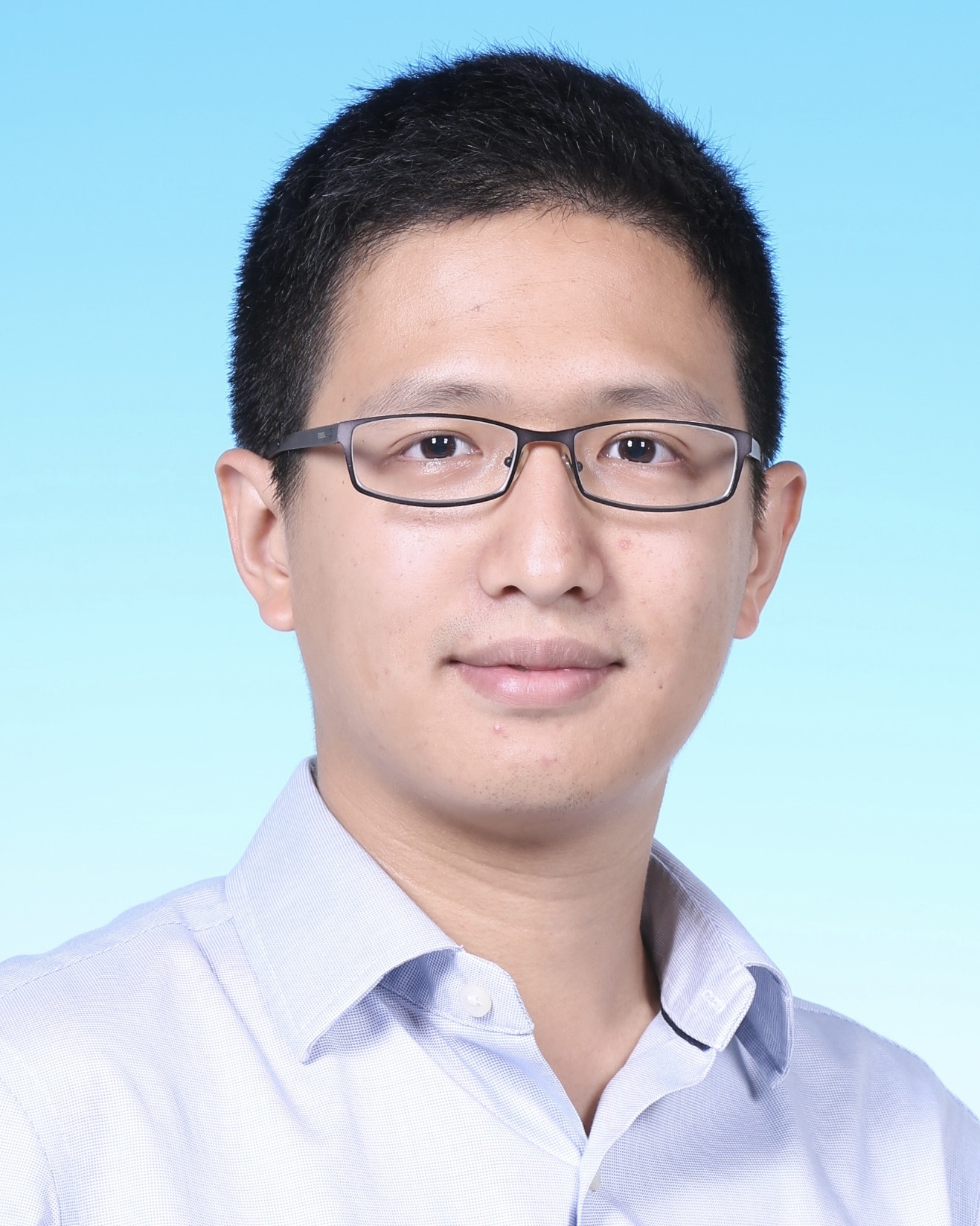}}]{Shaojie Shen}
    (Member, IEEE) received the B.Eng. degree in electronic engineering from the Hong Kong University of Science and Technology (HKUST), Hong Kong, in 2009, the M.S. degree in robotics and Ph.D. in electrical and systems engineering from the University of Pennsylvania, Philadelphia, PA, USA, in 2011 and 2014, respectively. 
    In 2014, he joined the Department of Electronic and Computer Engineering, HKUST, as an Assistant Professor, and was promoted to Associate Professor in 2020. He is the founding Director of the HKUST-DJI Joint Innovation Laboratory (HDJI Lab). His research interests include robotics and unmanned aerial vehicles, with focus on state estimation, sensor fusion, localization and mapping, and autonomous navigation in complex environments.
\end{IEEEbiography}

\end{document}